\documentclass{article}

\usepackage{rawfonts}

\usepackage{longtable}
\usepackage{multirow}
\usepackage{amssymb}
\usepackage{color}
\usepackage{graphicx}
\usepackage{booktabs}
\usepackage{algorithm}
\usepackage[noend]{algpseudocode}
\usepackage{caption}
\usepackage{rotating}
\usepackage{paralist}
\usepackage{setspace}
\usepackage{fullpage}
\usepackage{refstyle}
\usepackage{amsmath} 
\usepackage{amsthm}
\usepackage{authblk}
\usepackage{natbib}
\usepackage{tablefootnote}

\usepackage{dsfont}
\newtheorem{definition}{Definition}

\def\HiLi{\leavevmode\rlap{\hbox to 0.92\hsize{\color{lightgray}\leaders\hrule height .8\baselineskip depth .5ex\hfill}}}
\def\HiLiLong{\leavevmode\rlap{\hbox to 0.95\hsize{\color{lightgray}\leaders\hrule height .8\baselineskip depth .5ex\hfill}}}

\renewcommand{\epsilon}{\varepsilon}

\usepackage[T1]{fontenc}
\usepackage{kantlipsum}
\usepackage[usenames,dvipsnames]{xcolor}
\usepackage[breakable, theorems, skins]{tcolorbox}
\tcbset{enhanced}
\newtcolorbox{specialistbox}[1]{fonttitle=\bfseries,title=#1,colframe=gray!75!white}

\usepackage{caption}
\usepackage{subcaption}

\mathcode`*=\string"8000
\begingroup
\catcode`*=\active
\xdef*{\noexpand\textup{\string*}}
\endgroup

\definecolor{darkgreen}{rgb}{0,0.4,0.0}

\newcommand\identity{1\kern-0.25em\text{l}}

\DeclareRobustCommand{\Difficult}{$\textcolor{red}{*}$}

\makeatletter
\usepackage[implicit=true,unicode=true,
 bookmarks=true,bookmarksnumbered=true,bookmarksopen=true]
 {hyperref}
\hypersetup{
  pdftitle={my title},
  pdfauthor={my name},
  pdfsubject={my subject},
  pdfkeywords={my keywords},
  colorlinks=true,
  allcolors=blue,
  pdfpagelayout=OneColumn,
  pdfnewwindow=true,
}

\begin{document}
\author[1]{Natalia Ponomareva \thanks{dpfying-paper@google.com}}
\author[1]{Hussein Hazimeh}
\author[2]{Alex Kurakin}
\author[2]{Zheng Xu}
\author[3]{Carson Denison}
\author[3]{H. Brendan McMahan}
\author[1]{Sergei Vassilvitskii}
\author[2]{Steve Chien}
\author[2]{Abhradeep  Thakurta}

\affil[1]{Google Research, NYC}
\affil[2]{Google Research, MTV}       
\affil[3]{Google Research, Seattle}  

\title{How to DP-fy ML: A Practical Guide to Machine Learning with Differential Privacy}
\maketitle

\begin{abstract}
Machine Learning (ML) models are ubiquitous in real-world applications and are a constant focus of research. Modern ML models have become more complex, deeper, and harder to reason about. At the same time, the community has started to realize the importance of protecting the privacy of the training data that goes into these models.

Differential Privacy (DP) has become a gold standard for making formal statements about data anonymization. However, while some adoption of DP has happened in industry, attempts to apply DP to real world complex ML models are still few and far between. The adoption of DP is hindered by limited practical guidance of what DP protection entails, what privacy guarantees to aim for, and the difficulty of achieving good privacy-utility-computation trade-offs for ML models. Tricks for tuning and maximizing performance are scattered among papers or stored in the heads of practitioners, particularly with respect to the challenging task of hyperparameter tuning. Furthermore, the literature seems to present conflicting evidence on how and whether to apply architectural adjustments and which components are ``safe'' to use with DP. 

In this survey paper, we attempt to create a self-contained guide that gives an in-depth overview of the field of DP ML. We aim to assemble information about achieving the best possible DP ML model with rigorous privacy guarantees. Our target audience is both researchers and practitioners. Researchers interested in DP for ML will benefit from a clear overview of current advances and areas for improvement. We also include theory-focused sections that highlight important topics such as privacy accounting and convergence. For a practitioner, this survey provides a background in DP theory and a clear step-by-step guide for choosing an appropriate privacy definition and approach, implementing DP training, potentially updating the model architecture, and tuning hyperparameters.
For both researchers and practitioners, consistently and fully reporting privacy guarantees is critical, so we propose a set of specific best practices for stating guarantees.

With sufficient computation and a sufficiently large training set or supplemental non-private data, both good accuracy (that is, almost as good as a non-private model) and good privacy can often be achievable. And even when computation and dataset size are limited, there are advantages to training with even a weak (but still finite) formal DP guarantee. Hence, we hope this work will facilitate more widespread deployments of DP ML models.
\end{abstract}

\clearpage
\tableofcontents
\clearpage

\section{Introduction}
Differential Privacy (DP) \cite{dwork} has become a de facto standard for reasoning about information leakage. This well-established framework is starting to be adopted in industry \cite{thakurta22fldp_blog,xu2023gboard,snap_blog,covid,apple,msft21ppmlblog}~and the public sector \cite{census}, and is an active area of research. 

The term ``privacy'' has sometimes been used in the ML community quite loosely. Models are sometimes deemed ``private'' if they are robust to some empirical tests; for example membership inference, training data extraction, or private attribute inference attacks. ``Privacy'' is also often a shorthand term used to refer to DP, even though DP really only addresses data anonymization, and not other important privacy principles including transparency and consent into how data is used, or data minimization approaches that appropriately restrict access to raw data and intermediate computations \cite{bonawitz22cacm}. 

In this work, we concentrate on using differential privacy to provide anonymization\footnote{We use the term \textit{privacy guarantees} interchangeably with \textit{data anonymization} guarantees with respect to ML training data. For more precise definitions please refer to \citet{bonawitz22cacm}.} guarantees for the data used in training ML models.
Privacy protection, and even privacy definition is an active research topic, and DP is one of the most widely accepted concrete technologies which allows one to reason about data anonymization in a formal way. 
DP can be used for a wide variety of ML models, and DP methods can make models more robust to the aforementioned empirical privacy attacks. Of course, for ML to be effective, some information from the training data must be represented in the model, and so completely eliminating the chance of any possible inferences about the training data set being made from the model will be impossible if any utility is to be maintained. Arguably, the type of guarantee DP provides is thus the best one could hope for in terms of a domain-independent formal anonymization guarantee. 

In contrast, heuristic methods of protection against a particular attack do not provide the theoretical guarantees of DP.
The choice to apply DP-protection, heuristic methods, empirical privacy auditing, or a combination of these  is ultimately a business or policy decision, and sometimes none of these options is sufficient. Modern giant language models make this philosophical question concrete. \citet{brown2022does} find that neither DP nor data sanitation techniques nor robustification methods, while providing some data protection, fully reflect the privacy expectations of the people who contributed training data. This is due to multiple reasons, including the fact that for text data, coming up with an appropriate unit of privacy is hard (see additional discussion in Section \ref{sec:unitofprivacy}), there is a blurred line between public and private data, and the context where the data is revealed is important. 

While there has been much work applying DP to machine learning models, successfully doing this in practice remains challenging. First, privacy-utility tradeoffs (both real and perceived) discourage broad application of DP during training, and tricks to reduce this gap are scattered among different papers or are stored in the heads of practitioners. At the same time, many academic papers do not apply DP rigorously: in complex models like giant image or language models, the  simple application of popular DP training algorithms like DP-SGD is sometimes insufficient for a rigorous DP guarantee. For example, components like tokenizers \cite{DBLP:journals/corr/abs-1808-06226,wordpiece}, special layers like BatchNorm \cite{ioffe2015batch}, and hyperparameter tuning processes need to be adjusted or accounted for. 

This survey attempts to provide a comprehensive and self-contained guide on how to apply DP to complex ML models such as deep neural networks, with the goal of achieving the best performance and rigorous privacy guarantees.\footnote{Given the breadth and challenging nature of the topic, omissions and mistakes are quite possible. The authors welcome feedback on the work.} Our target audience is both academic researchers and practitioners. For academic researchers this work can serve as a one-stop survey of the current advances in the field of DP ML. Additionally, we cover in-depth important but often overlooked topics relevant for DP ML model training, e.g., privacy amplification via sampling, convergence of the DP-SGD algorithm, and user-level DP algorithms. We also touch upon the importance of providing fully quantified privacy statements, including private hyperparameter tuning and final $\epsilon$ reporting. Practitioners will benefit from clear definitions and explanations of what DP guarantees, descriptions of practical algorithms for obtaining DP models, discussion on how to choose their privacy budget, the importance of identifying the unit of protection, as well as tips and tricks for obtaining the best possible utility. 

Finally, we would like to highlight that this guide assumes that it is clear to the reader that DP is needed for their ML model. We do not discuss alternative methods like $k$-anonymity or heuristic methods of reasoning about or mitigating information leakage.\footnote{E.g., data deduplication can be an effective non-DP tool for reducing memorization \cite{DBLP:journals/corr/abs-2107-06499}.} We additionally assume a reasonable background in ML and deep learning in particular.

Throughout this work, we mark some sections as \Difficult, to illustrate that they provide additional in-depth theoretical details and can be skipped over without hurting the overall flow of thought. 

\begin{specialistbox}{Attention}
We use grey boxes like this one to draw the reader's attention to an important argument, conclusion or suggestion.
\end{specialistbox}

\subsection{Preview of the Later Sections}
This survey paper is organized as follows.
\begin{compactenum}
\item \textbf{Differential Privacy Basics} (Section \ref{sec:definitions}) provides the background information required to understand differential privacy. In particular, we introduce the two most common definitions of differential privacy (DP) and approximate differential privacy and discuss intuitive examples (Section \ref{sec:privacydeff}). Additionally we state the most important properties of DP (Section \ref{sec:properties}). We conclude this section with a sneak peak into popular mechanisms that can be used for achieving DP (Section \ref{sec:mechanisms}). These mechanisms will be employed throughout the later sections.
\item \textbf{DP-fying Basics} (Section \ref{sec:theory-basics}) describes the DP setting including threat models and release boundaries (Section \ref{sec:dp-settings}) and discusses where DP can be introduced (whether it is adding privacy to the data in Section \ref{sec:dp-input}, the serving process in Section \ref{dp:prediction-level}, or the algorithm in order to obtain a DP model. We explore algorithm modifications that provide partial training data protection in Section \ref{sec:label-protection} and we devote the full next section (Section \ref{sec:dp-training}) to modifications of the algorithm that results in full training data protection.
We also compare the guarantees each of these methods provide. 
\item \textbf{DP-Training for Full Training Data Protection} (Section \ref{sec:dp-training}) is devoted to  an in-depth discussion of the most common way of obtaining a DP ML model -- by modifying the training algorithm. We introduce one of the most popular algorithms for DP-Training -- DP-SGD (Section \ref{sec:dpsgd}) and discuss advanced topics on DP-SGD convergence (Section \ref{sec:dpsgd-convergence}) and privacy accounting (Section \ref{sec:dpaccounting} and \ref{sec:amplification}). 
We then explore DP-Training algorithms that provide user-level privacy (as opposed to example-level) guarantees (Section \ref{sec:dp-user-modifications}).
Finally, we conclude with a discussion on what makes the adoption of DP-SGD hard in practice (Section \ref{sec:problems-dp-sgd}).
\item \textbf{Practicalities of DP-Training} (Section \ref{sec:practicalities})  is specifically designed for practitioners and focuses on all stages of applying DP-Training to an ML model. We start by highlighting the importance of what unit of protection to use (Section \ref{sec:unitofprivacy}). We then discuss what is currently considered a ``good'' level of protection for an ML model and suggest privacy guarantees to target, as well as outline the reasoning as to why these guarantees have a meaning (Section \ref{sec:good-eps}). We state how privacy guarantees can be calculated and argue for a rigorous way of reporting such guarantees in Section \ref{sec:calculating-and-reporting-eps}. We then present an analysis of the importance of hyperparameter tuning for maximizing the utility of DP-Training methods (Section \ref{sec:tuning_hyperparams}), introduce step-by-step tuning algorithms and describe how to account for such tuning (Section \ref{sec:hyperparams_eps}). Finally, we highlight the need for careful model architectural design (Section \ref{sec:architecture}) and multi-device distribution consideration (Section \ref{sec:microbatches}). We conclude with a brief overview of popular DP libraries.

\end{compactenum}

\section{Differential Privacy: Definitions, Intuition and Properties}\label{sec:definitions}
 In this section we introduce common differential privacy definitions, outline DP properties and popular mechanisms which will be employed throughout the later sections.

\subsection{Definitions}\label{sec:privacydeff}

Differential privacy (DP) was originally introduced in the context of databases that store private information about individuals. For example, consider a hospital admissions dataset, where each row may contain sensitive information about a patient, such as their demographics, medical history, insurance, and payment information. As part of analyzing the dataset, an analyst may want to issue a query to obtain aggregate-level statistics, for instance, average bill, or median hospital stay length.
Informally, differential privacy requires that the result of the query be insensitive to the removal of any single row in the database. Thus, differential privacy protects against leaking information about individual rows (e.g., patients in this case). In what follows, we will use a standard machine learning terminology and refer to the database of interest as a \textsl{dataset}. Next, we formalize the notion of differential privacy.

\paragraph{Setup and notation.} Let $D$ be a dataset consisting of $n$ records. An analyst would like to query the dataset. Formally, the query is a function $f$ that takes a dataset as input and outputs a quantity of interest. The query could be as simple as computing the mean of a certain feature, or, more complex, such as training a neural network and then returning the network's weights. DP is achieved via a \textsl{mechanism} $\mathcal{A}$: a randomized algorithm that approximates the result of $f$. One popular class of mechanisms can be thought of as a ``noisy'' version of $f$, for example, adding judiciously chosen noise to $f$, i.e.,  $\mathcal{A}(D) = f(D) + Z$, where $Z$ is a random variable sampled from a specific noise distribution\footnote{We note that not all mechanisms used in the literature are additive as in this example.}.

\paragraph{Differential privacy.} What makes a mechanism $\mathcal{A}$ differentially private? We formalize this notion next. 

\begin{definition}[Differential Privacy - \cite{DMNS}] \label{def:exact_dp}
We say that two datasets $D$ and $D'$ are \textsl{neighbors} if they differ in exactly one record; more precisely, one dataset is a copy of the other but with a single record added or removed\footnote{While this is the most common notion of adjacency, we discuss other possible definitions in Section \ref{sec:recordadjacency}}. Let $\epsilon$ be a positive scalar. A mechanism  $\mathcal{A}$ guarantees $\epsilon$-differential privacy if for any two  neighboring datasets $D$ and $D'$, and for any $S \subseteq \text{Range}(\mathcal{A})$,
\begin{align}
    P[\mathcal{A}(D) \in S] \leq \exp(\epsilon) \times P[\mathcal{A}(D') \in S].
\end{align}
\end{definition}

In Definition \ref{def:exact_dp}, Range$(\mathcal{A})$ refers to the set of all possible outcomes of $\mathcal{A}$. Technically, the set $S$ in the definition must be measurable.

Definition \ref{def:exact_dp} guarantees that the probability of seeing a specific output on any two neighboring datasets can differ by at most a multiplicative factor of $\exp(\epsilon)$. When  $\epsilon$ is sufficiently small, the main implication of the definition is that including or excluding a single record from the dataset is not likely to change the output.  Thus, an adversary who only has access to the output of $\mathcal{A}$ will have a difficult job inferring whether any particular record is present in the dataset. 

The choice of what constitutes a  ``record'' (the \emph{unit of privacy}) is central to interpreting the definition of DP and the semantics of the guarantees it provides. Different units of privacy can be appropriate for different ML applications.  For simplicity, we will generally focus on the case where a record corresponds to a single training example, resulting in \emph{example-level DP} (also called \textit{instance-level DP}). However, in many applications (particularly those where training data is generated by users, and one user might contribute a large number of training examples), it may be preferable to define a ``record'' as encompassing all the data from a user (\emph{user-level DP}). In other applications, one might also partition data based on the time of generation. In any case, the neighbor definition will then guide the specific near-indistinguishability guarantees given by differential privacy.  When one record consists of multiple training examples, different DP mechanisms may be required (Section \ref{sec:unitofprivacy}). 
\\ \\
\noindent \textsl{Choice of $\epsilon$.}  The parameter $\epsilon$ is called the \textsl{privacy parameter} or the \textsl{privacy budget}. It controls the level of protection provided by Definition \ref{def:exact_dp} for the specific unit of privacy: smaller $\epsilon$'s provide more protection because the mechanism's output distributions on  neighboring datasets become closer. 
Generally, there is a trade-off between $\epsilon$ and the utility of the mechanism (e.g., accuracy of a neural network); smaller $\epsilon$'s typically lead to lower utility if other variables like the dataset size and batch size remain constant. As an extreme example, when $\epsilon=0$, it is easy to see that the output of the mechanism becomes independent of the input, i.e., all datasets will lead to the same output distribution. Of course, such an  input-independent mechanism is expected to have very limited use. In practice, we need to achieve a balance between an $\epsilon$ that provides a good level of privacy without sacrificing much on  utility. The particular choice usually depends on the application. For common statistical database queries (e.g., mean of a column), $\epsilon$ is typically chosen to be less than one. In deep learning, this choice is usually relaxed to $\epsilon \leq 10$ (see Section \ref{sec:good-eps} for a discussion). We also emphasize that the shape and location of the privacy-utility tradeoff curve is strongly influenced by dataset size and the amount of computation used during training (e.g., batch size). With a sufficiently large training set and sufficient computation, for a fixed model both good accuracy (that is, almost as good as a non-private model) and good privacy can often be achievable. Hence, the relevant question is not usually ``\emph{Will DP work for my model?}'' but rather ``\emph{How much computation and data do I need to achieve reasonable privacy and utility?}''.

\paragraph{Approximate differential privacy.} In context of private ML models, a relaxation of pure $\epsilon$-DP Definition~\ref{def:exact_dp} has been commonly used instead. This is due to a number of reasons, including obtaining better utility and other advantages like easier and tigher privacy accounting for composing several DP mechasims (see Section \ref{sec:properties}), while preserving the strong semantics of DP~\cite{semantics}. In this work, we primarily  concentrate on the following \textit{Approximate DP} relaxation~\cite{ODO}:

\begin{definition}[$(\epsilon, \delta)$-Differential Privacy, \cite{ODO}] \label{def:apprx_dp}
Let $\epsilon$ and $\delta \le 1$ be two non-negative scalars. A mechanism  $\mathcal{A}$ is $(\epsilon, \delta)$-differentially private if for any two neighboring datasets $D$ and $D'$, and for any $S \subseteq \text{Range}(\mathcal{A})$,
\begin{align}\label{eq:dwork}
    P[\mathcal{A}(D) \in S] \leq \exp(\epsilon) \times P[\mathcal{A}(D') \in S] + \delta.
\end{align}
\end{definition}
The $(\epsilon, \delta)$ definition is a relaxation of the $\epsilon$ definition, which allows the two probability terms in Definition \ref{def:exact_dp} to differ by the additive scalar $\delta$. Thus,  $\delta$ controls the strength of the relaxation, with smaller values leading to stronger privacy guarantees. While for $\delta > 0$, this definition generally ``fails'' to satisfy $\epsilon$-DP, it is important to make a distinction between two types of failure that the definition allows. The first is ``catastrophic'' where parts of, or even the whole dataset, is likely to be output publicly. The second type is ``graceful'', in the sense that the $\epsilon$ definition does not hold exactly, but a looser bound may still hold. As an example of graceful degradation, consider an $(\epsilon, \delta)$ mechanism that is also guaranteed to be $(2\epsilon, 0)$-DP. While this mechanism fails to satisfy $\epsilon$-DP, it does satisfy exact DP with a privacy level of $2\epsilon$, so it cannot fail catastrophically. Fortunately, common mechanisms for $(\epsilon, \delta)$-DP in the literature, such as the Gaussian mechanism that we discuss in Sec. \ref{sec:mechanisms}, do not fail catastrophically \footnote{Rather, for any arbitrarily small $\delta$, there exists an $\epsilon$ value such that the mechanism has $(\epsilon, \delta)$ guarantees  \cite{DBLP:journals/corr/Mironov17}}. 

Since $\delta$ controls the strength of the relaxation, it is important to make sure that a sufficiently small $\delta$ is used. The general recommendation in the literature is to choose $\delta \ll \frac{1}{n}$, where $n$ is the number of records in the dataset \cite{dwork}. This recommendation stems from a worst-case analysis. Specifically, consider the following worst-case assumption on every record: if the record $r$ is present in the dataset, the $(\epsilon, \delta)$ mechanism will generate a certain output $E_r$ with probability $\delta$, and furthermore, $E_r$ cannot happen otherwise. If an attacker observes $E_r$, they can directly deduce that the record $r$ is in the dataset. Thus, each record in the dataset has a probability $\delta$ of being successfully identified by the attacker in this worst-case scenario. The expected number of successful attacks is $\delta n$. Choosing $\delta \ll \frac{1}{n}$, will ensure that the expected number of successful attacks is much smaller than $1$.

\subsubsection{Alternative Neighboring Criteria} \label{sec:recordadjacency}
\begin{specialistbox}{Neighboring criteria}
The DP definition can be parameterized with different ways records are allowed to change to form a neighboring dataset:  \emph{add-or-remove} one record, \emph{zero-out} one record, or \emph{replace-one} record. The first two have comparable semantics for a fixed $\epsilon$, whereas the guarantee for \emph{replace-one} is approximately twice as strong. Care should therefore be taken when comparing $\epsilon$s based on different criteria.
\end{specialistbox}

The choice of what constitutes neighboring datasets is key to Definition \ref{def:exact_dp}. The primary question of what constitutes a single record (the unit of privacy) was discussed above, and is treated in more depth in Section \ref{sec:unitofprivacy}.  There is also a more technical aspect to the definition, which is how records are allowed to change between neighboring datasets (independent of what defines a ``record''). The addition or removal of a single record (\emph{add-or-remove}, as in Definition \ref{def:exact_dp}) is particularly common. However, because this changes the size of the dataset, complications can arise when applying this definition in some settings. For this reason, it may be technically preferable to instead use a \emph{zero-out} notion where datasets are adjacent if any one record is replaced with a special ``zero'' record (often exactly zero for numeric data) \cite{erlingsson20esa,kairouz21practical}. While this technically produces a slightly different guarantee, $\epsilon$'s for add-or-remove and zero-out DP are essentially semantically equivalent.

A third common definition is \emph{replace-one} which allows one record to be replaced with an arbitrary different record \cite{vadhan2017complexity}. This is equivalent to combining the addition and the removal of a record; this definition can roughly be thought of as producing guarantees that are ``twice as strong'' as the other two.\footnote{Technically, the $\ell_2$-sensitivity (see Definition \ref{def:sensitivity}) will typically be twice as large under replace-one. To see this, imagine we are computing a sum of scalar records from the range $[-1, 1]$. Add-or-remove and zero-out can both change the sum by at most 1, but replacement can change the sum by 2 (switching a $-1$ to a $+1$).} Hence, when comparing specific $\epsilon$ values it is essential to confirm that a comparable adjacency criteria and unit-of-privacy is being used.

\subsection{Properties of DP} \label{sec:properties} 

Definitions \ref{def:exact_dp} and \ref{def:apprx_dp} satisfy two important properties: composition and invariance to post-processing. Specifically, composing or post-processing multiple DP mechanisms is guaranteed to remain differentially private (albeit the privacy parameters do degrade upon composition).
Thus, DP procedures for complex systems can be designed in a modular way by combining and transforming the outputs of many building-block DP mechanisms. As an example, pre-processing a dataset using a DP algorithm and then training a model using another DP algorithm is guaranteed to be DP. Next we discuss these properties in detail.

\paragraph{Sequential composition.} Applying multiple DP mechanisms to the same dataset remains differentially private but with some degradation in the privacy parameters. There are different composition bounds in the literature for quantifying this degradation. One basic composition bound states that the $\epsilon$ and $\delta$ after applying multiple mechanisms is the sum of the $\epsilon$'s and $\delta$'s of the individual mechanisms. More formally, let $\mathcal{A}_1, \dots, \mathcal{A}_t$ be a set of $t$ mechanisms  where the i-th mechanism satisfies $(\epsilon_i, \delta_i)$-DP. Sequential composition states the joint output of the mechanisms, i.e., $(\mathcal{A}_1, \dots, \mathcal{A}_t)$, is $(\epsilon', \delta')$-DP where $\epsilon' := \sum_i \epsilon_i$ and $\delta' := \sum_i \delta_i$ \cite{dwork}. The $\epsilon'$ in the latter bound can be improved at the expense of some degradation in $\delta'$, using  \textsl{advanced composition} bounds  \cite{dwork,kairouz2015composition}. Alternatively, tighter bounds can be obtained for sequential composition by exploring more fine-grained properties of $\mathcal{A}_1, \dots, \mathcal{A}_t$:  e.g., for the composition of exponential mechanisms \cite{pmlr-v119-dong20a}.

\paragraph{Parallel composition.} Recall that in sequential composition all mechanisms were applied to the same dataset. In contrast, parallel composition assumes that the dataset is partitioned into mutually disjoint subsets, and each mechanism is applied to one unique subset. As before, we denote the set of mechanisms by  $\mathcal{A}_1, \dots, \mathcal{A}_t$,  where the i-th mechanism satisfies $(\epsilon_i, \delta_i)$-DP. Parallel composition guarantees that the combined mechanism, i.e., $(\mathcal{A}_1, \dots, \mathcal{A}_t)$, is $(\max_i \epsilon_i, \max_i \delta_i)$-DP. The guarantee here is stronger than that of sequential composition. Intuitively, this statement holds because in parallel composition the combined mechanism uses each record once, whereas in sequential composition each record is used multiple times.

\paragraph{Invariance to post-processing.} Applying any data-independent transformation to a DP mechanism is guaranteed to remain differentially private (with the same privacy parameters) \cite{dwork}. This property has two important implications. First, it is impossible for an attacker to weaken the DP guarantee by post-processing the mechanism's output. Second, this property can be used to simplify the design and analysis of complex DP systems. For example, training a neural network with SGD is essentially a post-processing of gradients computed at successive iterations. Thus, based on the post-processing property, differentially private training of a neural network can be achieved by using differentially private gradients in each iteration; this method will be discussed in more detail in Section \ref{sec:dpsgd}.

\paragraph{Converting from example-level to user-level privacy (group privacy guarantees).} In some cases, it is possible to use \emph{group privacy} theorems to convert guarantees for a ``smaller'' unit of privacy to a guarantee for a ``larger'' unit of privacy.  For example, consider a domain where we train a model on examples coming from users.  If we train with an example-level $(\epsilon, \delta)$-DP guarantee, we can infer a $(k \epsilon, k e^{k\epsilon}\delta)$-DP guarantee when up to $k$ examples are changed arbitrarily, following e.g. \citet[Lemma 2.2]{vadhan2017complexity}.

Suppose the maximum number of examples any one user can contribute to the training data is capped at $k=20$, and we train with a example-level $(\epsilon{=}2, \delta{=}10^{-24})$-DP guarantee. Using the above result, we can infer a user-level $(\epsilon{=}40, \delta{=}4.7{\times}10^{-6})$-DP guarantee. The substantial degradation of both $\epsilon$ and $\delta$ in this case suggests that using DP mechanisms that directly provide user-level privacy may be preferable (Section~\ref{sec:dp-user-modifications}), or that a smaller cap on the number of examples allowed per user should be chosen.

\subsection{Alternative Stronger Relaxations of DP\Difficult} As discussed earlier, approximate DP (Definition \ref{def:apprx_dp}) is a relaxation of exact DP (Definition \ref{def:exact_dp}). However, composition bounds using ($\epsilon, \delta$)-DP have been shown to be loose even for advanced composition. To address this issue, stronger relaxations of exact DP with much tighter bounds for composition have been introduced in the literature. Popular examples include \textsl{zero-Concentrated Differential Privacy (zCDP)} \cite{bun2016concentrated}  and \textsl{R\'enyi Differential Privacy (RDP)}  \cite{DBLP:journals/corr/Mironov17}. Similar to the setup of Definition \ref{def:exact_dp}, let us consider two arbitrary neighboring datasets $D$ and $D'$ and a mechanism $\mathcal{A}$. On a high level, both zCDP and RDP guarantee that the ``distance'' (technically, the R\'enyi divergence) between the distributions of $\mathcal{A}(D)$ and $\mathcal{A}(D')$ is below a certain threshold, for any two neighbors $D$ and $D'$. Intuitively, since the two distributions are close, it is improbable for an attacker to deduce which of the neighboring datasets was used by the algorithm. These two definitions do not allow for catastrophic failures  and are stronger than approximate DP. Specifically, any zCDP or RDP guarantee can be converted to an approximate DP guarantee. In Section \ref{sec:dpaccounting}, we will present and discuss RDP more formally.

\subsection{Basic DP Mechanisms\Difficult} \label{sec:mechanisms}
As discussed earlier, differential privacy is typically integrated in complex systems in a modular fashion, by relying on building-block mechanisms. While many mechanisms have been proposed in the literature, we will focus here only on three fundamental mechanisms that are essential to the training and hyperparameter tuning algorithms discussed in the rest of the paper. Specifically, we will discuss (i) the Laplace and Gaussian mechanisms for queries with numerical outcomes, and (ii) the Exponential mechanism for queries with arbitrary outcomes (not necessarily numeric).

We start with the Laplace and Gaussian mechanisms. We assume that the query $f$  returns an output in $\mathbb{R}^{k}$. As the names suggest, the Laplace and Gaussian mechanisms add noise sampled from the Laplace and Gaussian distributions, respectively, to $f$. The variance of these noise distributions will depend on the \textsl{$\ell_p$-sensitivity} of $f$, which we discuss next. 

\paragraph{$\ell_p$-Sensitivity.} The $\ell_p$-sensitivity refers to the maximum possible change in the function output (measured using the $\ell_p$ norm) when a single record is added or deleted from the input. We define this notion more formally below.
\begin{definition}[$\ell_p$\text{-sensitivity}] \label{def:sensitivity}
Let $f$ be a query mapping from the space of datasets to $\mathbb{R}^k$. Let $N$ be the set of all possible pairs of neighboring datasets, i.e., \\$N = \{ (D,D') \ | \  D \text{ and } D' \text{ are neighbors}\}$. For a fixed positive scalar $p$, the $\ell_p$-sensitivity of $f$ is defined by
\begin{align} \label{eq:l1_sensitivity}
  S(f;p) = \max_{D, D' \in N} \| f(D) - f(D') \|_p.
\end{align}
\end{definition}
Note that this definition of sensitivity is \textsl{global} in the sense that it does not depend on the dataset we want to run the algorithm on, but on a worst-case pair of neighbors.  
As an example of Definition \ref{def:sensitivity}, a query $f$ that counts the number of records in $D$ has $S(f;1) = 1$  (because, by definition, one of the neighboring datasets has exactly one additional record). In other cases, however, the sensitivity may be unbounded or difficult to estimate. For example, assuming that the entries of the dataset can take arbitrary values, the query that adds (or averages) all of the entries of the dataset has an infinite sensitivity, since the additional record could take on an arbitrarily large value. Another important example is the gradient of an arbitrary function such as the loss of a neural network. When no assumptions are placed on the function, the gradient can generally have infinite sensitivity, or possibly a finite sensitivity that is difficult to compute or bound. As we will discuss, the Laplace and Gaussian mechanisms require the sensitivity of the query to be bounded and this bound to be known. For queries with unbounded or unknown sensitivity, this issue is commonly solved by \textsl{clipping} either the entries of the dataset or the output of the query to be within a bounded range (or to have a bounded norm). The choice of the range is a critical parameter, and leads to a bias-variance trade-off \cite{AKMV}. For typical queries, such as the evaluation of the sum, mean, or gradient, clipping leads to a bounded sensitivity that is easy to compute. (Gradient clipping will be used in Section \ref{sec:dpsgd} for one of the most common DP-Training algorithms).

\paragraph{Laplace mechanism.} Before presenting the mechanism, we first review the Laplace distribution. Given a positive scalar $b$, the Laplace distribution centered at zero is characterized by the probability density function $g(x | b) = \frac{1}{2b} \exp(-|x|/b)$ where $x \in \mathbb{R}$. The scalar $b$ is called the \textsl{scale parameter}, and the variance is given by $2b^2$. 
Given the output of the query $f(D) \in \mathbb{R}^{k}$, the Laplace mechanism adds noise sampled from the Laplace distribution to each of the $k$ dimensions in the output, with the variance of the noise calibrated to the $\ell_1$-sensitivity, i.e., $S(f;1)$. Specifically, the Laplace mechanism is defined by 
\begin{equation*} 
    \mathcal{A}_{L}(D; f, \epsilon) := f(D) + (Z_1, Z_2, \dots, Z_k), \qquad Z_i \overset{\text{ind.}}{\sim} \text{Laplace}\Big( S(f;1)/\epsilon \Big) \ \ \forall i,
\end{equation*}
where Laplace($b$) denotes the Laplace distribution with parameter $b$. 
The Laplace mechanism $\mathcal{A}_{L}(D; f, \epsilon)$ is guaranteed to be $\epsilon$-DP \cite{dwork}. 
The scale parameter (and the variance) used by the Laplace mechanism is increasing in $S(f;1)$. Intuitively, this is expected because queries with larger sensitivity can change more significantly with the addition or deletion of a single record, and thus require more noise to ``hide'' the change. Also note that the scale parameter decreases with $\epsilon$, meaning that tighter DP guarantees require higher levels of noise. The Laplace mechanism is used in a range of DP training algorithms, as we discuss in Section \ref{sec:dp-training} and Appendix  \ref{app:nondif}.

\paragraph{Gaussian mechanism.} The Gaussian mechanism is an alternative to the Laplace mechanism, which has a similar mode of operation but samples noise from a normal distribution. The Gaussian mechanism cannot guarantee pure $\epsilon$-DP but can instead ensure approximate $(\epsilon, \delta)$-DP\footnote{The Gaussian mechanism also satisfies stronger relaxations of DP, such as zCDP and RDP.}. Despite its weaker guarantee, the Gaussian mechanism is commonly used in machine learning, e.g., it is the main mechanism behind DP-SGD (Section \ref{sec:dpsgd}). A main reason behind its wide adoption is that it can work with less noise than the Laplace mechanism, especially when the output of the query is high-dimensional (i.e., large $k$). 

Formally, assuming $\epsilon \in (0,1)$, the classical Gaussian mechanism\footnote{The classical Gaussian mechanism we discuss here is only guaranteed to satisfy $(\epsilon,\delta)$-DP for $\epsilon \in (0,1)$. See \citet{balle2018improving,zhao2019reviewing} for improved versions of the Gaussian mechanism that work for $\epsilon \geq 1$.} samples noise from $N(0, \sigma^2)$ with $\sigma = S(f;2) \sqrt{2\ln(1.25/\delta)}/\epsilon$ \cite{dwork}, where $S(f;2)$ is the $\ell_2$-sensitivity. This is in  contrast to the Laplace mechanism, which uses $\ell_1$-sensitivity. Since $\| x \|_2 \leq \| x \|_1$, we always have $S(f;2) \leq S(f;1)$. In fact, $S(f;2)$ can be significantly smaller in high dimensional settings, allowing the Gaussian mechanism to use noise with less variance than the Laplace mechanism (assuming the term $\ln(1.25/\delta)$ is sufficiently small). Moreover, the tails of the normal distribution decay faster than those of the Laplace distribution. Therefore, even if the two distributions have the same variance, the Gaussian distribution is more likely to sample noise with a smaller magnitude for tail events.

\paragraph{Exponential mechanism.} As discussed earlier, the Laplace and Gaussian mechanisms can only handle queries with numeric output. In many applications, the answer to a query may not be numeric or possibly numeric but discrete (e.g., fractional values are not allowed). For example, tuning a neural network requires answering queries such as ``what is the best model architecture that maximizes performance?''. The answer to this query is an architecture, which cannot be directly privatized by adding noise. The exponential mechanism is a differentially private selection algorithm \cite{mcsherry2007mechanism}, which can be useful in such applications where queries output arbitrary ``objects'', such as models, text, or numbers.

Given a public set of objects $\mathcal{R}$ (e.g., candidate model architectures), the exponential mechanism seeks to (approximately) pick the ``best'' object in the set. The notion of object quality is quantified using a scoring function and depends on a dataset of interest. Specifically, given a private dataset $D$ and an object $r \in \mathcal{R}$, we define a scoring function $G(D, r)$, which returns a scalar that quantifies how good $r$ is w.r.t. $D$ (where  higher scores are interpreted as better). The set $\mathcal{R}$ and the function $G$ are assumed to be public and are chosen by the analyst, while we recall that  the dataset $D$ is private. The goal is thus to make sure that releasing some $r \in \mathcal{R}$ does not reveal sensitive information about the records of $D$. To achieve this, the exponential mechanism $\mathcal{A}_E(D; G, R, \epsilon)$ randomly samples a single element  from $\mathcal{R}$, where the sampling probability is defined by:
\begin{equation} \label{eq:exponential_mechanism}
P\Big(\mathcal{A}_E(D; G, R, \epsilon ) = r \Big) \propto  \exp{\Bigg(\frac{\epsilon G(D, r)}{2 \Delta}}\Bigg), ~ \forall r \in \mathcal{R},
\end{equation}
and $\Delta := \max_{r \in \mathcal{R}} S\big(G(., r); 1 \big)$ is the maximum sensitivity of the scoring function. The mechanism $\mathcal{A}_E(D; G, R, \epsilon)$ is guaranteed to satisfy $\epsilon$-DP \cite{mcsherry2007mechanism}. As evident from Eq.~(\ref{eq:exponential_mechanism}), the mechanism assigns exponentially higher probabilities to better objects (i.e., ones with higher scores). Moreover, the less sensitive the scoring function is, the more likely the best object will be selected. We remark that the exponential mechanism is very general and can recover a wide class of DP mechanisms (including the Laplace mechanism) for suitably chosen scoring functions. In Section \ref{sec:tuning_hyperparams}, we discuss how some existing approaches for DP hyperparameter tuning rely  on the exponential mechanism. 

Another popular alternative for private selection is the report-noisy-max mechanism \cite[Chapter 3]{dwork}, which requires the set of objects $\mathcal{R}$ to be finite. Unlike the exponential mechanism, report-noisy-max adds noise (e.g., from a Laplace distribution) to the object scores and then outputs the object with the maximum score\footnote{Under Gumbel noise, report-noisy-max is known to be equivalent to the exponential mechanism.}. We also note that there is an extensive literature on alternative private selection mechanisms, e.g., see \cite{beimel2013private,chaudhuri2014large,lantz2015subsampled,minami2016differential,raskhodnikova2016lipschitz,liu2019private,awan2019benefits,mckenna2020permute}. These alternative mechanisms may work better than the exponential mechanism in specific settings, such as when the scoring function has a high sensitivity or when the set $\mathcal{R}$ is large.

Besides the basic mechanisms discussed above, there are many other fundamental DP techniques and frameworks in the literature. For example, the Sparse Vector Technique \cite[Chapter 3]{dwork} can be used to obtain tight DP guarantees in settings where there is a stream of numeric queries and where the goal is to identify one (or a small number of queries) whose output lies above a certain threshold. There are also several popular frameworks that can guarantee DP in settings where the global (worst-case) sensitivity in Definition \ref{def:sensitivity} is large; for example, the Sample-and-Aggregate \cite{sampleaggregate} and Propose-Test-Release frameworks \cite{dwork2009differential}. These frameworks rely on a local notion of sensitivity, which is typically smaller than global (worst-case) sensitivity. For more discussion and a survey of additional techniques and frameworks, we refer the reader to \citet{dwork}.

\section{DP-fying Basics: Settings and Methods }\label{sec:theory-basics}
This section provides details on how to achieve training data protection via DP datasets, models and predictions. 

We first cover an important question that a practitioner should answer before choosing a DP method: how will the model be used (accessed) and what threat mode do we need to mitigate (e.g., protection from a rogue user that has access to ML model predictions vs protection from an untrusted service provider). We cover this topic in Section \ref{sec:dp-settings}. We proceed to explore where DP can be introduced in Section \ref{sec:wheretoapply} and then we  present different DP techniques, including \textit{modifications to the training data} (Section \ref{sec:dp-input}), \textit{inference process} (Section \ref{dp:prediction-level}) and the modification to the \textit{training algorithm} that results in \textit{partial training data} protection (Sections \ref{sec:label-protection}). We explore the \textit{modification to the training algorithm that protects full training data} in the next Section \ref{sec:dp-training}.

While there has been some work that attempts to reason about \textit{intrinsic} Differential Privacy of some \textit{unmodified/standard ML components} (e.g., SGD \cite{hyland2019on} and bagging \cite{DBLP:journals/corr/abs-2008-09845}), this subfield is still very much nascent and will not be explored in this paper. 

\subsection{DP Settings: Threat Models and Release Boundaries} \label{sec:dp-settings}

The type of DP guarantee necessary (as well as its strength) should depend on the threat model(s) of concern. The more plausible it is that some (raw, intermediate, or final) piece of data is visible to a potentially adversarial actor, the stronger the DP anonymization requirements should be. Depending on the threat model, a (potentially adversarial) actor could have access to different components of an ML system or workflow:
\begin{compactenum}[B1.]
  \item The raw data from which training examples are derived. In this work we presume this raw data contains privacy-sensitive information. \label{item:rawdata}
  \item The training dataset itself. Access to the training data might be a concern when for example releasing a dataset for use in an ML competition. Arguably the hardest setting, we briefly survey techniques for privatizing datasets in Section \ref{sec:dp-input}. \label{item:traindata}
  \item Gradients or updates from an individual user. This might be a concern if data is transmitted from devices without on-the-wire encryption, or if the adversary has access to the intermediate state of the aggregator / training algorithm.\label{item:localdpadversary}
  \item Intermediate models or aggregated gradients. This might be a concern in federated learning, where partially trained models are sent to client devices. \label{item:intermediary}    
  \item The final (production) model parameters. This might be a concern if model parameters are open-sourced or deployed for on-device inference. \label{item:finalmodel}
  \item Predictions made by the production model. This might be a concern if the models predictions are used in a public web service or app.  We cover options for directly protecting predictions in Section \ref{dp:prediction-level}. \label{item:predictions}
\end{compactenum}
For access at levels B\ref{item:rawdata} - B\ref{item:intermediary}, data minimization approaches to privacy can often provide the primary defense---for example using appropriate security and data access procedures to limit visibility to a small number of trusted system administrators or ML engineers.  Hence, data anonymization (and DP in particular) is most salient for B\ref{item:finalmodel} and  B\ref{item:predictions}, as these may necessarily be exposed to some threats during the intended use of the model.

Importantly, the invariance to post-processing guarantee of DP (Section \ref{sec:properties}) plays a critical role here, in that as long as the data passes through the DP mechanism before the potential threat, the DP guarantee applies.  For example, if a DP mechanism protects B\ref{item:intermediary}, then B\ref{item:finalmodel} and B\ref{item:predictions} benefit from the same guarantee. This is, in fact, the most common approach:  the majority of the methods we discuss in Section \ref{sec:dp-training} actually provide protection at the level of B\ref{item:intermediary}. For example, in the usual analysis of DP-SGD (Section \ref{sec:dpsgd}), formally the output of the DP mechanism is the full sequence of per-iteration noised gradients, \textit{even though DP-SGD is commonly used when access to the final model, as in B\ref{item:finalmodel}, is the primary concern}.

For the DP guarantee to be meaningful, one needs to establish trust or verify that the mechanism is implemented correctly, and that the raw data and pre-DP intermediate values are suitably protected. There are several ways that DP can be applied that take different approaches to these requirements. 

\paragraph{Central or trusted-aggregator DP.}
In this setting, a trusted service provider (often called the aggregator) has access to the raw data and is in charge of ``privatizing'' the model by applying DP. \textit{This is the setting we concentrate on throughout this paper.}

Users contributing data need to trust this aggregator, and the primary privacy concern relates to the output of the DP mechanism the aggregator implements, or something post-processed from that output (e.g., the final model will be made public as in B\ref{item:finalmodel}, and users contributing data want a DP guarantee to ensure that this final model cannot be used to learn something private from the training data contributed). That is, any adversary is assumed to only have access to the released output of the trusted aggregator. In the ML context, this corresponds to a setting in which many users contribute their raw data to a dataset which is typically stored centrally by the aggregator, and used to train a model which is eventually released. In some settings (e.g., federated learning), intermediate versions of the model are also released. 

\paragraph{Local DP.}
{\em Local differential privacy} is an alternative setting motivated by cases where users contributing their data do not fully trust the central aggregator (e.g., are concerned about data breaches or insider threats at the entity coordinating training).  We discuss applications of Local DP to learning in Section \ref{sec:dp-input}.
Formally, Local DP \cite{kasiviswanathan2011can} is defined as follows:
\begin{definition}[Local Differential Privacy]\label{def:local_dp}
Let $\epsilon$ be a positive scalar. A mechanism  $\mathcal{A}$ guarantees {\em $\epsilon$-local differential privacy} if for any two values $x$ and $x'$, and for any $S \subseteq \text{Range}(\mathcal{A})$,
\begin{align}
    P[\mathcal{A}(x) \in S] \leq \exp(\epsilon) \times P[\mathcal{A}(x') \in S].
\end{align}
\end{definition}

In this setting, an adversary can see the output of a transformation on any individual's record before any aggregation (as in Item B\ref{item:localdpadversary} above), and must still not be able to distinguish anything about that individual regardless. The requirement of local differential privacy is much stronger than that of central differential privacy, as it requires an algorithm to give indistinguishable output on any possible pair of data points, no matter how distinct. Often, this results in a much more substantial drop in utility compared to central DP for the same problem.

\paragraph{Distributed DP\Difficult.}
Distributed DP seeks to recover the utility of central DP without having to rely on a fully trusted central aggregator \cite{ODO,bittau17prochlo,kairouz2021distributed,agarwal2021skellam}. These techniques are essentially based on running the core of the DP mechanism (typically aggregating and noising) in a `secured box' that even the organization administrating the mechanism cannot look into, thus rendering the output differentially private before it becomes visible to the aggregator. 
Currently, such approaches are most feasible in the federated learning setting, where a collection of \emph{clients} (mobile devices or even different organizations) holds the raw data. In a typical setup, these clients compute minimal reports (e.g., gradients) as in local DP, and perturb these slightly with random noise. However, if for a given DP guarantee the local approach would require noise of magnitude 1 on each client, distributed DP typically would only require noise of magnitude $1/\sqrt{n}$ (where $n$ is the number of clients in the aggregation) \cite{kairouz2021distributed}.  The server then has access only to the output of the private aggregation protocol. The noise added by individual clients is typically insufficient for a meaningful local DP guarantee on its own. After private aggregation, however, the output of the private aggregation protocol provides a stronger DP guarantee based on the total sum of noise added across all clients. This applies even to someone with access to the server under the security assumptions necessary for the private aggregation protocol, which could be provided cryptographically, e.g. via Secure Aggregation \cite{bonawitz2017practical}, or via hardware trusted execution environments (TEEs).

In contrasting Local and Distributed DP, it is worth remarking that one can view Local DP as using a data anonymization approach where arguably a data minimization approach should be preferred. That is, in the ML context, there is no need to release the contributions of individual users/clients (e.g. gradients), as their only use is to be aggregated into a final batch gradient and eventually into a final model. Thus, using cryptographic protocols or Trusted Execution Environments (TEEs) to simply remove access to the non-aggregated non-privatized values entirely (as in Distributed DP) is likely preferable to noising them and assuming they can be accessed by an adversary as in Local DP, assuming the security properties of the TEEs or protocols are sufficiently strong. 
\\
\begin{specialistbox}{Choosing an appropriate DP approach}
The setting chosen determines the set of privacy threats that can possibly be addressed by a DP guarantee. We focus on the \textbf{Central DP setting}, where the entity training the model is considered trusted and has access to the raw data. However, this setting can be insufficient if e.g. insider threats or data breaches are a primary concern (as these might bypass the DP outputs entirely). Local DP is an intuitive approach for decentralized data, but typically suffers from severe utility loss, and where feasible Distributed DP methods are likely to be preferred as they can offer protection against similar threat models with much higher utility.
\end{specialistbox}

\subsection{Where to Apply DP}\label{sec:wheretoapply}
We now examine various ways of adding differential privacy to machine learning  workflows. Keeping in mind the post-processing property of DP, we have a choice of enforcing differential privacy in three different phases of the typical ML pipeline:
\begin{compactenum}
    \item \textbf{Adding DP at Input/Data level}: If the input data is made differentially private, any model trained using that data will also be differentially private, as will be all outputs of that model. This is the most challenging place to introduce DP, but in Section \ref{sec:dp-input} we explore several methods that make progress in this direction.
    \item \textbf{Adding DP during ML model training process}: This is by far the most common approach to obtain DP ML models. Even if the input data is sensitive, if the model training algorithm is differentially private, then the resulting model and its outputs will be differentially private. Here one can distinguish between
    \begin{compactenum}
    \item \textit{Label only protection}. In this setup, only the labels of the training data are considered private, while features are treated as public. We explore methods for DP-Label protection in Section \ref{sec:label-protection}
    \item \textit{Full training data protection}. In this setup, which is probably the most common, both features and labels of the data are considered private and need to be protected. Section \ref{sec:dp-training} describes methods to achieve such protection.
    \end{compactenum}
    Gradient perturbation methods, which we will cover in Section \ref{sec:dp-training} are the most common and practical methods for DP-Training. They work by making the gradients differentially private. As such, by postprocessing property, weights are also DP, so all the checkpoints and the final model weights are DP and can be released \footnote{Assuming the privacy guarantee is deemed sufficient for the application.}.
    \item \textbf{Adding DP to the predictions of an ML model} is possible when the model itself does not need to be released. This level of DP protection is appropriate if users are only allowed to access model predictions through some trusted server by providing their own inputs. 
\end{compactenum}

\begin{specialistbox}{Where DP modifications are introduced.}
As a general rule, \textit{the task of introducing DP becomes ``easier'' the further from the data DP modifications are introduced}, with the hardest being DP at the input level and the easiest (resulting in the smallest hit to the ML model utility) being at the model prediction level (assuming only a limited number of predictions are made). 
\end{specialistbox}

These methods clearly come with different levels of guarantees. Table \ref{tab:dp-vectors} summarizes the interaction between required mode of access to the model (which is dictated by the threat of concern) and where the DP-related modifications can be introduced. 
While the majority of this work will focus on \textit{DP-Training for the full model protection} (Section~\ref{sec:dp-training}), we provide a brief overview of other aforementioned methods below.

\setlength{\tabcolsep}{5pt}
\begin{table}[!ht]
\small
\begin{tabular}{lcccl}
\toprule
\multirow{2}{*}{\textbf{Mode of access}} &
  \multicolumn{3}{c}{\textbf{What is considered "public"}} &
  \multirow{2}{*}{\textbf{Where DP is added}} \\
                              & \textit{Training data} & \textit{Model weights} & \textit{Model predictions} &                   \\
\midrule
Model predictions     &                        &                                        & x\tablefootnote{Inference budget must be set, access to only a limited number of predictions is allowed}           & Predictions       \\
Model weights &                        & x                                     & x~                             & Training process\tablefootnote{Assuming gradient perturbation methods that make gradients DP.}  \\
Access to data     & x                      & x                                     & x~               
& At the data level \\
\hline
\end{tabular}
\caption{\small The connection between where the DP mechanism is introduced, the mode of access to an ML model, what can be released freely (e.g., considered public). Note that all methods aim to protect \textit{the original training data}, however the mode of access determines how broad of interface is revealed to the public to query the result of a DP-algorithm.}
\label{tab:dp-vectors}
\end{table}

\subsubsection{DP at the Input Level}\label{sec:dp-input}
Input arguably is the most challenging place to apply DP. Intuitively it is the case due to the fact that this option gives the broadest privacy coverage: releasing a differentially private version of the dataset allows the use of an arbitrary training algorithm, but also must ensure privacy for any possible use of the anonymized data, including the inspection of individual training examples.

\paragraph{Local DP approaches.}
When the dataset is formed by collecting anonymized examples from users (without ever collecting the non-privatized data), a local DP guarantee is possible.
The resulting anonymized dataset can be then passed to an arbitrary training algorithm. However, the noise of the DP mechanism introduced in such setting will typically be far too large.
Achieving this requirement has proven difficult enough that researchers have tended towards using relaxations of local DP instead.  

One such relaxation is the idea of ``limited-precision local privacy'' (LPLP), introduced by~\cite{pmlr-v97-schein19a} in the context of algorithms for analyzing count data. Essentially, LPLP modifies the definition of local differential privacy by only requiring it to apply when the two elements in question fall below a given distance threshold. The authors then devise a new LPLP algorithm for the problem of Bayesian inference for Poisson factorization.

A more continuous relaxation called d$\chi$-privacy was proposed by~\citet{CABP_dchi}. In this setup, an adversary's success probability is allowed to depend on a context-specific distance between the two elements under consideration. (Formally, this is done by replacing the $\epsilon$ in Definition~\ref{def:local_dp} with $\epsilon d(x, x')$, where $d(x, x')$ is the (problem-specific) distance between $x$ and $x'$);  this notion has been successfully used in a number of works on text data.

Recently \citet{FBDD2020} devised a d$\chi$ mechanism to modify a text string by taking each word, adding noise to its embedding, and then replacing the original word with the word closest to the noisy embedding. They prove that this mechanism satisfies d$\chi$ privacy, and present experiments showing that the output is still useful for text analysis models.

\citet{FDM2019} define a measure they call ``earth mover's privacy'' based on the idea of d$\chi$ privacy, in which the distance between examples is the well-known earth mover metric. Based on this new relaxation, authors describe an algorithm for adding noise to a bag-of-words representation of a text document and demonstrate its effectiveness in obscuring author identity.

The application of d$\chi$ privacy to language models was further systematically explored by \citet{QKYZBN2021}. They experiment with three different d$\chi$-privacy techniques for privatizing text data at the token, text, and sequence level, and explore the effects of these method when fine-tuning pretrained BERT models in a variety of settings. They also explore the idea of using privacy-adaptive pretraining.

\paragraph{Synthetic private data generation.}
Another broad category of approaches for adding DP at the input level is \textit{synthetic data generation}, which is generally done in the central DP setting. This line of work moves away from adding noise to individual examples in a dataset and instead seeks to generate fully private synthetic examples that can be freely shared. In order to generate such synthetic data, some sort of probabilistic model describing the underlying distribution is created and then subsequently sampled, and the fidelity of this model is crucial for the utility of the underlying synthetic data. 

Synthetic data for \textit{query release} was explored in depth extensively \cite{DBLP:journals/corr/abs-1109-2229}, \cite{DBLP:journals/corr/abs-1012-4763}. \textit{Differentially private query release} is a task of releasing accurate answers to a number of statistical queries (e.g., counts, sums etc). While answering a small number of such queries can be done by adding noise perturbation to the query results, for a large number of queries, approaches that generate synthetic data and subsequently answer the queries using this data have been quite prominent in the literature. The utility of this type of synthetic data is measured by the quality of the answers to statistical queries. For this setting, \citet{DBLP:journals/corr/abs-1012-4763} introduced MWEM-mechanism based on private multiplicative weights, \citet{DBLP:journals/corr/abs-1202-3807}'s mechanism is based on Matrix Mechanisms; Dual Query by \citet{balle2018improving} algorithm views the synthetic data generation setup as a zero-sum game. \cite{DBLP:conf/icml/VietriTBSW20} introduced 3 algorithms (FTPL, FEM, sepFEM) algorithms that rely on black-box optimization routines. RAP \cite{DBLP:journals/corr/abs-2103-06641} follows the select-measure-generate paradigm that generates synthetic data to closely match noised answers to chosen queries.  \citet{DBLP:journals/corr/abs-2106-07153} unifyied these approaches under a common paradigm.  \citet{DBLP:journals/corr/abs-2201-12677}'s AIM used the same select-measure-generate paradigm, with the modification to select stage where authors iteratively and greedily select the most useful queries that took into account the value of these queries for approximating the original data.  \citet{DBLP:journals/corr/abs-2108-04978} then applied similar paradigm for generation of synthetic data that can be released on its own (as opposed query release task where synthetic data is not released but used to release the answers to statistical queries).

 The topic of generating private \textit{tabular} synthetic data that would be useful for an \textit{ML model} training has recently gained popularity \cite{DBLP:journals/corr/abs-2112-09238}. In this setting, synthetic data utility is evaluated based on the performance of an ML model trained on synthetic data. 
 One common strategy for high-dimensional synthetic data is to generate a set of low-dimensional marginals over the input data, and use them to approximate the underlying distribution. For example, \citet{10.1145/2588555.2588573} proposed a PrivBayes method that constructs a Bayesian network to model the interactions of the features. The noise is then injected into the marginals to ensure differential privacy. Synthetic dataset is constructed thereafter by sampling from this approximate DP distribution. JunctionTree method \cite{junctiontree} subsequently improved upon PrivBayes by learning DP-protected pairwise correlations of the attributes and applying junction tree algorithm to infer joint data distribution via noisy marginals. More recently, \citet{10.14778/3476249.3476272} introduced PrivMRF which uses Markov Random Field model to represent the correlations among the data attributes. 
 Another line of research foregoes directly modeling the  marginals and adopts ML approach to learning the underlying data distribution automatically. For example, GAN-based methods were explored in DP-GAN \cite{DBLP:journals/corr/abs-1802-06739} and PATE-GAN \cite{yoon2018pategan}, and the ensembling approach was recently proposed in \cite{DBLP:journals/corr/abs-2106-07153}. However a recent benchmark finds that 
for tabular data the aforementioned Marginal-based methods seem to outperform GAN-based in terms of final ML model utility \cite{DBLP:journals/corr/abs-2112-09238}. 

The field of private synthetic data in the context of \textit{complex data such as text, images, or audio/video} is still very much nascent.

\subsubsection{DP at the Prediction Level: Privacy Preserving Predictions}\label{dp:prediction-level}
Adding DP at the prediction level is used in a setting when a trained ML model is accessible only through a secure interface \cite{DBLP:journals/corr/abs-1803-10266}. In particular, a (potentially adversarial) user has only the ability to obtain predictions from the model \textit{using its own data\footnote{If the model is queried on private data at inference time, in general the model output will still be privacy sensitive (for example, consider a model that simply changes the style of an input image but leaves it semantically unchanged).}}. Such an access mode is popular for pay-per-use models like various cloud-based ML prediction APIs. Importantly, the goal is still to protect the privacy of the training data used to train the model(s) making the predictions, as with all the approaches we consider. Private prediction methods introduced at prediction level come with an \textit{inference budget}, which limits the number of predictions a user can access \cite{DBLP:journals/corr/abs-2007-05089}. 

Techniques based on Sample-and-aggregate framework \cite{sampleaggregate} are commonly used in order to allow to answer multiple user queries without privacy degradation. They work by splitting the training data into a number of disjoint subsets, training an ML model for each of these subsets and then during the prediction time aggregating the predictions from these non-private models while taking into account the level of consensus from these models (adding less noise with similar predictions and more noise otherwise).  

Sample-and-aggregate is the workhorse of privacy preserving prediction methods, with the differences being how the aggregation happens and the amount of noise that needs to be added during such an aggregation. Additionally, \citet{DBLP:journals/corr/abs-1803-10266} introduced non-aggregation based approaches that instead rely on subsampling and uniform stability.

Finally, it is worth mentioning that empirical study by \citet{DBLP:journals/corr/abs-2007-05089}  argues that introducing DP during \textit{the training process} (instead of at the prediction level), provides a better privacy-accuracy-tradeoff for private predictions in many cases like when large inference budget is required or large amount of training data is available (while also providing stronger access guarantees, refer back to Table \ref{tab:dp-vectors}).

\subsubsection{DP During The Training Process: Protecting Only Labels (Label-DP)}\label{sec:label-protection}
In general, modification of the training process can result in either full model protection (that we will explore in detail in Section \ref{sec:dp-training}) or label-only protection. We explore the latter setting in this section.

Label-level DP is a relaxation of ($\epsilon, \delta)$ DP for ML models. This definition considers only the labels of the data to be sensitive. This is in contrast to both labels and features being treated as private/sensitive, as in standard DP. 
An example of such setting is online advertisement where models predicting either conversion on clicks are trained. For such models, the data about the advertisement (e.g., link, product etc) is considered public, whereas the label (whether a user clicked or converted on an ad) is private and should be protected \cite{https://doi.org/10.48550/arxiv.2202.12968}

\begin{definition}[Label Differential Privacy \cite{pmlr-v19-chaudhuri11a}]\label{def:label_dp}
Let $\epsilon, \delta$ be a positive scalars. A mechanism  $\mathcal{A}$ guarantees {\em label $(\epsilon,\delta)$ differential privacy} if for any two datasets $D$ and $D'$ that differ only in the label of one instance, and for any $S \subseteq \text{Range}(\mathcal{A})$,
\begin{align}
    P[\mathcal{A}(D) \in S] \leq \exp(\epsilon) \times P[\mathcal{A}(D') \in S] + \delta.
\end{align}
\end{definition} 
\noindent This definition is the same as the classical $(\epsilon,\delta)$ definition with a notion of what makes two datasets $D$ and $D'$ neighboring modified.

\citet{DBLP:journals/corr/abs-2102-06062} show that Label-DP is significantly ``easier'' than providing full level protection (e.g., protecting both features and the labels), therefore achieving small performance drop due to DP should be possible with small $\epsilon$ values.

There are a number of ways to achieve Label DP protection. The first is using classical randomized response (RR) \cite{10.2307/2283137} -- by randomly flipping the training labels using some predefined probability before labels are used for training/model updates. For example, for models trained with SGD, labels are randomly flipped before gradient is calculated.

While early works use pre-determined prior distribution for labels' change, recent work by \citet{DBLP:journals/corr/abs-2102-06062} proposed to instead learn a prior by bootstrapping it from a uniform prior and progressively updating this prior during multi-stage training. This works by splitting the training data into a multiple subsets and training a model on each subset. The top-k previous model predictions are used as a new prior for the model trained on the next subset.

Two additional methods were introduced by \citet{esmaeili2021antipodes}. Their first method called PATE-FM works by first splitting the training data into $K$ disjoint subsets, and then training a teacher model on each of these labeled subsets while also incorporating unlabeled data from all other subsets. Then a student is trained using the votes from all the teachers as the labels for the data. The second method proposed by \citet{esmaeili2021antipodes} is ALIBI, based on Randomized Response, that perturbs the one-hot encoding of training labels with Laplace noise before training, making the labels soft (as opposed to hard label switching as in RR). After additional normalization using Bayesian Inference to make label distribution probabilities per instance sum to one, a model is trained conventionally (e.g., via back-propagation).

More recently, \citet{pmlr-v151-esfandiari22a} proposed another solution for achieving Label DP using clustering. In their method, training data points are clustered using their features, then labels are randomly re-sampled using the labels of other examples in the cluster, producing a new training data with noisy labels. Subsequently, a model is trained with this new training data and a modified loss. Authors show that such approach significantly improves privacy-utility trade-off compared to direct application of RR to the labels.

\section{DP-Training: Protecting Full Training Data }\label{sec:dp-training}

The term ``\textit{DP training}'' often refers to a modification of the training process of ML models which guarantees that the resulting ML models are differentially private. In this section when we talk about DP-Training we aim to provide full model guarantees that state that the model would not be sufficiently different, no matter whether a particular instance was or was not included in the training data.

In this section we first provide a literature overview of DP-Training methods (Section \ref{sec:dptraining-survey}). 
We then proceed to introduce one of the most popular algorithms for DP-Training -- DP-SGD (Section \ref{sec:dpsgd}) and discuss advanced topics on DP-SGD convergence (Section \ref{sec:dpsgd-convergence}) and privacy accounting (Section \ref{sec:dpaccounting} and \ref{sec:amplification}). 
We additionally explore DP-Training algorithms that provide user-level privacy (as opposed to example-level) guarantees (Section \ref{sec:dp-user-modifications}). 
Finally, we conclude with a discussion on what makes the adoption of DP-SGD hard in practice (Section \ref{sec:problems-dp-sgd})

\subsection{Survey of DP-Training Methods}\label{sec:dptraining-survey}
Broadly speaking, DP Training can be categorized into noise injection methods and alternate methods. Noise injection methods can be further categorized by where in the training process the noise is introduced. While the most common method for deep learning ML models is gradient noise injection, below we survey most of the approaches that were explored in academia.

\subsubsection{Trained Weights Noise Injection Methods}\label{sec:noise-injection} 
These methods modify the \textit{already trained model weights} and are also sometimes referred to as an \textit{output perturbation methods} \cite{10.5555/3361338.3361469}.
This is one of the first lines of work that stems directly from Dwork's definition of privacy and an introduction of randomized  mechanisms. These methods work by injecting the noise proportional to the sensitivity of the training output (which describe how much the weights can change on neighboring datasets) \cite{NIPS2008_8065d07d}. The analysis to bound such sensitivity can be performed only for relatively simple models like linear regression, due to complex dependencies between the data and the weights \cite{DBLP:journals/corr/abs-1208-0219}. For example, \citet{NIPS2008_8065d07d} introduce an algorithm to inject the noise into trained logistic regression models.

More recently \citet{DBLP:journals/corr/0001KCJN16} proposed a ``bolt-on'' approach to obtaining DP models that were trained with SGD. They utilized output perturbation and injected the noise at the end of the training. In order to provide privacy guarantees, authors conducted analysis of $L_2$ sensitivity of SGD for convex and strongly convex losses.

\subsubsection{Objective/Loss Modification Methods} 
These types of methods also assume (some form) of convexity of the loss. 
Loss modification methods work by perturbing the loss function with noise, which is subsequently optimized normally using SGD or other optimizers \cite{JMLR:v12:chaudhuri11a,kifer2012private,phan-1}. For example, 
\citet{NIPS2008_8065d07d} introduced a modified regression loss, similar to logistic regression, for achieving privacy for convex and twice differentiable loss functions. They demonstrated that for logistic regression such loss perturbation requires less noise than weights noise injection (for situations when small regularization coefficient is used).
\citet{DBLP:journals/corr/abs-1208-0219} introduced the 
Functional Mechanism to both linear and logistic regression and showed that the noise magnitude is constant with respect to the dimensionality of the training data. This mechanism works for functions that can be represented or approximated as finite polynomials, expanding the loss as a polynomial of its weights and adding Laplace noise into the coefficients. For functions that cannot be represented using finite-degree polynomials, a truncated Taylor's expansion can be used as an approximation. For example for the Logistic Regression loss function, authors use a second-order Taylor expansion. While this expansion allows to bound the sensitivity, it limits the expressive power of the models. \citet{phan-1} extended this work to deep auto-encoders, where the authors approximated both the data reconstruction and standard losses with a Taylor expansion.

It is important to point out that for privacy guarantees to hold, both types of aforementioned noise injection (noising weights and objective perturbation) require strong convexity assumptions. Further, they require convergence to a global optimum. At the cost of these strong assumptions, these specialized methods achieve truly impressive privacy guarantees (with $\epsilon$ at most one ) with only slight degradation of utility. Several works attempted to remove the convergence to the global optimum requirement, for example \cite{8835258} presented an alternative loss perturbation where privacy guarantees hold if the model reaches  the vicinity of a global optimum, however the convexity remains a requirement. Later \citet{DBLP:journals/corr/abs-1909-01783} attempted to relax the convexity assumption. In particular, they introduced an algorithm that required boundedness of the loss function and Lipschitz continuity in the model weights for its privacy guarantee. However, the utility (accuracy) bounds of this algorithm still required convexity and boundedness of the loss assumption. The algorithm works by solving polynomially many problems with perturbed losses, each with an independently introduced random perturbation. Then the average of these models with an addition of Laplace noise is employed. Therefore it is computationally expensive and feasible only for relatively small datasets (the authors report success on a dataset of 15k instances with 23 dimensions).

Most deep ML models both have non-convex losses and are not trained to the global optimality, due to time constraints and the difficulty of the problems. Additionally, they are often trained on a huge corpus of data, and thus training even one copy of the model is already expensive. Achieving such strong privacy guarantees as those for simpler models ($\epsilon \le 1$) is usually impossible without a severe degradation of utility\footnote{Please refer to Section \ref{sec:good-eps} for more in-depth discussion of what privacy guarantees can be achieved for complex models like deep neural nets.}.
Further, almost all work on DP Training for more complicated models uses the approximate  ($\epsilon$, $\delta$) notion of DP  \cite{10.5555/3361338.3361469}, with the most popular class of methods that is applicable to any generic differentiable ML model being gradient noise injection.

\subsubsection{Gradient Noise Injection Techniques} 
These methods are applicable to any ML model that is optimized using a gradient-based method, explaining their popularity. They work by introducing the noise into the gradients before the gradient step update is applied to the weights. In order to provide privacy guarantees, these methods require a bound on gradient norm, which is hard to provide for deep learning models because oftentimes gradients are unbounded or have bounds that are difficult to compute. The standard way to address this requirement is to clip the gradient so it's norm is less than a specified value. 

The most commonly used algorithms for gradient noise injection  are variants of SGD. \citet{6736861} introduced differentially private stochastic gradient descent (DP-SGD). Their algorithm was modifying SGD updates with the noise for linear classification problems. Due to the nature of this loss, they were able to bound the sensitivity of SGD updates without gradient clipping and used strong composition to achieve the final bound. \citet{DBLP:journals/corr/BassilyST14} refined the analysis for privacy budget by taking into account the randomness of the batch sampling (a.k.a. privacy amplification by sampling~\cite{kasiviswanathan2011can}, which allowed them to run the algorithm for more steps without a significant cost to privacy. Additionally,~\cite{DBLP:journals/corr/BassilyST14} demonstrated that DP-SGD is optimal for DP convex optimization under $(\epsilon,\delta)$-DP. 
\citet{Abadi_2016} refined the DP-SGD method to use it for training deep learning models, and presented a tighter privacy accounting (a.k.a. moments accountant), that gained a lot of popularity and remains the standard method implemented in many DP-Training libraries. 
We introduce this algorithm in detail in Section \ref{sec:dpsgd}.

Briefly, this method works by introducing two simple modifications to the standard SGD algorithm. First, the per-example gradients are clipped to some maximum predefined norm. Second, Gaussian noise is added to the average of the per-example gradients, and  the resulting noised gradient is used to perform the gradient update. 
Different privacy accounting techniques can be used to accumulate the total privacy cost incurred by using the Gaussian mechanism on each step. Critical to all of these analysis techniques is the concept of \textit{privacy-amplification-via-sampling}, discussed in depth in Section \ref{sec:amplification}. This concept requires that data processing samples either fixed or variable-sized batches of examples \emph{with replacement} on each iteration. 
The final $\epsilon$ guarantees depend on the noise level, the total number of steps (batches) used for training and the  sampling ratio (ratio of batch size to the total dataset size). The utility of models trained with DP-SGD depends heavily on the choice of hyperparameters. We discuss related work on hyperparameter tuning and provide an algorithm for tuning the related parameters in Section \ref{sec:tuning_hyperparams}. Additionally, we discuss a broad body of research on  challenges faced by DP-SGD and some proposed solutions in Section \ref{sec:problems-dp-sgd}.

In contrast to DP-SGD techniques, which use an independent application of the Gaussian mechanism to release a private estimate of the per-iteration gradient on each round, DP-FTRL algorithms \cite{kairouz21practical} use a \emph{stateful} DP mechanism which observes the (true) gradient on each iteration, and then releases a privatized estimate of the \emph{cumulative sum of gradients so far}. These prefix sums are sufficient to implement the SGD algorithm, and in fact DP-FTRL combined with a matrix-factorization mechanism can directly privatize the iterates of SGD with momentum and a learning-rate schedule \cite{denisov22matfact}.  The stateful nature of the DP mechanisms used in DP-FTRL is critical, as this allows the mechanism to ``hide'' the release of information about gradient $g_{t}$ over all rounds $t' \ge t$. By taking differences of the gradient prefix sums, one can equivalently view these stateful mechanisms as release DP estimates of individual gradients $g_t$ with (anti)correlated noise, so that the noise in the estimate of $g_{t}$ cancels out some of the noise introduced in the private estimates of previous gradients. These capabilities allow DP-FTRL to provide strong privacy guarantees \emph{without assuming any random sampling} --- it is sufficient to process the data in an arbitrary order so long as each example occurs a bounded number of times. This is particularly useful in the federated learning setting where random sampling is generally infeasible. 
However, even for centralized training this shuffled rather than sampled data processing pattern may better fit common ML infrastructure (see Section \ref{sec:amplification} for further discussion). Further, \cite{choquette22multiepoch}  showed that Matrix Factorization DP-FTRL (MF-DP-FTRL) can outperform DP-SGD in some settings for small values of $\epsilon$, often substantially. More recently, by introducing banded matrices, \cite{choquette23amplified} suggested that MF-DP-FTRL can subsume prior state-of-the-art algorithms in both federated and centralized training settings.

\begin{specialistbox}{Practical methods for DP-Training}
Gradient perturbation-based methods are so far the most practical methods for achieving rigorous privacy guarnatees for non-convex problems like large scale deep neural nets. Even for small scale models with strongly convex losses, where alternative techniques like output and loss perturbation methods are applicable, well tuned implementations of noisy gradient descent has been shown to result in better utility \cite{DBLP:journals/corr/abs-1911-11363}
\end{specialistbox}

\subsubsection{Alternative Methods for DP Training:} 
While the Sample And Aggregate framework (Section \ref{dp:prediction-level}) is commonly used for Prediction-level protection, \citet{pate} introduced an extension that provides full data protection \textit{assuming availability of the public data of similar distribution to the data that is being protected}. In particular, authors
 introduced Private Aggregation of Teacher Ensembles (PATE) method that is applicable to any multi-class classification model, including non differentiable models \cite{pate}.
 
 The idea behind PATE is to utilize the algorithm used for private prediction and 
to create disjoint subsets of the training data and then train a separate teacher model on each of these subsets. The private student model then is trained using non-sensitive (public) unlabeled data and voted labels from the teachers. To strengthen the privacy guarantees, only a limited number of teacher votes is used and Laplace noise is added before the top vote is chosen. A clear upside of this approach is that it is intuitively understandable by a non DP expert, at the same time providing rigorous DP guarantees.  
While the student can be trained with distillation, authors state that the most successful way of training the student is using GAN-like approach for semi-supervised training. This involves a discriminator and a generator that are co-trained together, where
the generator produces the samples by modifying the Gaussian data, and the the multi class discriminator attempts to classify real samples into their real correct classes, generated samples  as an additional "fake" class and unlabeled real samples into any of the real classes. The labels are obtained from the teachers via the aforementioned voting. Authors report the improvements in utility over DP-SGD method by \citet{Abadi_2016} on MNIST and SVHN data and improved privacy guarantees (from $\epsilon=8$ to $\epsilon=1.9$). While the method was introduced for classification tasks only, it can be extended to the regression tasks by allowing the teachers to produce a regression estimate and modifying the analysis to account for the variance in the predictions of such votes.
The downsides of this approach is that it is computationally expensive, requiring training of many teacher models, each of which has its own hyperparemeters. For example, in \citet{pate}'s experiments, 250 teacher models were trained. With modern giant models like language models taking days to train, this approach is prohibitively expensive. Further, having more teacher models should improve the privacy but it limits the amount of data available for each teacher, so this method is also hard to apply to small datasets. Additionally, this method assumes access to unlabeled public data of similar distribution. Finally, it is not clear whether this voting approach can be extended to generative models like decoders.  

A more recent extension of Sample and Aggregate framework is presented in \cite{NEURIPS2018_aa97d584}. Similar to PATE, this work assumes availability of unlabeled public data and trains private teacher ensemble using disjoint subsets of training data. Using novel analysis techniques (subsample stability and sparse-vector techniques), authors were able to provide actual formal guarantees on the resulting private student model utility. They also demonstrated that their proposed framework could be used to achieve Label-only protection. Several other modifications of the teacher-student idea were proposed recently, often referred to as ``mimic learning'' \cite{BOULEMTAFES202021}

Please refer to \citet{10.5555/3361338.3361469} for an additional in-depth overview and comparison of various methods of DP Training.

For the remainder of this paper, unless specified otherwise, \textbf{we will use the term DP-Training interchangeably with DP-SGD}, since DP-SGD is one of the most popular methods for DP-Training. At the same time, DP-SGD and other \textit{gradient noise injection} methods cannot be used with discrete/not end-to-end trainable ML models and only methods of sample-and-aggregate framework \cite{sampleaggregate}(e.g., PATE) can be applied out of the box. Such models are not the focus of this paper, due to the fact that most recent/modern models seem to be based on giant neural networks. Nevertheless, we provide  some discussion on DP-Training for non-differentiable models in Appendix \ref{app:nondif}.

\subsection{DP-SGD Algorithm}\label{sec:dpsgd}
First-order methods with gradient noise injection, such as SGD proposed by \citet{Abadi_2016},  are the workhorse of many privacy libraries and the most common way of adding Differential Privacy to differentiable models. Algorithm \ref{algo:abadi} outlines two modifications introduced to standard SGD which make it $(\epsilon, \delta)$ differentially private \cite{Abadi_2016}. The first step is clipping of the per-example gradients to have a maximum norm of $C$, which bounds the influence of each example on the gradient. Note that this step happens before averaging of the gradients and it works on each individual \textit{per-example} gradient. The noise is then added to the aggregated gradient of the batch before this gradient update is applied to the model parameters. Having this noise proportional to the clipping norm ensures that impact of each individual (clipped) example is properly masked.

\begin{algorithm}
\caption{DP-SGD algorithm}\label{algo:abadi}
\begin{algorithmic}
\Require Training data, consisting of features  $X := \{x_1, x_2, ..., x_N\}$ and labels $Y := \{y_1, y_2, ..., y_N\}$. \\
$f(x; \theta)$ is the model applied to an input $x$ and parameterized by $\theta$. \\
$L(y, y')$ is the loss function for label $y$ and prediction $y'$.\\
SGD hyperparameters: $\eta$ learning rate, $T$ number of iterations, $B$ batch size. \\
DP hyperparameters: $C$ clipping norm, $\sigma$ noise level, $\delta$ (used only for privacy accounting).
\Ensure $\theta_T$ final model parameters
\State $\theta_0 \gets $ randomly initialized values
\For{$t \gets 1$ to $T$}                     
     \State Randomly sample a batch $B_t$ with sampling probability $B/N$ for each data point. 
     \State Data are sampled with replacement for each batch.
     
     \For{$i \in B_t$}  
        \State$g_t(x_i) \gets \nabla_{\theta_t}{L(y_i, f(x_i; \theta_t))}$
        \Comment{Compute per-example gradient wrt the weights }
        
         \State  \HiLi $g_t(x_i) \gets g_t(x_i)/\max(1, \frac{||g_t(x_i)||_2}{C})$
        \Comment Clip the per-example gradient
        
     \EndFor
     
     \State  \HiLiLong $\bar g_t  \gets \frac{1}{B}(\sum_i{g_t(x_i)}+\mathcal{N}(0,\,\sigma^{2} C^2 \identity ) )$
     \Comment Add noise

    \State $\theta_{t+1} \gets \theta_t - \eta \bar g_t $
    \Comment Gradient descent step
     
\EndFor

\end{algorithmic}
\end{algorithm}
Similar modification of clipping and noise can be used to easily obtain private versions of other optimizers, such as Adam, Adagrad, etc.  \cite{DBLP:journals/corr/abs-1812-06210}.

There are several caveats that should be highlighted. Firstly, many auto-differentiation libraries do not provide easy access to the per-example gradients required for clipping. The computation of the (accumulated over the batch) gradients required for a weight update can be represented as a sequence of matrix multiplications and element wise products, and some of these steps can be performed in parallel. Many auto-differentiation frameworks take advantage of vectorization for such matrix operations \cite{DBLP:journals/corr/abs-2009-03106}. 
In order to compute per-example gradients, some implementations, like TensorFlow Privacy, compute gradients for each example one at a time (as if the batch contained one example only). This results in a loss of GPU/TPU parallelism and forgoes GPU bulk data transfer benefits \cite{DBLP:journals/corr/abs-2009-03106}. In general, computing per-example gradients remains the slowest part of DP-SGD. 

Another caveat is the noise added to the accumulated gradient. 
The most common formulation (as in Algorithm \ref{algo:abadi}) is to decouple the parameters $\sigma$ and $C$, and report them separately. The noise added at each step will thus be sampled from $\mathcal{N}(0,\,\sigma^{2} C^2 \identity )$. This formulation allows to reason about the noise as essentially a percentage of the maximum gradient norm. This, in turn, allows choices of the noise level to be somewhat transferable between datasets and models, since the optimal clipping norm is data and architecture dependent. However, some works, e.g.,  \citet{9423193}, report the Gaussian noise level in a form of $\mathcal{N}(0,\,\sigma^{2})$, essentially making the $\sigma$ clipping-norm dependent. Since for calculating the privacy guarantees $C$ will not be used directly, care should be taken to make sure the calculations are performed with a decoupled variance that does not include the clipping norm.  

\subsubsection{Convergence of DP-SGD Variants\Difficult} \label{sec:dpsgd-convergence}

\newcommand{\calL}{\mathcal{L}}

Typically, the utility of DP-SGD (and other DP ML algorithms) can be measured in terms of \textit{excess empirical risk}, i.e., for a given model $\theta\in\mathbb{R}^p$ (weights) output by the algorithm, this error measure is defined as 
\begin{equation*}
    R_{\sf ERM}(\theta)=\calL(\theta;D)-\min_\theta\calL(\theta;D)
\end{equation*}
Here, $\calL(\theta;D)=\frac{1}{n}\sum\limits_{i=1}^n\ell(\theta;d_i)$ corresponds to the loss on the training data set $D$. 

Alternatively, one can measure the utility in terms of \textit{excess population risk} with respect to a fixed distribution $\tau$ as follows:
\begin{equation*}
R_{\sf Pop}(\theta)
= \mathbb{E}_{d\sim\tau} \left[\ell(\theta;d)\right] 
     - \min_\theta\mathbb{E}_{d\sim\tau}\left[\ell(\theta;d)\right] 
\end{equation*}
\citet{DBLP:journals/corr/BassilyST14,bassily2019private,bassily2020stability} show that for (strongly) convex and Lipschitz losses variants of DP-SGD obtain optimal excess empirical risk and excess population risk. For variants of DP-SGD one can output models $\theta_{\sf priv}$ that have excess empirical risk of $R_{\sf ERM}(\theta_{\sf priv})=\tilde{\mathcal{O}}\left(\frac{\sqrt{p}}{\epsilon n}\right)$ and $R_{\sf Pop}(\theta_{\sf priv})=\tilde{\mathcal{O}}\left(\frac{1}{\sqrt n}+\frac{\sqrt{p}}{\epsilon n}\right)$ for convex and Lipschitz loss functions, and \\ $R_{\sf ERM}(\theta_{\sf priv})=\tilde{\mathcal{O}}\left(\frac{p}{\epsilon^2 n^2}\right)$ and $R_{\sf Pop}(\theta_{\sf priv})=\tilde{\mathcal{O}}\left(\frac{1}{n}+\frac{\sqrt{p}}{\epsilon^2 n^2}\right)$ for strongly convex loss functions (where $p$ is the dimensionality of the model, e.g., number of weights and $n$ is the training dataset size.)\footnote{$\tilde{\mathcal{O}}$ refers to a variant of the big-O notation that ignores logarithmic factors.}

One can obtain better dependence on dimensions (and possibly dimension independence) for DP-SGD (Algorithm \ref{algo:abadi}) and some of its variants if the loss functions satisfy special properties (e.g., for generalized linear models~\cite{song2021evading}). For context, in the non-private setting one can obtain an excess empirical risk of zero, and excess population risk of $\tilde{\mathcal{O}}\left(\frac{1}{\sqrt{n}}\right)$ for convex losses, and $\tilde{\mathcal{O}}\left(\frac{1}{n}\right)$ for strongly convex losses~\cite{SSSS09}.

Unfortunately, in the non-convex setting convergence of DP-SGD to the optimal excess empirical risk or population risk is unknown, and the algorithm may generally diverge. However, one can obtain convergence to a stationary point\footnote{For a differentiable objective function, a (first-order) stationary point is one where the gradient is zero. For SGD and DP-SGD, convergence to a stationary point is commonly established in expectation, e.g., the expected norm of the gradient converges to zero. We  note that stronger notions such as second-order stationarity are also employed in some of the literature.} (both for the empirical loss and the population loss) under certain assumptions~\cite{wang2019differentially,chen2020understanding,song2021evading,bu2021convergence,das2022beyond,arora2022faster}. For example, convergence to a stationary point can be established if the distribution of the gradients encountered by DP-SGD is symmetric \cite{chen2020understanding} or heavy-tailed \cite{das2022beyond}. \citet{chen2020understanding} empirically demonstrates that CNNs trained with DP-SGD on MNIST and CIFAR have nearly symmetric gradient distribution, so these models are expected to converge to (approximate) stationary solutions.
While the notion of convergence to a stationary point is considered important in optimization theory, we note that it may not be as important from a practical perspective in deep learning. In fact, many popular, deep neural nets do not converge to stationary points but nevertheless achieve good performance and a stable training loss  \cite{pmlr-v162-zhang22q}.

\subsubsection{DP-SGD Privacy Guarantees: Theory \Difficult} \label{sec:dpaccounting}
As mentioned earlier, the final ($\epsilon_f$,$\delta_f$) privacy guarantees describe the overall privacy of the DP-Trained model. While $\delta_f$ is usually set to be less than the  inverse of the training data size, the final $\epsilon_f$ value can be calculated based on the level of noise, sampling ratio (defined as $\frac{\text{batch size}}{\text{total dataset size}}$) and the number of training iterations of DP-Training.

To calculate such $\epsilon_f$, one must be able to keep track of the evolution of the privacy loss for the mechanism. Privacy loss is defined as a random variable, and providing bound on its tail is equivalent to saying that the mechanisms is $(\epsilon_f, \delta_f)$-DP. More specifically, privacy loss of an outcome is defined as a log ratio of probabilities of an outcome on two neighbouring datasets. Obviously, the better accounting procedure that provides tighter bounds on the tail directly translates into better utility, since the noise magnitude required to obtain the same $\epsilon$ guarantees will be lower and/or DP-Training can run for more steps.

Examining DP-SGD Algorithm \ref{algo:abadi}, the Gaussian mechanism applied to a random batch of the data in each DP-SGD step achieves the same $(\mathcal{O}(q(e^{\epsilon_s}-1)), \mathcal{O}(q\delta))$\footnote{For $\epsilon \le 1$ it is often approximated as ($\mathcal{O}(q \epsilon_s), \mathcal{O}(q\delta)$).} guarantee using the Amplification theorem (please refer to Section \ref{sec:amplification} for a discussion), where $q$ is a sampling ratio and $\epsilon_s=\sqrt{2 \log{\frac{1.25}{\sigma}}}$. Direct application of composition will give a bound of the whole DP-Training procedure to be  $\mathcal{O}(q (e^{\epsilon_s}-1) T, q \delta_f T)$, while the strong composition theorem \cite{dwork}, which states that  $\epsilon$ parameter increases only with a square root of the number of steps in composition, can achieve the total privacy bound of $(\mathcal{O}(q (e^{\epsilon_s}-1) \sqrt{T \log{\frac{1}{\delta_f}}}), \mathcal{O}(q \delta_f T))$.

\citet{Abadi_2016} introduced a stronger accounting method called the moment accountant, which allowed them to forego the direct application of composition theorem, dealing instead with R\'enyi definition of Differential Privacy (RDP), which has tighter composition rules. Once the R\'enyi DP guarantees are obtained, they can be mapped back to the $(\epsilon, \delta)$ DP definition. Several works were able to improve upon these conversion rules, with \citet{DBLP:journals/corr/abs-2001-05990} providing an optimal conversion procedure from RDP to DP. 

Next we will briefly examine the R\'enyi DP and how it is used to bound the privacy loss. 

R\'enyi Differential Privacy (RDP) was introduced by \cite{DBLP:journals/corr/Mironov17} and is based on R\'enyi divergence:
\begin{definition}[R\'enyi divergence\label{def:renyi} \cite{DBLP:journals/corr/Mironov17}]
Let $P$ and $Q$ be two probability distributions defined over $R$. Then  R\'enyi divergence of order $\alpha > 1$ is defined as ,
\begin{align}
    D_\alpha(P||Q)=\frac{1}{\alpha-1}\log E_{x\sim Q} \left( \frac{P(x)}{Q(x)} \right)^\alpha
\end{align}
\end{definition}  
When $\alpha \to 1$, this divergence metric is equal to well known Kullback-Leibler divergence (relative entropy). Additionally, when $\alpha \to \infty$, there is a connection with $\epsilon$-DP definition: a randomized mechanism $f$ is $\epsilon$-DP iff for its distribution over any two adjacent datasets $D$ and $D'$ it holds that 
$D_\alpha(f(D)||f(D')) \le \epsilon$

This connection inspired the introduction of $(\alpha, \epsilon)$-R\'enyi Differential Privacy:
\begin{definition}[ $(\alpha, \epsilon)$-RDP \label{def:rdp} \cite{DBLP:journals/corr/Mironov17}]
Let $f$ be a randomized mechanism $f:\mathcal{D} \to R$, where $\mathcal{D}$ is the space of datasets. $f$ is said to conform to $\epsilon$-RDP definition of order $\alpha$ if for any two adjacent datasets D, D' 
\begin{align}
    D_\alpha(f(D)||f(D')) \le \epsilon
\end{align}
\end{definition}  
Definition \ref{def:rdp} holds iff the tail event random variable between two neighbouring datasets has an $\alpha$ moment upper bounded by $\epsilon$ \cite{DBLP:journals/corr/abs-2001-05990}. Based on this intuition, the Moment Accountant \cite{Abadi_2016} limits all moments of this random variable.

The beauty of RDP is that providing guarantees for composition of many steps of a private process is straightforward: a composition of a number of mechanisms $f_i$ with each $(\alpha, \epsilon_i)$-RDP satisfies $(\alpha, \sum_i \epsilon_i$) RDP
and it is a tighter bound, unlike (even) strong composition of ($\epsilon, \delta$) which has been shown to be loose for many practical mechanisms, including the workhouse of DP-Training -- the Gaussian mechanism \cite{DBLP:journals/corr/Mironov17}.

The following result allows for converting from $(\alpha, \epsilon)$-RDP to ($\epsilon', \delta$) DP: for any $\alpha, \epsilon$ and  $\epsilon'>\epsilon$ $(\alpha, \epsilon)$-RDP implies $(\epsilon, \delta)$-DP where $\delta=exp(-(\alpha-1)(\epsilon'-\epsilon))$ \cite{DBLP:journals/corr/Mironov17,Abadi_2016}. Since this result holds for all orders $\alpha$, to obtain the best guarantees, the Moments Accountant needs to optimize over continuous $1 < \alpha < 32$. \citet{DBLP:journals/corr/Mironov17} however showed that the using only a restricted set of discrete $\alpha$ values is sufficient to preserve the tightness of privacy analysis.

The aforementioned conversion from RDP then allows to obtain an overall DP-Training bound of ($\mathcal{O}(q(e^{\epsilon_s}-1)\sqrt{T}),\delta_f)$ \cite{Abadi_2016}, which is an improvement over strong composition. Roughly, one can obtain these bounds by calculating RDP guarantees for various orders of $\alpha$ and converting them to $(\epsilon, \delta)$ guarantees. Then, the best order that gives the lowest $\epsilon$ is chosen and reported.

Consequently, \citet{DBLP:journals/corr/abs-2001-05990} showed that the conversion from RDP is suboptimal and provided a better bound for going from $(\alpha, \epsilon)$-RDP to ($\epsilon', \delta$)
This resulted in slighter better DP-Training bound than \cite{Abadi_2016}. 
Refer to Table \ref{tab:compositions} for an overview of bounds obtained by various Moment Accounting schemes.

Finally, general tighter bounds were achieved recently for many popular mechanisms by using privacy loss distribution (PLD) accounting instead of RDP accounting \cite{koskela2020tight}.\footnote{As an example, consider a dataset of 1 million examples, a batch size of 5000, and a noise multiplier of 1. Renyi DP (converted to $(\epsilon, \delta)$-DP) gives $(\epsilon, \delta) = (1.2, 1e-6)$ for one epoch, and $(\epsilon, \delta) = (4.95, 1e-6)$ for 100 epochs, while the PLD accountant gives $(\epsilon, \delta) = (0.59, 1e-6)$ for one epoch, and $(\epsilon, \delta) = (4.62, 1e-6)$ for 100 epochs.}

\setlength{\tabcolsep}{5pt}
\begin{table}[!ht]
\footnotesize
\begin{tabular}{ l p{0.3\linewidth} p{0.53\linewidth}}
 \toprule
\textbf{DP-type} & \textbf{DP-Training bound} & \textbf{Comments}\\
\midrule

 $(\epsilon, \delta)$  & ($\mathcal{O}(q (e^{\epsilon_s}-1) T), \mathcal{O}(q \delta T)$) &  Straightforward application of composition, very loose bounds\\
\hline

 $(\epsilon, \delta)$ & $(\mathcal{O}(q (e^{\epsilon_s}-1) \sqrt{T \log{\frac{1}{\delta}}}), \mathcal{O}(q \delta T))$ &  Application of strong composition\\
\hline

 R\'enyi & $(\mathcal{O}(q (e^{\epsilon_s}-1) \sqrt{T}),\delta)$  & Conversion from RDP to ($\epsilon, \delta)$ via Theorem 2  \cite{Abadi_2016}. \\
\hline

 R\'enyi &  $(\mathcal{O}(q (e^{\epsilon_s}-1) {T}^k),\delta)$, $k$ slightly less than 1/2  & Conversion from RDP to $(\epsilon, \delta)$ via Lemma 1 \cite{DBLP:journals/corr/abs-2001-05990}. See Figure 3 \cite{DBLP:journals/corr/abs-2001-05990} for comparison of DP-Training bounds. \\
\hline
\end{tabular}
\caption{\small Evolution of DP-Training bounds at a glance. Assumes that each of $T$ iterations of  DP-SGD Algorithm \ref{algo:abadi}  achieves the same $(\mathcal{O}(q(e^{\epsilon_s}-1)), \mathcal{O}(q\delta))$ guarantee, where $q$ is a sampling ratio and $\epsilon=\sqrt{2 \log{\frac{1.25}{\sigma}}}$.}
\label{tab:compositions}
\end{table}

\normalsize

\subsection{Privacy Amplification via Sampling \Difficult}\label{sec:amplification} 

In the previous section we discussed  DP-Training bounds, which characterize the DP guarantees as a function of several hyperparameters. Some of these bounds rely on the fact that only a portion of the dataset (e.g., a batch) is used during each training step. This section discusses in more detail the nuances of obtaining these bounds.

Stochastic methods of training, where each training step uses a subset (instead of all) of training data are extremely popular in ML. For example, an unbiased stochastic gradient is estimated over a batch and subsequently used for each SGD step. Intuitively, the uncertainty of whether a sample has contributed can ``help'' the privacy of this sample, and a technique named \textit{privacy amplification by subsampling} is widely used to achieve strong privacy-utility trade-offs for practical algorithms like DP-SGD \cite{DBLP:journals/corr/BassilyST14,Abadi_2016}. 

Informally, an algorithm that satisfies $(\epsilon, \delta)$-DP (e.g., based on the  Gaussian mechanism) can achieve stronger $(\mathcal{O}(q(e^\epsilon-1)), \mathcal{O}(q\delta))$-DP with respect to the whole dataset when the algorithm is applied on a subset of data \emph{randomly sampled} with probability $q$ \cite{kasiviswanathan2011can}. The above dependence on $\epsilon$ can be approximated with $q\epsilon$ (when $\epsilon\leq 1$). Semantically, this expression implies that the privacy guarantee roughly gets amplified by a factor of $q$, when the starting privacy guarantee is small. One crucial aspect of privacy amplification is that as $\epsilon$ gets larger, because of the exponential dependence, the amplification guarantee gets weaker.

An alternative view is that in order to achieve the same privacy guarantees, smaller noise that is  $\mathcal{O}(q\sigma)$ can be used when the privacy accounting method is based on amplification by subsampling \cite{kasiviswanathan2011can}. Compared to another commonly used practice of reducing effective noise per iteration by directly increasing batch size (\cite{McMahan2018dplm,anil2021large}~ and Figure \ref{fig:batchsize}), amplification by subsampling can have diminishing returns: increasing batch size is usually more ``efficient'' (for obtaining better $\epsilon$) than relying on the amplification introduced by small batches. Amplification by subsampling is used in the moments accountant with RDP \cite{Abadi_2016,mironov2019r}~for privacy accounting of DP-SGD in practice. Somewhat tighter bounds can be achieved by the recent  privacy loss distribution (PLD) accounting instead of RDP accounting, which also uses amplification by subsampling \cite{koskela2020tight}. 

While amplification is intuitive, the conditions of when it holds and the semantics of the resulting guarantees are sometimes overlooked.
Poisson sampling (selecting each record with probability $q$, leading to variable-sized minibatches) is commonly analyzed using the add-or-remove notion of adjacency, while uniform subsampling (of fixed sized batches independently on each round) is analyzed with the replace-one notion of adjacency  \cite{wang2019subsampled,balle2018privacy}. As discussed in Section~\ref{sec:recordadjacency}, the $\epsilon$'s from these two notion of adjacency have different semantics.

For modern ML models where training data does not fit into memory, it is common to forego true random sampling and to instead perform several passes over a randomly \textit{shuffled} version of the  dataset. In fact, as even fully shuffling the data may be computationally expensive, often the data is only shuffled within a buffer much smaller than the total dataset size. This approach does not satisfy either Poisson or uniform sampling. Therefore, the amplification by sampling results cannot be applied~\emph{as is}. Recent studies suggest that shuffling can also amplify privacy~\cite{erlingsson2019amplification,feldman2022hiding}, but the best known amplification guarantees are weaker than what one would achieve via sampling. It is an important open question to get comparable RDP/PLD amplification guarantees via shuffling. It is common, though inaccurate, to train without Poisson subsampling, but to report the stronger DP bounds as if amplification was used. We encourage practitioners at a minimum to clearly disclose both the data processing and accounting methods (refer to Section~\ref{sec:dp-gur-practice-reporting} for reporting guidelines).  When sampling cannot be guaranteed in the actual training pipeline, alternative approaches such as DP-FTRL \cite{kairouz21practical} that do not rely on amplification may be a preferable option. A practical summary of common data processing patterns (including sampling and shuffling) as well as the algorithm and analysis techniques that can be used for each can be found in Table~\ref{tab:dataprocessing} in Section~\ref{sec:dataprocessing}.

In addition to amplification by sampling or shuffling, privacy can also be amplified by iterations \cite{feldman2018privacy,altschuler2022privacy}~or ``convergence'' \cite{chourasia2021differential} when only the last iterate (checkpoint), instead of all iterates of the model in DP-SGD is released. However, these analyses in the literature require stronger assumptions than amplification by sampling, and can only be applied to convex and smooth functions.

\subsection{Modifications for User-Level DP-Training}\label{sec:dp-user-modifications} 

\begin{algorithm}[ht]
\caption{DP-FedAvg for user-level DP. \cite{McMahan2018dplm}.}
\label{algo:fedavg}
\begin{algorithmic}
\Require Training data $\mathcal{D}$, consisting of $U$ users $\mathcal{D}=\bigcup\limits_{u=1\ldots U} \mathcal{D}_u$ \\
and each user data $\mathcal{D}_u=(X_u, Y_u)$ consisting of both features and labels.  \\
$f(x; \theta)$ is the output of a model parameterized by $\theta$ and applied to an input $x$. \\
$L(y, y')$ is the loss function for label $y$ and prediction $y'$.\\
FedAvg hyperparameters: learning rates $\eta_s$ for global update, $\eta_c$ for local updates, \\
\qquad $T$ number of rounds, $K$ number of local iterations \\
\qquad $B_c$ number of users per round, $B_m$ local batch size. \\
DP hyperparameters: $C$ clipping norm, $\sigma$ noise level.
\Ensure $\theta_T$ final model parameters
\State $\theta_0 \gets $ randomly initialized values
\For{global round $t \gets 1$ to $T$}                     
    \State Randomly sample a subset of users $\mathcal{S}^t$ of $B_c$ users \Comment{Challenging in cross-device FL}
    \For{each user $u \in \mathcal{S}^t$} \Comment{Process data of each user} 
        \State Initialize $\omega_u^0 = \theta_{t-1}$
        \For{local iteration $k \gets 1$ to $K$} \Comment{Local updates}
            \State Sample minibatch $\mathcal{B}^{(t,k)}_u \subset \mathcal{D}_u$ of $B_m$ examples.
            \State $g_u^{(t,k)} \gets \frac{1}{B_m} \sum_{j \in \mathcal{B}^{(t,k)}_u} \nabla_{\omega_u}{L(y_j, f(x_j; \omega_u^{k-1}))}$
            \State $\omega_u^k \gets \omega_u^{k-1} - \eta_c g_u^{(t,k)}$
        \EndFor
        \State  $\Delta^t_u \gets  \omega_u^0 - \omega_u^K$ \Comment{User's model delta}
        \State \HiLi $\Tilde{\Delta}^t_u \gets \Delta^t_u/ \max(1, \frac{||\Delta^t_u||_2}{C})$ \Comment{Clip each user's model delta}
    \EndFor
    \State \HiLiLong $\bar \Delta^t  \gets \frac{1}{B_c}(\sum_{i \in \mathcal{S}^t} {\Tilde{\Delta}^t_i}+\mathcal{N}(0,\,\sigma^{2} C^2 \identity ) )$
     \Comment{Add noise}

    \State $\theta_{t} \gets \theta_{t-1} - \eta_s \bar \Delta^t$ 
    \Comment{Global model update}
    
\EndFor

\end{algorithmic}
\end{algorithm}

The neighboring datasets in the definition of differential privacy (Definition \ref{def:exact_dp} and \ref{def:apprx_dp}) differ in one record, which can be considered the unit of privacy (see Section \ref{sec:unitofprivacy} for an additional discussion about the unit of privacy). The record can be one example in the training data, i.e., \emph{example-level DP} which we have focused on so far. We can also take the record to be the combination of all training examples for a user who contributed their data, i.e., \emph{user-level DP} \cite{dwork2010differential}. In this section we focus on DP-Training algorithms that can achieve user-level DP.

We consider both the decentralized or federated setting where users' data is stored on their own personal devices and the centralized setting where users' data is collected and stored in a datacenter. We focus on DP-FedAvg,  Algorithm \ref{algo:fedavg}, of \citet{McMahan2018dplm} as it is a natural extension of DP-SGD to multi-example as units of privacy, it is a popular choice for user-level DP, and can be used in both centralized and decentralized settings.

\paragraph{Decentralized setting.} User-level DP is a natural choice in federated learning, where decentralized training is used to minimize the exposure of users’ private data \cite{kairouz2021advances,bonawitz22cacm}. 
Federated averaging (FedAvg) \cite{mcmahan2017communication} is the most widely used algorithm for federated training, with DP-FedAvg \cite{McMahan2018dplm} its natural extension to provide user-level DP. Similarly to DP-SGD, DP-FedAvg works by applying Gaussian mechanism to FedAvg. DP-FedAvg can also be considered a variant of DP-SGD with one key change: in DP-SGD, the gradient of each example (or microbatch) is clipped and then aggregated and noised; in DP-FedAvg, a few steps of local updates on model weights are performed on the private data of each user, and the global model delta is created by clipping and then averaging the local updates, and adding appropriate noise to the global update. 

\paragraph{Centralized setting.}
DP-FedAvg described in Algorithm \ref{algo:fedavg} can be a good choice for user-level DP in scenarios beyond federated learning, e.g., when users' data are collected and stored in a datacenter. Unlike on-device training in federated learning, the concept of \emph{local updates} is generalized to training on the data from a specific user collected in a datacenter. However, the local updates in FedAvg that enable low frequency communication between the aggregation server and the devices also introduce challenges for convergence (e.g., in terms of loss stabilization, in both centralized and decentralized settings) due to the heterogeneity of data from various users \cite{wang2021field}. A special case of 
DP-FedAvg called DP-FedSGD is studied in \cite{McMahan2018dplm}: DP-FedSGD will only aggregate the local gradients without updating the local models, i.e., $\Delta^t_u = \sum_k g_u^{(t,k)}$. As such, DP-FedSGD is very similar to DP-SGD with mircobatches (refer to Algorithm \ref{algo:microbatch} in Section \ref{sec:microbatches}) where
each microbatch is constructed using the (sample of) a particular user's data (instead of sampling from multiple users' data), and setting local learning rate $\eta_c=1$. 

In terms of utility, DP-FedAvg has been shown to outperform DP-FedSGD, mostly due to the fact that the convergence of FedAvg can be superior for a wide range of practical applications \cite{wang2022unreasonable}. Recently \citet{xu2022learning} proposed \emph{virtual clients} by extending microbacthes of examples to groups of users, which can also be used to mitigate the heterogeneity issue for convergence. 

\paragraph{Additional discussion.} 
We highlight that in Algorithm \ref{algo:fedavg} there are more hyperparameters to tune than in standard DP-SGD, i.e., additional learning rate $\eta_c$ and local iteration number $K$. The hyperparameter tuning strategy outlined in Section \ref{sec:tuning_hyperparams} can be applied with some modifications: $K$ and $\eta_c$ can be tuned once and fixed for other experiments. Additionally, the clipping norm $C$ exhibits dependency on $K$ and $\eta_c$ (contrary to the clipping norm in DP-SGD that does not depend on the learning rate). Clipping norm can also be estimated via adaptive clipping \cite{andrew2021adaptiveclip}.

Additionally, even though user-level DP has recently been studied in the centralized setting with collected data in datacenter \cite{xu2022learning}, its primary application still remains to be in federated learning with decentralized data. Amplification by subsampling, as one of the key techniques for achieving strong privacy guarantees, is very challenging in current cross-device federated learning systems \cite{balle2020privacy}. The recent DP-FTRL algorithm \cite{kairouz21practical,choquette23amplified} that does not rely on subsampling is much easier to deploy on such systems,  and has been applied in practice to train production models with user-level DP \cite{thakurta22fldp_blog,xu2023gboard}. For more discussion of differential privacy in federated learning, including local differential privacy and other system considerations, we refer readers to Section 5 of \citet{kairouz2021advances} and Section 7.1 of \citet{wang2021field}. 

\subsection{Challenges with DP-Training}\label{sec:problems-dp-sgd}

While DP-training provides strict privacy guarantees, there are multiple obstacles preventing its widespread adoption.

\paragraph{Loss of utility.} Private training usually comes with a decrease in utility (where ``utility'' is a collective name for evaluation metrics); that is, private models often perform worse than their non-private counterparts in terms of accuracy, precision, or any other metrics measuring model quality. Typically, a lower $\epsilon$ (i.e.,  stricter privacy guarantees) corresponds to a more significant loss of utility.
Especially for datasets that are small relative to model capacity, the loss of utility required to achieve a small $\epsilon$ may be so significant that the private model may be no longer useful for practical tasks. For example, the best known private ImageNet model,  when trained without extra data~\cite{deepmind2022dpimagenet}, achieves only $32.4\%$ accuracy at $\epsilon=8$. In contrast, non-private training of the same model (NF-Resnet50~\cite{brock2021nfresnet}) achieves $76.8\%$ accuracy.

There are eight main themes that attempt to mitigate the performance drop. We discuss these themes below:
\begin{compactenum}
    \item \textit{Use more computation}. Training models with DP requires tradeoffs between model utility, the strength of the privacy guarantee, and (importantly) the amount of computation used. Specifically, using larger batch sizes and/or more DP-SGD iterations can significantly help model accuracy at a fixed privacy cost. In fact, for a sufficiently large dataset, a large computation budget can possibly offer nearly non-private utility together with an $\epsilon \le 10$. More details on hyperparameter tuning including batch size and number of iterations can be found in Section~\ref{sec:tuning_hyperparams}.
    
    \item \textit{Tuning other hyperparameters} can also significantly improve the utility of DP-SGD training. In particular, joint tuning of learning rate and clipping norm has been shown to have large impact on utility~\cite{kurakin2022dpimagenet} (see Section~\ref{sec:tuning_hyperparams} for details).
    
    \item \textit{Increasing the amount of data available for training}. \citet{DBLP:journals/corr/abs-2011-11660} argue that the larger training dataset, the better is utility of the private model. Thus, collecting more training data could help boost utility.
    In a similar vein, \citet{DBLP:journals/corr/BassilyST14} derive upper and lower bounds for excess risk in a DP version of (convex) empirical risk minimization. 
    Specifically, they show that excess risk bound in DP-training exhibits inverse polynomial dependency on dataset size. Intuitively, this means a DP-trained model would perform better with an increase of training dataset size, in the limit reaching the performance of the non-private model, all other things fixed.
    \item \textit{Handcrafted features.} 
    \citet{DBLP:journals/corr/abs-2011-11660} show that using handcrafted wavelet-like ScatterNet features improves the  accuracy of DP-trained models.
    They argue that the use of handcrafted features results in an easier learning task and faster convergence (e.g., loss stabilization).
    At the same time, one can argue that handcrafted features also leak the private information.
    Moreover, choosing good handcrafted features might be not an easy task by itself.
    \item \textit{Utilizing public data.}
    Utilizing public data from a distribution which is similar to the distribution of private data could significantly boost utility. The most straightforward way to do this is to pre-train a model using public data and then fine-tune this model with DP-Training methods like DP-SGD using private data~\cite{kurakin2022dpimagenet,deepmind2022dpimagenet}. Equivalently, one can start with public checkpoints (like ImageNet, ResNet for image data and BERT \cite{DBLP:journals/corr/abs-1810-04805} or GPT for text data) and fine-tune (with DP) these checkpoints using private data. 
    Some other more sophisticated ways to utilize public data during DP training were reported recently. For example, \citet{amid2022publicdata} utilizes public data to better guide gradient descent during private training. However, care must be taken in selecting the ``public'' dataset, as even publicly available datasets might contain private information \cite{tramer22considerations}.
    
    \item \textit{Model weights averaging}. An extremely simple and computationally cheap idea that is related to ensemble learning is to average intermediate weight values obtained from different checkpoints during DP-training.  For example, \cite{deepmind2022dpimagenet} report improvements from this strategy using an exponential moving average. This method does not incur additional privacy costs since all weight values (including from intermediary checkpoints) are considered public when obtained from DP-Training.
    \item \textit{Architectural adjustments}. 
    In practice, it is common to transfer the architecture of a non-private model and reuse it for DP-training, while tuning the batch size, clipping norm, and learning rate. However, several works argue that appropriate architectural decisions can result in better privacy/utility tradeoffs. For example, \citet{papernot2020tempered} advocated for the use of bounded activation functions (as opposed to unbounded like common RELU and Sigmoid) when using DP. Additionally, various other architectural adjustments such as increasing the batch size, or using batch/layer normalization were proposed, for example in \citet{davody2020effect}. We discuss a number of these suggestions in detail in Section \ref{sec:architecture}.
    \item \textit{Relaxation of privacy guarantees}. When utility drop remains unacceptable, practitioners may consider aiming for weaker privacy guarantees (see Section \ref{sec:target-eps}). Alternatively, heuristic methods can also be employed, in place of providing theoretical privacy guarantees. For example,  \citet{DBLP:journals/corr/abs-1802-05214} demonstrate  ``empirical privacy'' by preventing discovery of some predefined ``private attributes'' from the data (so the model is unable to infer, for example, the race or income level of the participants).
\end{compactenum}

\paragraph{Slower training.} 
As we mentioned in Section \ref{sec:dpsgd}, modern machine learning frameworks are optimized for standard backpropagation, in which operations (such as computation of gradients) are performed on an aggregate, batch level. 
However, DP-Training procedures like DP-SGD perform a  non-standard backpropagation that requires the computation and clipping of per-example (not batch-level) gradients. 
A naive implementation of DP-Training involves the computation of per-example gradients, clipping of per-example gradients,  and finally aggregation of clipped gradients. This process is typically much slower than computing per-batch gradients, e.g., by orders of magnitude in TensorFlow ~\cite{subramani2020fastdpsgd,kurakin2022dpimagenet}.

In general, there is no known way to compute per-example gradients as fast as aggregated gradients. The following ways are explored in attempt to mitigate this issue:

\begin{compactenum}
    \item \textit{Efficient implementation of per-example gradient clipping.}
    It is possible to ensure per-example gradient clipping without fully computing per-example gradients. Instead, it is enough to compute only the norms of per-example gradients and then use them to re-weight the model loss during the backward pass.
     This trick allows to perform per-example gradient clipping at a cost of one forward and two backward passes through the network. This idea was first explored in~\cite{goodfellow2015trick} for fully connected and later expanded to other types of layers ~\cite{DBLP:journals/corr/abs-2009-03106,li2022llmdp}.
    
    \item \textit{Choosing an efficient DP framework.}
    Some of the existing DP frameworks can perform efficient per-example clipping automatically, thus relieving the practitioner from the need to manually optimize the code.
    PyTorch Opacus~\cite{opacus} implements efficient per-example gradients for some types of neural network layers.
    JAX can automatically perform efficient vectorization of per-example gradient computation~\cite{subramani2020fastdpsgd}.
    This allows DP-SGD to run $\approx 1.5\times$ slower compared to regular SGD, which could be considered acceptable cost when privacy is at stake. \citet{ponomareva-etal-2022-training} reported that using modern JAX primitives like \texttt{vmap}, their DP-Training version that takes per example gradients is only $25\%$ slower than the version that does not taker per-example gradients, with all other things fixed.
    \item \textit{Gradient clipping at microbatch level.} Instead of clipping the norm of each example's gradients in the batch, some frameworks like Tensorflow Privacy allow clipping at the microbatch level: a batch of examples is split into a number of microbatches, the average gradient per microbatch is calculated and clipped according to the clipping norm, and these clipped averages are aggregated across all the microbatches from the batch and the noise is subsequently added. 
    While this approach preserves the same privacy guarantees and allows to reduce memory requirements while improving the speed of training, it adds more noise compared to per-example clipping and thus tends to hurt model utility. Refer to Section \ref{sec:microbatches} for in-depth discussion on microbatches.
\end{compactenum}

\paragraph{Increased memory footprint.}
In addition to being slower, per-example gradient clipping requires more accelerator memory compared to regular per-batch gradients.
Additionally, as discussed in Section~\ref{sec:tuning_hyperparams}, practitioners may want to consider increasing the batch size for DP-SGD training to improve model utility.
This can significantly elevate the memory requirements of  DP-SGD, especially for large models. There are several ways to overcome this aforementioned issue:

\begin{compactenum}
    \item \textit{Increase the number of accelerators in distributed training.}
    That is the most straightforward way if extra accelerators are available.
    \item \textit{Use gradient accumulation.}
    The idea of gradient accumulation (also sometimes referred to as \textit{virtual batch}) is somewhat similar to microbatching. At each step, a small batch is drawn, its per-example gradients are clipped. Then instead of adding the noise and applying the gradient update as per DP-SGD, the sum of the clipped gradients is saved/accumulated. Then the next batch is drawn and the sum of it is clipped gradients is added to the running sum. After a number of steps (after a large number of example has been processed, essentially representing a large enough batch), the gradient step update (with the added noise) is applied to the model's weights. This approach allows to simulate an arbitrarily large batch on an accelerator with limited memory~\cite{opacus,kurakin2022dpimagenet}.
    \item \textit{Efficient algorithm tailored to specific models/layers.}
    Some of the algorithms designed for efficient per-example gradient calculation also help with memory consumption. E.g., ghost clipping~\cite{li2022llmdp} significantly reduces memory footprint of training transformers by optimizing per-example gradient clipping for sequential inputs.
\end{compactenum}

\section{Practicalities of DP-Training}\label{sec:practicalities}
In this section we first outline the decisions that a practitioner should make, then we discuss reporting and hyperparameter tuning. We also look into how architectural choices affect privacy and performance and conclude with information about the DP tooling options. 

\subsection{Choosing the Right Unit to Protect}\label{sec:unitofprivacy}

\begin{specialistbox}{Choose unit of protection}
One of the important decisions when applying DP to a complex model is to determine what unit of the data needs to be protected. The unit of protection determines what makes two datasets ``neighbouring'' in DP Definition \ref{def:exact_dp}, essentially defining a ``sample''.  
\end{specialistbox}
In the context of a machine learning model, the most common units are:
\begin{compactenum}

\item \textit{Instance-level protection} or \emph{example-level DP} protects both the features and the labels of each instance (sample) in the dataset.  Unless stated otherwise, all the DP-Training methods we discuss work at this level, thus assuming that the training data is simply a collection of independent instances. If instances are not unique (e.g., an  instance can be repeated multiple times in the training data), the guarantees for such repeated instances are ``diluted'' by the number of repetitions. 
\item \textit{Sub-instance level protection} can be used where only a subset of the features is considered private, or if only the labels are considered private (Label-DP). For example, for tabular data, a single instance can have a number of features, some of which can be private (e.g., the name of a respondent) and some may be considered public (e.g., the city). It is possible to choose to protect only these private attributes; however, this is often achieved by heuristic methods during training, such as using adversarial models \cite{DBLP:journals/corr/abs-1911-10143}; such methods do not provide DP guarantees. Labels-only protection on the other hand can be achieved with DP-Training (Section \ref{sec:label-protection})

\item \textit{User-level/Group-level protection}. If the data was generated by multiple users, a user-level protection might be better suited for an ML model. User-level protection would mean that the neighbouring datasets definition is based on inclusion/exclusion of all the data of any one user (potentially a very large number of examples). Similar to user-level protection, group-level protection uses some grouping function for the data and the definition of ``neighbouring'' datasets is modified to include all  samples from one group.
There are modifications to the training process (DP-Training) that can guarantee user-level DP-Protection (refer back to Section \ref{sec:dp-user-modifications} for modifications to the standard DP-Training algorithm). We also note that user-level protection often is explored in context of Federated Learning \cite{kairouz2021advances,wang2021field}. 

\item \textit{Units of privacy for text and sequence data}. For many applications, e.g. typical classification tasks on feature vectors or images, the notion of an example is a well-defined semantic concept. However, for sequence models where conceptually the training data might be a single very long sequence of text (e.g., text corpora like c4 \cite{DBLP:journals/corr/abs-1910-10683}), more care needs to be taken in defining the unit of privacy.  The basic application of DP-SGD will generally protect ``one row in a batch'', or essentially a number of tokens that depends on a sequence-length or unroll-length hyperparameter. However, it is important to note that this hyperparameter no longer only influences model performance characteristics, but also fundamentally controls the semantics of the DP guarantee. At the same $(\epsilon, \delta)$-DP statement, a sequence length of say 32 vs 128 have substantially different privacy properties. In order to decouple the batch width from the privacy guarantees, it is possible to use algorithms for group-level protections described above to for example protect sequence data at a more semantically meaningful level, e.g. for text data one might desire sentence-level, paragraph-level, document-level, or user-level guarantees for different purposes.
\end{compactenum}
At the end of the day, the choice of unit of protection is both extremely important and application dependent. The privacy guarantees (e.g. specific $(\epsilon, \delta)$ guarantee) and/or model accuracy will also depend on the unit chosen, because slicing the data with various levels of granularity essentially changes the training dataset size, which is one of the most important factors that influences the guarantees achieved (see Section \ref{sec:dp-gur-practice}).

\subsection{What is a Good $\epsilon$ for an ML Model} \label{sec:good-eps}
Below we first summarize our recommendations for selecting $\epsilon$ value, followed by the discussion of these recommendations in Section~\ref{sec:eps-justification}.

\subsubsection{Our Recommendations for $\epsilon$ Values for ML models}\label{sec:target-eps}

\begin{specialistbox}{Target $\epsilon$ for DP ML models} We encourage practitioners to choose the lowest possible tier from the below.
We consider \textbf{user-level} DP (or \textbf{example-level} where a single user or other appropriate group contributes at most one example) with the add-or-remove or zero-out adjacency.

\begin{compactenum}

\item \textbf{Tier 1: Strong formal privacy guarantees.} Choosing $\epsilon \le 1$ provides a strong privacy guarantee directly via the DP definition. However such $\epsilon$ values frequently result in large utility drop for large ML models, and may be infeasible.
\item \textbf{Tier 2: Reasonable privacy guarantees.}
In this tier, we advocate for the currently undocumented but still widely used goal for DP ML models of achieving an $\epsilon \le 10$ in order to provide a reasonable level of anonymization for many applications. 
\item \textbf{Tier 3: Weak to no formal privacy guarantees.}
Any finite $\epsilon$ is an improvement over a model with no privacy protections, for several reasons: 1) A finite $\epsilon$ moves the model into a regime where further privacy improvements can be quantified; and 2) as discussed below in Section \ref{sec:eps-justification}, even large $\epsilon$s can indicate a substantial decrease in a model's ability to memorize user data. However, for a large $\epsilon$ (e.g., $> 10$), the DP guarantee on its own cannot be taken on as sufficient evidence of data anonymization, and additional measures (e.g, empirical privacy auditing, demonstrated robustness to attacks, or pre-processing to remove privacy-sensitive data from the training set) may be necessary before releasing or deploying the model.
\end{compactenum}
\end{specialistbox}

\subsubsection{Discussion and Justification}\label{sec:eps-justification}
Answering the question of what level of protection is appropriate for a given application requires balancing factors including the strength of the formal privacy guarantee at an appropriate unit-of-privacy, known vulnerabilities and memorization characteristics of the model architecture at hand, the cost to model performance (e.g. accuracy or top-line user interaction metrics), the computational cost of training (e.g., larger batch sizes), and the possible costs of acquiring additional data. Additionally, the range of good $\epsilon$ is determined by both where DP is applied (Section \ref{sec:wheretoapply}) as well as the unit of privacy (Section \ref{sec:unitofprivacy}), and the precise notion of record adjacency (Section \ref{sec:recordadjacency}). Our recommendations above are informed by the choice of $\epsilon$ in real-world applications of DP and evidence from the academic literature on DP Training, that we present below. We follow by discussion on empirical evidence from privacy auditing and attacks, and then provide additional arguments outlining why we still advocate for DP even when only large values of $\epsilon$ result in acceptable utility. 

\paragraph{Real-world DP deployments}
For aggregate statistics (e.g., raw data and not ML models), low single and double digit $\epsilon$'s are commonly adopted. For example, \citet{census} used $\epsilon=12.2$ to release its demonstration data (privacy unit is a person), \citet{covid} employed $\epsilon=2$ to quantify the mobility changes of Facebook users (user-day privacy unit), and Apple collected various data from end users running iOS or macOS using $\epsilon$ ranging from $2$ to $16$, again using user-day privacy unit with add-or-remove adjacency \cite{apple,apple_patent}.
  
The story gets murkier for DP-Training of ML models, particularly since the use of DP in production settings is currently very limited. Microsoft \cite{msft21ppmlblog} mentions using DP with $\epsilon\!=\!4$ covering all contributions from a user in a six month window (a notion stronger than example-level privacy, but weaker than full user-level privacy), but does not detail any specific production uses of DP ML. In fact, we are aware of a single DP launch for a model trained on private data with a publicly stated DP guarantee, Gboard's use of DP for Spanish-language next-word prediction \cite{thakurta22fldp_blog}. This work used \emph{device-level DP} (protecting all of the examples on any one user's device, equivalent to user-level DP if each user has one device) with $\epsilon=8.9$ and zero-out adjacency \cite{ftrl-colab}; these values are derived from the more precise guarantee of $\rho=0.81$ zCDP (see Section \ref{sec:definitions}). About twenty follow-up Gboard language models were subsequently trained and launched with zCDP guarantees $\rho \in (0.2, 2)$ \cite{xu2023gboard}.

\paragraph{Academic literature on DP training.}
Below we attempt to draw some conclusions about privacy/accuracy tradeoffs from the academic literature, but it is worth emphasizing that the datasets used in such experiments are typically small, and much better privacy/accuracy tradeoffs are generally possible by using larger datasets or more computation. \textit{To the best of our knowledge, all the examples below focus on example-level privacy with add-or-remove adjacency unless otherwise noted.}

For small models (e.g., one or two hidden-layers) achieving reasonable performance is possible with $\epsilon$ between $0.1$ and $10$ \cite{Abadi_2016}. 
For giant models, such as Large Language Models (LLMs), the most common application of DP is \textit{DP fine-tuning} a publicly pretrained model, which can achieve good performance for low-digit $\epsilon$'s. For example, \citet{yu2021differentially} reported a privacy budget of $\epsilon=6.7$ on RoBerta models with approx 3\% (relative) drop in performance compared to a non-private model (unknown length of privacy unit). Similar to LLMs, good performance can be achieved for ResNet with public pretraining and DP-fine-tuning \cite{deepmind2022jax-privacy}.

For DP training of Large Language Models \textit{from scratch}  \citet{ponomareva-etal-2022-training} reported large pre-training performance drop for low-digit privacy budgets (with 626 SentencePiece tokens of text as privacy unit and add-or-remove adjacency). For example, a T5 model with an $\epsilon=6.06$ exhibited a 34\% relative drop; albeit the authors highlight that on the final non DP fine-tuned task performance was not affected. Similarly, \citet{anil2021large} reported that pre-training Bert with DP using  mega-batches with $\epsilon=5.36$ results in 14\% relative  drop (unit of privacy is 128 WordPieces tokens, add-or-remove adjacency). 

In contrast to full training data protection, Label-DP is easier to achieve and requires less noise; label-level DP algorithms can work well with small $\epsilon$'s--e.g., achieving only 3\% relative performance drop for an $\epsilon=8$ on CIFAR 100 for Resnet model \cite{DBLP:journals/corr/abs-2102-06062}.

\paragraph{Evidence from empirical privacy attacks} 
Membership inference (MI) attacks seek to determine whether a particular training example was present in the training data (e.g., a particular patient was in cancer dataset). There is empirical evidence that demonstrates that if robustness w.r.t membership inference attack is the ultimate goal, one might get equally good empirical protection without DP methods. For example, \citet{DBLP:journals/corr/abs-1902-08874} argues that privacy leakage is exacerbated by overfitting and \citet{Blanco_Justicia_2022} demonstrates that in some cases, the amount of protection against privacy leakage that DP provides for large values of $\epsilon$ (Tier 3 in our guidelines) is comparable with other non-DP noise addition/regularization techniques like dropout or l2 regularization, which don't come with increased computation cost. \citet{cummings2023challenges} argue that while DP requires bounding the sensitivity and noise injection, just the sensitivity-bounding step like clipping gradients can mitigate many state-of-the-art privacy attacks like membership inference attacks. 

Privacy Auditing of machine learning models has been proposed to empirically measure the privacy leakage of ML training algorithms~\cite{Jagielski2020,nasr2021adversary}. While membership inference attacks can be used to perform empirical privacy auditing~\cite{10.5555/3361338.3361469}, recent literature introduced stronger attacks to provide better empirical estimation of the $\epsilon$ privacy parameter. \citet{Jagielski2020} proposed the idea of crafting worst-case data poisoning examples that increase the success of the adversary in performing a distinguishing test between neighboring datasets and result in sharper lower bounds on $\epsilon$ than standard membership inference attacks. Follow-up work explored several designs of data poisoning canaries for auditing in both centralized~\cite{nasr2021adversary,lu2022a}~and federated learning~\cite{CANIFE} under different threat models. While initial methods for privacy auditing required training of thousands of models~\cite{Jagielski2020,nasr2021adversary,lu2022a}, privacy auditing can be made efficient by performing the $\epsilon$ estimates with fewer models~\cite{pillutla2023unleashing}, and even in ``one-shot’’, by training a single model~\cite{andrew2023oneshot,steinke2023privacy}.  \citet{nasr2023tight} showed that privacy auditing results in tight estimates of $\epsilon$ for the Gaussian mechanism, when the adversary gets access to intermediary model updates during training. However, in a more realistic setting in which the adversary only observes the final model’s predictions or does not know the specifics of the privacy mechanism, there is still a large gap between the empirical estimates and theoretical analysis even under strong attacks~\cite{nasr2021adversary,nasr2023tight}. For these settings, in which tight theoretical analysis might not prove feasible, privacy auditing techniques provide empirical estimation of privacy leakage, which could inform practitioners on the choice of privacy tiers. For instance, a Tier 2 level $\epsilon$ upper bound might be acceptable for releasing access to a model’s predictions, if the strongest known privacy auditing attack provides an order-of-magnitude lower $\epsilon$ estimate.

\paragraph{Evidence from empirical reconstruction attacks}
There is a growing literature providing evidence that DP training with even large $\epsilon$'s can result in protection against a variety of specific threats, particularly various forms of reconstruction attacks like training data extraction attacks. While these results are necessarily limited to specific threats, they nevertheless provide evidence that DP training can provide useful privacy benefits even if the formal DP guarantee is relatively weak. 
Empirically, \citet[Table 3]{carlini19secret} showed that example-level $\epsilon$ values as high as $10^9$ produced a significant decrease in memorization for a language model. \citet{ponomareva-etal-2022-training} similarly demonstrated that for example-level DP with $\epsilon=320$, success of training data extraction attack was reduced $15\times$  for large language models.
\citet[Fig. 9]{balle22reconstructing} showed that example-level $\epsilon$s in the  $10^2$ to $10^4$ range significantly decreased the effectiveness of a reconstruction attack with almost no impact on test accuracy for the DP model. 
Formal relationships between DP guarantees and reconstruction attacks have also been established \cite{bhowmick19protecting,balle22reconstructing,guo22boudning,guo22fano,stock22defending}, often with the goal of directly informing the choice of $\epsilon$ if the primary concern is a specific notion of reconstruction.

There is a natural intuition for why larger $\epsilon$s provide effective protection in these works --- the attacks generally consider an adversary attempting to answer a high-dimensional question (e.g., reconstructing a full training example) with only limited information about the dataset (e.g., distributional). 
This is in sharp contrast to the adversary implicitly encoded by the DP definition: an adversary that knows that the model was trained on the precise dataset $D$ or a specific neighboring one $D'$, and needs to answer only a single binary question (which dataset was used, $D$ or $D'$?). Recent work on empirical privacy auditing has shown that a strong adversary that better matches the assumptions of DP \emph{can} construct attacks that are almost as successful as the lower bound $\epsilon$ would predict (that is, you really need a small $\epsilon$ to protect against these attacks) \cite{nasr2021adversary}.  Hence, the degree to which memorization-measurement and reconstruction results should be used to justify a larger $\epsilon$ depends strongly on the types of adversaries that are a concern.

\paragraph{Additional discussion}
The aforementioned discussion demonstrates that there is no consensus as to what $\epsilon$ to aim for large ML models trained with DP-Training methods. From a practical point of view, $\epsilon=10$ (or its vicinity) seems to be a ``sweet'' spot where it is possible to preserve an acceptable utility for complex ML models. However from the DP point of view, the $\epsilon\sim10$ guarantees might seem dubious. After all, referring back to Definitions \ref{def:exact_dp} and \ref{def:apprx_dp}, this value of $\epsilon$ would translate into the probability of a particular outcome changing by 22026 times on two datasets that differ only by one instance (in case of instance level privacy). On one hand, this does not represent particularly strong privacy guarantees. However, most DP-Training methods (e.g., DP-SGD) are \textit{iterative} (as opposed to one-shot) algorithms whose final guarantees are obtained by composition of guarantees from each iteration (Section \ref{sec:dpaccounting}). This composition assumes that all intermediate results are released, which is not what happens in practice when only the final checkpoint is used for subsequent inference \cite{feldman2018privacy}. Our current understanding of DP-Training accounting relies on a number of techniques like RDP composition and privacy amplification (Section \ref{sec:amplification}, \ref{sec:dpaccounting}).  We believe that better accounting methods will demonstrate that DP-guarantees for ML models are actually better than currently thought. As a first step, \citet{feldman2018privacy} recently argued that not releasing intermediate results during training allows (under certain conditions on iterative process) to significantly amplifies the privacy guarantees of iterative models. This approach can amplify privacy even in settings where privacy-amplification-by-sampling can't be used (e.g., the noise level is too low). The downside of this new technique is that privacy guarantees depend on when an instance was visited --- instances from earlier batches enjoy stronger privacy guarantees than those observed closer to the end of the training process.

\paragraph{Discussion of Tiers 2 and 3}
While there is general consensus about guarantees for $\epsilon < 1$, application of DP for larger $\epsilon$ values (Tier 3, and to an extent, Tier 2 in our guidelines), might be controversial for some in DP community. Tier 3 essentially offers little to no formal privacy guarantees (as \citet{Blanco_Justicia_2022} calls it, these guarantees are "DP in the name only"). The downside of foregoing the DP completely (for example, in Tier 3) and relying on privacy auditing is that such auditing attacks provide a \textit{lower bound} on privacy, where acceptable performance during the attack is not a sufficient condition, and a new attack discovered later on could demonstrate more privacy leakage than was previously determined \cite{cummings2023challenges}. DP on the other hand provides an \textit{upper bound} and allows to quantify privacy improvements between different versions of the model. As stronger attacks have been developed, the gap between lower and upper bounds got tighter. Additionally, new line of work demonstrated that it is possible to derive the bounds of success of a particular empirical class of attacks like membership inference \textit{from DP bounds} \cite{DBLP:journals/corr/abs-1709-01604,DBLP:journals/corr/abs-1908-03566,sablayrolles2019whitebox,DBLP:journals/corr/abs-2005-10881}, without having to do empirical privacy auditing \cite{cummings2023challenges}. Such estimates would hold even for new previously undiscovered attacks of the class at hand. Therefore, we do believe that empirical auditing is beneficial in Tier 3, and possible in Tier 2, and can complement our DP privacy guarantees but is not a replacement for training with DP. We refer reader to \cite{cummings2023challenges} for much richer discussion on this topic.

\subsection{Calculating and Reporting Privacy Guarantees} \label{sec:dp-gur-practice}
In this section we first draw attention to the need of understanding data processing to implement correct privacy accounting. We then describe how to calculate DP-SGD guarantees in practice, and how hyperparameters affect the $\epsilon$. We then provide recommendations for rigorous reporting of privacy guarantees that we hope will result in better reproducibility and fair comparison between various DP ML models.  

\begin{table}[h]
\footnotesize
\renewcommand{\arraystretch}{1.3}
\begin{tabular}{p{1.8cm} p{5cm} p{7.5cm}}
 \toprule
\textbf{Data \mbox{processing}} & \textbf{ Minibatch Construction}  &  \textbf{Algorithms and accounting} \\
\midrule
Poisson \mbox{sampling}
& Independently samples each example with a probability of inclusion $q$, therefore resulting in batches of different sizes. 
& DP-SGD can be analyzed using RDP or PLD accounting with the add-or-remove neighboring relation; amplification-via-sampling (Section~\ref{sec:amplification}) provides a substantial improvement in the privacy guarantees. \\
\hline
Uniform \mbox{sampling}
& Samples a fixed size batch from the training data without replacement for each batch, but with replacement across batches.
& DP-SGD can be analyzed via the methods of \citet{balle2018privacy} under the replace-one neighboring relation; again, sampling provides a substantial improvement. \\
\hline
Shuffling 
& Permutes all examples, producing an ordering, and then partitions the examples to batches of fixed size. This strategy represents one special case of single- or multi-epoch rows below.
& Commonly implemented and used in centralized ML training infrastructure, though care must be taken to ensure the whole dataset is randomly permuted. DP-FTRL does not directly leverage shuffling, but can provide strong guarantees, see the following two rows.
\\
\hline
\hline
Single epoch
& Each example participates once, in an arbitrary order.
& DP-FTRL can provide strong guarantees, via either tree aggregation \cite{kairouz21practical} or with improved results via matrix factorization \cite{denisov22matfact}; DP-SGD's guarantees tend to be weak since no amplification applies.
\\
\hline
Multiple \mbox{epochs}
& Each example participates a fixed number of times.
& If participations from the same example can be separated by a sufficient number of iterations, DP-FTRL can provide strong guarantees, either via tree aggregation \cite{kairouz21practical} or with improved results via matrix factorization \cite{choquette22multiepoch,choquette23amplified}; DP-SGD guarantees tend to be weak since no amplification applies. \\
\bottomrule

\end{tabular}
\caption{\footnotesize Data processing patterns in training and privacy accounting. Row groups are not mutual exclusive: the single- and multi-epoch cover cases where sampling was not used or cannot be verified.}
\label{tab:dataprocessing}
\end{table}

\subsubsection{Data Processing Patterns, Amplifications, and Accounting} \label{sec:dataprocessing}
\begin{specialistbox}{Privacy accounting assumptions should match training reality}
The data processing workflow used to select training examples and form them into batches has a substantial impact on the privacy properties of the training mechanisms, and should influence the choice of DP algorithm and accounting technique. 
\end{specialistbox}
Ideally DP ML systems should be fully integrated with accounting approaches, so all parameters from training required for privacy accounting are logged programmatically and can be automatically consumed by appropriate accounting libraries. For example, the \texttt{DpEvent} representations in the Google DP library are one effort to establish such a representation (though it is not yet fully integrated with TF Privacy). However, currently some manual steps are often involved in selecting and running accounting routines.  As an example of potential mismatches, as recently noted by \citet{choquette22multiepoch}, numerous papers have (technically incorrectly) reported $\epsilon$ guarantees using the RDP or moments-account analysis of Poisson sampling when the actual training used (partial or full) shuffling of the training data with fixed sized batches. This inaccuracy goes back to the experiments reported by \citet{Abadi_2016}. While it is plausible to hypothesize that the shuffling with fixed sized batches might produce similar privacy amplification gains to Poisson sampling, this remains an important open theoretical question. 

Thus, currently the burden is on practitioners to 1) understand the data processing pattern used by their ML infrastructure, 2) appropriately transfer the necessary parameters to the accounting library, 3) at a minimum accurately document any mismatch between analysis assumption and infrastructure, such as the Poisson-vs-shuffling issue noted above. Table~\ref{tab:dataprocessing} summarizes commonly used families of data processing patterns, the recommended algorithms and accounting techniques.

\subsubsection{Calculating Training Process Guarantees for DP-SGD}\label{sec:calculating-and-reporting-eps}
\begin{specialistbox}{Convention for setting $\delta$}
While the mechanisms and algorithms that we discuss throughout this paper (like DP-SGD) do not suffer from catastrophic failure, it is still recommended to set $\delta$ to a small value, hence the convention to use $\delta \ll \frac{1}{n}$, for example $\delta=\frac{1}{n^{1.1}}$, where $n$ is the training dataset size (measured in terms of the unit-of-privacy)\footnote{The above suggestion considers dataset size $n$ to be non-sensitive information. When $n$ is unknown or considered private, you can set the value of $\delta$ based on an estimate}.
\end{specialistbox}

Most major libraries that implement DP-SGD provide a routine to post-hoc calculate the achieved $\epsilon$ value of the training process. It is expected that $\delta$ is set to be less than the inverse of the training data size. Most of these routines currently assume \textit{example-level unit of protection, the add-or-remove definition of neighbouring datasets, and that data is processed using variable-sized batches formed via Poisson sampling, in the central DP setting}. When these assumptions hold, there are only three parameters that affect the final $\epsilon$:
\begin{compactenum}
\item \textbf{Noise multiplier $\sigma$} for the Gaussian mechanism applied at each step. Note that during DP-Training this noise is amplified by the clipping norm $C$ (e.g., Gaussian noise is drawn from $\mathcal{N}(0,\,\sigma^{2} C^2 \identity )$ as per Algorithm \ref{algo:abadi}), so $\epsilon$ does \textbf{not} depend on the clipping norm.
\item \textbf{Example sampling rate}, the probability of each example being selected (independently) for the batch. Alternatively, some implementations ask for the batch size and the dataset size. 
\item \textbf{Number of training steps}. Some routines ask for the batch size and \# of epochs.
\end{compactenum}

If the user instead wants to find the appropriate level of noise or batch size to use in order to achieve a desired $\epsilon$, a  binary search can be performed by relying on  these routines to evaluate the $\epsilon$ for each $\sigma$.  For example, Google's DP libraries\footnote{https://github.com/google/differential-privacy/tree/main/python} provide the \texttt{dp\_accounting.calibrate\_dp\_mechanism} routine to facilitate such searches.

\paragraph{$\epsilon$ scaling laws.}
While these routines are essentially black-box for the end-user, there is a very rough approximation using advanced composition (refer to Appendix \ref{app:epsilon_epoch_derivation}) that helps understand the ``scaling laws'' --- how $\epsilon$ guarantees change with the change in the three parameters discussed above:
$
\epsilon \approx A \dfrac{q \sqrt{k}}{\sigma} + B \dfrac{k q^2}{\sigma^2}$
where $k$ is the number of steps in DP-Training, $q$ is the sampling rate (larger for a larger batch size), and $A$ and $B$ are some ``constants'' that hide a (small) dependence on $q$, $\delta$, and clipping norm $C$. 
As expected, $\epsilon$ increases with $k$ at the rate of $\epsilon \approx O(\sqrt{k})$ in a good privacy regime where $k \ll (\sigma/q)^2$ and $O(k)$ otherwise. Increasing the batch size increases the sampling ratio and increases the overall privacy cost while improving the signal-to-noise ratio in average gradients. \footnote{Increasing batch size is one of the most important ways of improving utility of DP-SGD-like methods. To preserve the same privacy, slightly more noise will need to be added; see Figure~\ref{fig:batchsize}.} Moreover, more noise (larger $\sigma)$ means smaller (better) $\epsilon$.

\subsubsection{Reporting Privacy Guarantees for ML Models}\label{sec:dp-gur-practice-reporting}
Works on DP ML vary in details on formal guarantees, often reporting only the $\epsilon$ and possibly $\delta$. We argue that proper reporting requires more information, especially considering the nuances highlighted in Sections \ref{sec:amplification} and \ref{sec:target-eps}, and upcoming in \ref{sec:hyperparams_eps} and \ref{sec-privacy-choices}. We believe that practitioners should report all the following in order to provide a complete picture of the resulting model guarantees and allow for a fair comparison between different methods. 

\begin{specialistbox}{Reporting Privacy Guarantees}
\begin{compactenum}
\item \textbf{DP setting.} For example ``This a central DP guarantee where the service provider is trusted to correctly implement the mechanism''. Or ``This is a local DP that protects data directly when it leaves a user device'' (Section \ref{sec:wheretoapply}).
\item \textbf{Instantiating the DP Definition.} All parts of the abstract DP definition should be clearly mapped to aspects of the concrete application.
\begin{compactenum}
\item \textit{Data accesses covered.} Private data can be accessed for many reasons during the process of building and deploying ML models.\footnote{For example, model architecture search, computation of statistics to understand the data distribution and perhaps inform featurization, training multiple models as part of an architecture search or hyperparameter tuning, as well as training a final model for deployment.} DP guarantees should include a description of which of these data uses are covered and which are not.  E.g., does the DP guarantee apply (only) to a single training run or it also covers hyperparameter tuning (Section \ref{sec:tuning_hyperparams})? 
\item \textit{What the final mechanism's output is.} The formal guarantee is stated in terms of a mechanism (randomized function) $\mathcal{A}$ and the mechanism output(s) $\mathcal{A}(D)$ should be clearly defined.  E.g., only the final model checkpoint is released, however the mechanism's output is technically \textit{the full sequence} of privatized gradients, and the guarantee also applies at this level (all the checkpoints are also protected).
\item \textit{Unit of privacy}, e.g. example-level, user-level, etc (Section \ref{sec:unitofprivacy}). This includes discussing whether protection applies to the \textit{full data} (both labels and features), \textit{labels only} or \textit{predictions only} (Section \ref{sec:wheretoapply}).
\item \textit{Adjacency definition} that was used for ``neigbouring'' datasets --- e.g. add-or-remove, replace one, zero-out one (Section \ref{sec:recordadjacency} and Section \ref{sec:amplification}).
\end{compactenum}
\item \textbf{Privacy accounting details.} 
\begin{compactenum}
\item Type of accounting used: RDP-based accounting, PLD accounting, etc.
\item Accounting assumptions and whether they hold (e.g., Poisson sampling was assumed for privacy amplification but shuffling was used in training).\footnote{In this case, we would recommend also reporting a guarantee that does not utilize amplification.} 
\item The formal DP statement for the final model and for the tuning process.  E.g., the specific $(\epsilon, \delta)$-DP or $\rho$-zCDP values.
\end{compactenum}
\item \textbf{Transparency and verifiability.} When possible, complete open-source code using standard DP libraries for the key mechanism implementation and accounting components should be provided.\footnote{In the future, we hope stronger verification methods perhaps based on secure enclaves and verifiable tool chains will become standard.}
\end{compactenum}

\end{specialistbox}

\subsection{Hyperparameter Tuning} 
In this section we first describe which hyperparameters are important for maximizing the utility of DP models and how hyperparameter tuning can be done in practice, followed up by the techniques to account for such hyperparameter tuning if the original sensitive data was used for this purpose.
\subsubsection{How to Tune the Hyperparameters for DP-Training}\label{sec:tuning_hyperparams}

Several papers study the influence of hyperparameters on the  privacy and utility of the trained model.
In particular,~\citet{kurakin2022dpimagenet} provide a  detailed analysis of how various hyperparameters affect the privacy and utility of convolutional image models on ImageNet.
\citet{li2022llmdp} also discusses hyperparameter tuning in the context of language models and share similar observations with \citet{kurakin2022dpimagenet}. 
Below, we first describe general observations about optimal hyperparameters and then suggest a number of specific algorithms for hyperparameter tuning.

\paragraph{Which hyperparameters are important.}

DP-SGD has two main privacy hyperparameters, the clipping norm $C$ and the noise multiplier $\sigma$, but there are many other training hyperparameters that can drastically affect the utility of the trained classifier. Below is a summary of various hyperparameters and how they affect DP-SGD:

\begin{figure}[h]
\centering
 \includegraphics[width=0.8\textwidth]{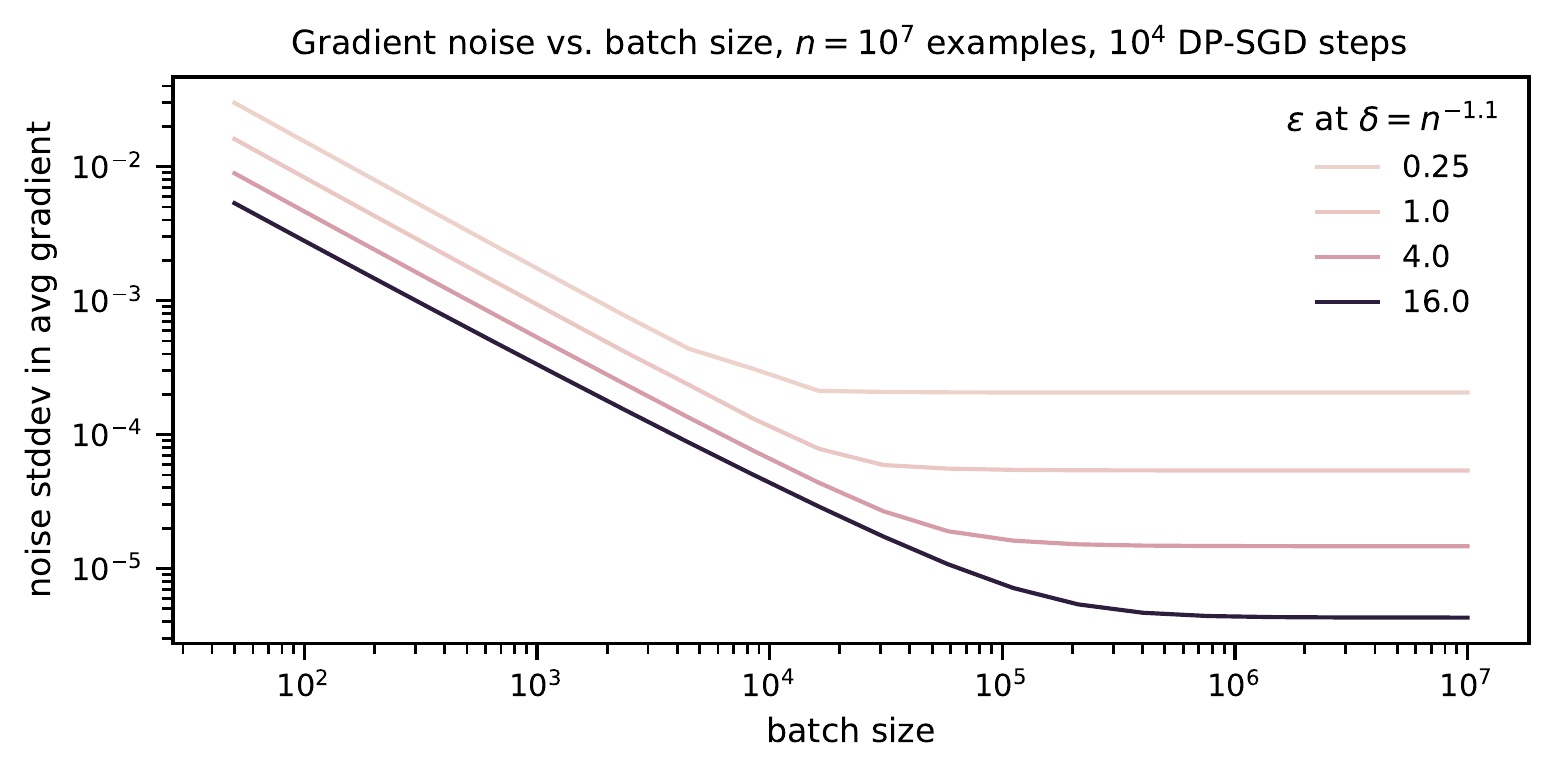}
 \caption{For a fixed dataset size, increasing the batch size can decrease the standard deviation of the noise in the average batch gradient $\bar{g_t}$ almost linearly, up to a point of diminishing returns determined by $\epsilon$, the dataset size, and the number of iterations. Often the point of diminishing returns is at a batch size \emph{much larger} than used for non-private training. This figure shows the tradeoff curves for a dataset of size $10^7$ for $\epsilon$ values from 0.25 to 16, assuming 10,000 steps of DP-SGD training with Poisson sampling, using RDP accounting.  \citet[Fig. 1]{anil2021large} gives a similar figure. The point of diminishing returns for a larger batch size can be approximated by $n \sqrt{\epsilon / T}$.}\label{fig:batchsize} 

\end{figure}

\begin{figure}[h]
\begin{subfigure}{.5\textwidth}
  \centering
  \includegraphics[width=\linewidth]{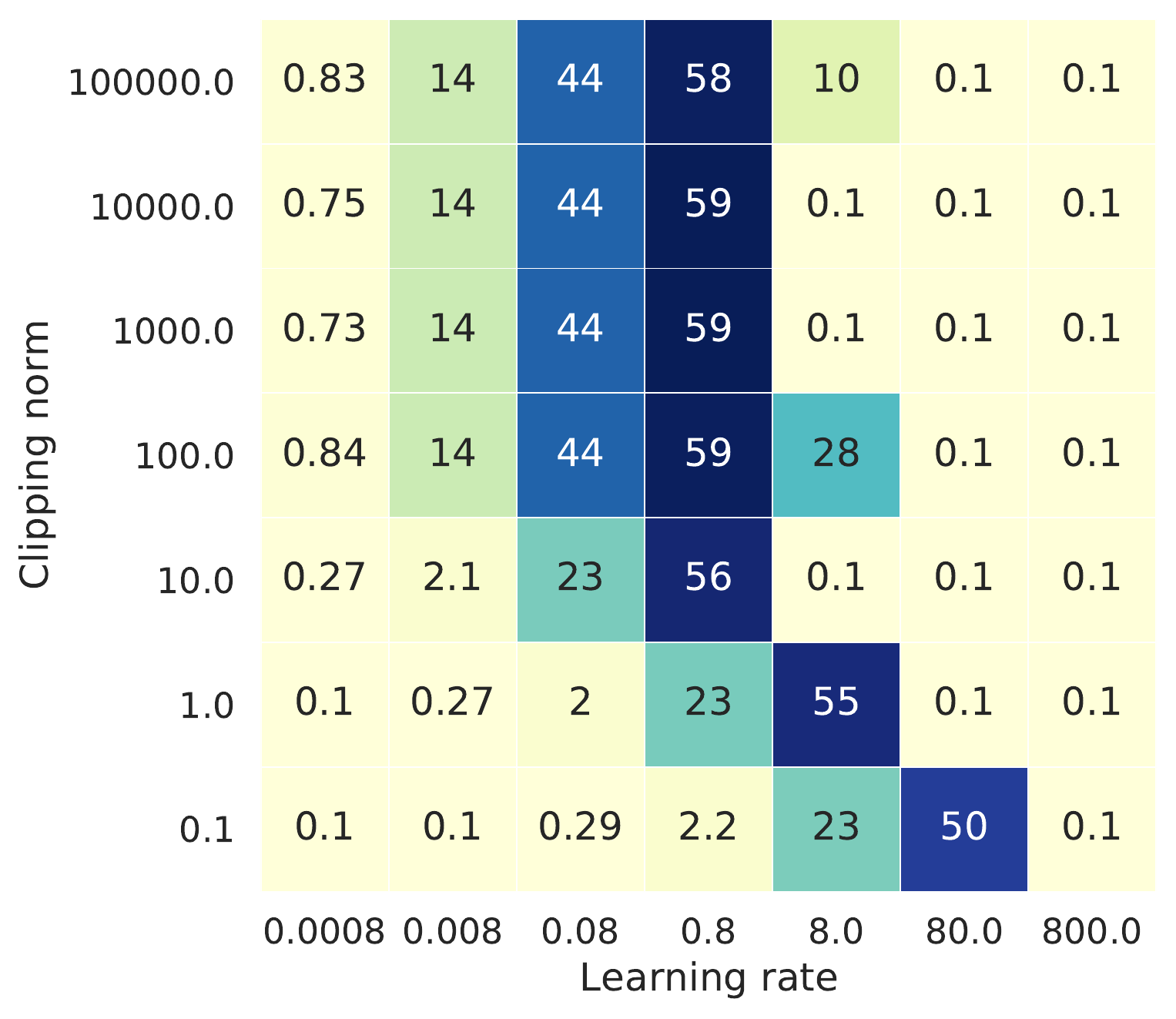}
  \caption{$\sigma = 0$}
\end{subfigure}%
\begin{subfigure}{.5\textwidth}
  \centering
  \includegraphics[width=\linewidth]{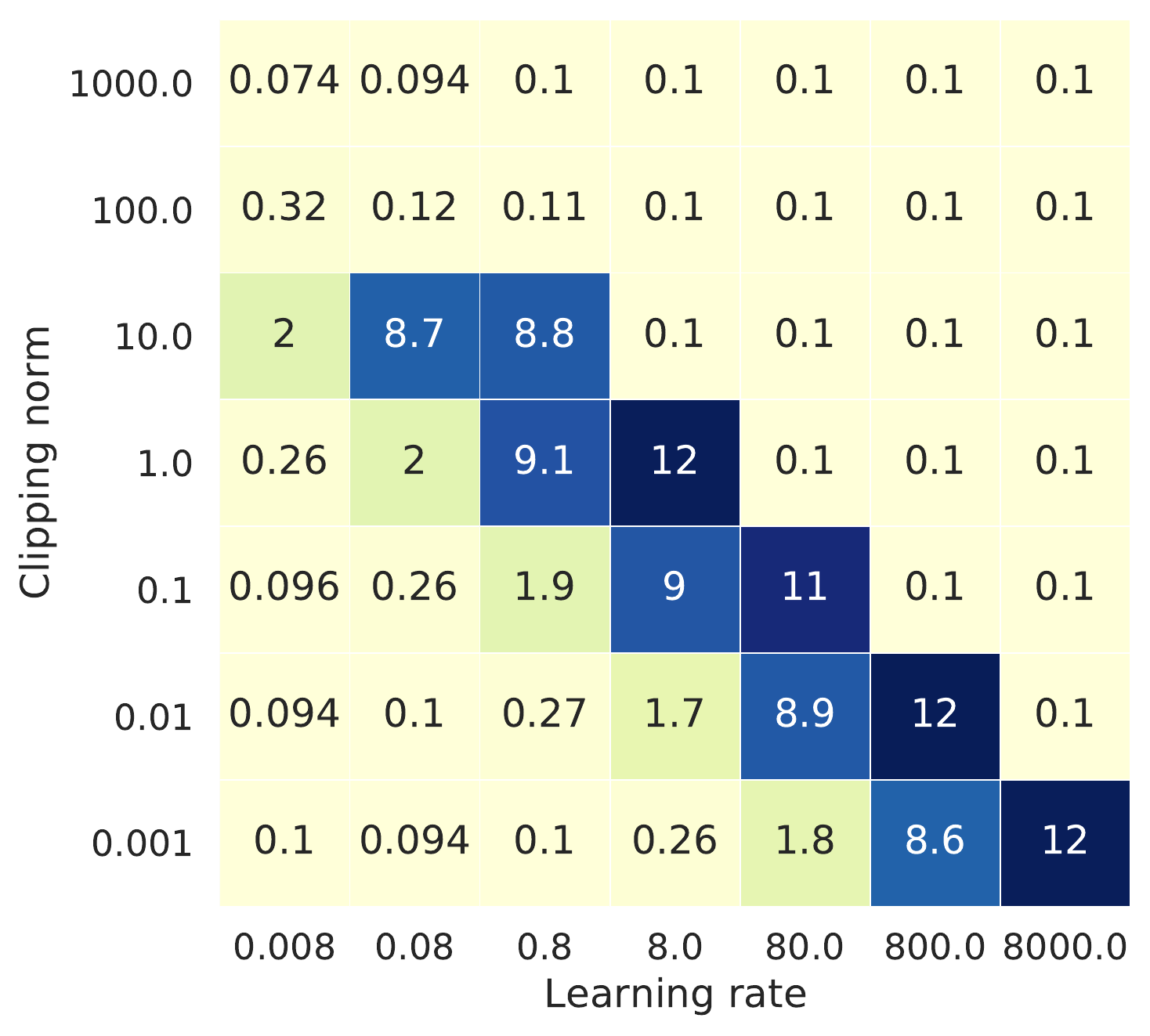}
  \caption{$\sigma \approx 0.28$}
\end{subfigure}
\caption{Relationship between the learning rate and the clipping norm in a non-private (a) and private (b) settings. Values in the grid represent the final training accuracy for an image classification task.
This Figure is adapted from~\cite{kurakin2022dpimagenet}.
In a private case ($\sigma > 0$), the best accuracy is achieved on a diagonal where the product of the clipping norm and the learning rate remains the same. However if the clipping norm gets larger than the norm of gradients, the accuracy quickly drops to zero because the noise would be larger than the magnitude of the gradients. Similar diagonal could be observed in a non-private case as well ($\sigma = 0$). The difference is that in a non-private case the clipping norm could be increased indefinitely. Additionally, the optimal learning rate stays the same once the clipping norm becomes larger than norm of gradients for a non-private case.}
\label{fig:c_lr_sweep}
\end{figure}

\begin{compactitem}
\item \textit{Batch size $B$.} 
For a sufficiently large dataset, increasing the batch size by a factor of $K$ (while keeping the number of DP-SGD iterations constant) will reduce the standard deviation of the noise in the estimate of the average batch gradient ($\bar{g_t}$ in Algorithm~\ref{algo:abadi}) by almost a full $1/K$ for the same the privacy cost $\epsilon$;\footnote{In order to maintain privacy, because the sampling probability $q$ will increase slightly with larger batches, a slightly larger noise-multiplier will be needed when we are in the left-hand regime of Figure~\ref{fig:batchsize}.} see the left-hand regime in Figure~\ref{fig:batchsize}. Hence, for a fixed model architecture and a sufficiently large dataset, it should be possible to increase the batch size and thereby reduce the noise added by DP-SGD to a level that has less pronounced impact on model accuracy. 
Empirically, for a wide range of deep networks, it has been shown that reducing the noise in the average gradient leads to improved model accuracy~\cite{McMahan2018dplm,kairouz21practical,anil2021large,kurakin2022dpimagenet,deepmind2022dpimagenet,choquette22multiepoch}. In convex settings, this observation has been formalized, with \citet{bassily2020stability} and \citet{talwar2014private} showing that larger batches improve utility of DP-SGD, with the best utility achieved by using a  full batch (i.e., batch size equal to dataset size). 
It is important to note that increasing the batch size while keeping the number of iterations constant leads to a corresponding increase in the number of training epochs and hence the total computational cost of model training.

\item \textit{Number of training epochs $N$.}
Even when the batch size is fixed, increasing the number of epochs and therefore the number of DP-SGD iterations (while keeping the same $\epsilon$ by increasing the noise multiplier) is typically beneficial for the utility of private training \cite{kurakin2022dpimagenet}. 
At the same time, there is an effect of saturation when increasing number of epochs beyond a certain point does not seem to help anymore.
\citet{deepmind2022dpimagenet} further showed the existence of an optimal number of training epochs for private training which is significantly larger compared to typical number of training epochs used in non-private setting due to both increasing the batch size and the number of iterations.

It is important to note that training for more iterations does not mean that the practitioner starts the training process and later decides when to stop, because the privacy budget will be increasing with each training step. 
Instead, the practitioner should first fix the total privacy budget and number of epochs $N_{max}$, compute the noise multiplier for DP-SGD (which depends on the privacy budget and $N_{max}$), and then train either for exactly $N_{max}$ epochs or stop early if it allows practitioner to achieve higher utility.

\item \textit{Noise multiplier $\sigma$} is the ultimate factor which is used in privacy analysis to compute $\epsilon$. In addition, increasing the  noise multiplier typically results in a decrease of utility. We recommend setting the noise multiplier after the number of training epochs and the batch size are fixed, based on a desired privacy budget. 

\item \textit{Gradient clipping norm $C$} is another parameter of DP-SGD, and it is used to clip the norm of the gradient of each example. Moreover, the total noise added to the sum of gradients has the standard deviation of $C\sigma$.
As a result, the clipping norm should be typically chosen in a way so that most gradients are either clipped or are near the clipping threshold~ \cite{li2022llmdp}.
If the clipping norm is too high, the noise magnitude $C\sigma$ would exceed the magnitude of gradients, which would make it harder for the algorithm to converge and will adversely affect the utility.
Increasing the learning rate could, to an extent, compensate for a too-small clipping norm, as we describe below.
Nevertheless, if the clipping norm is too small, the model utility would suffer as well.

\item \textit{Other gradient clipping strategies.}
One possibility is to use adaptive clipping instead of fixing the  clipping norm a priori~\cite{andrew2021adaptiveclip}.
However, its implementation is more complicated than static clipping, and thorough tuning of hyperparameters with static clipping norm usually results in the same utility as adaptive clipping \cite{andrew2021adaptiveclip}.

Another clipping strategy is per-layer clipping~\cite{McMahan2018dplm}, in which the clipping norm is set individually for each layer to accommodate different scales of the gradients.

\item \textit{Learning rate $\alpha$} typically has to be tuned to get the
best utility.
In particular, the learning rate has to be re-tuned once a private optimizer is used. \citet{kurakin2022dpimagenet} observed an interesting relationship between the optimal learning rate and the clipping norm. The clipping norm $C$ and learning rate $\alpha$ could be varied in a wide range with no change in the model's utility, as long as the product $C\alpha$ stays constant--see Figure ~\ref{fig:c_lr_sweep}.
An intuitive explanation for this phenomenon can be as follows. Let us say we use clipping norm $C$ and learning rate $\alpha$,
and all gradients are being clipped (i.e. all gradient norms are above clipping threshold $C$). In such a case, if we decrease the clipping norm $k$ times and increase learning rate $k$ times, then the outcome of one step of DP-SGD would remain the same.

It is important to note that in the non-private setting, practitioners commonly use adaptive optimizers (like Adam and Adagrad) and these optimizers are often used without extensive tuning of the learning rate.
Such optimizers generally do work well (e.g., lead to a good training loss stabilization) for a relatively  wide range of the learning rate. 
Nevertheless, the value of the learning rate does have an effect on how fast adaptive optimizers converge. Thus, when the number of training steps is fixed (which is the typical setting in  DP-Training), the utility of the final model can still benefit from tuning the learning rate, even when an adaptive optimizer is used.

\end{compactitem}

Generally, choosing hyperparameters requires optimizing over three inter-independent objectives: 1) model accuracy, 2) privacy cost $\epsilon$, and 3) computation cost (e.g., number of epochs and batch size); tuning strategies will take two of these three as constraints, and focus on optimizing the third subject to those constraints.

Below we summarize all the findings and observations discussed above into several practical strategies which practitioners can use to choose hyperparameters for DP-Training. Firstly we introduce a strategy for determining the optimal clipping norm. This building-block step will be shared by other hyperparameter tuning strategies. Then we demonstrate the first tuning strategy that assumes fixed $\epsilon$ and computational budget and optimizes for the best possible utility. The second strategy instead fixes privacy and utility as constraints and finds the smallest batch size (computation cost) that can achieve these.
This approach usually works well, when a practitioner has a relatively small model, relatively large dataset and unbounded compute.

\paragraph{Choosing an optimal clipping norm.}
No matter what tuning strategy is chosen, a practitioner typically needs to first choose a clipping norm $C$.
The choice of this parameter is essentially the same for all tuning strategies, thus we describe it first as a separate subroutine.

As demonstrated in Figure~\ref{fig:c_lr_sweep} and described previously, there is typically a range of good values of clipping norm $C$ which allows a practitioner to achieve the best utility (assuming that the learning rate is tuned afterwards). The following strategy assumes that we already know good hyperparameter values for a non-private training setting.

\begin{specialistbox}{Strategy to tune clipping norm with zero noise multiplier (\textbf{ClipSearch}).}
\begin{compactenum}
\item Use DP-Training optimizer (e.g., DP-SGD) with the noise multiplier $\sigma$ set to $0$.
\item Run a set of experiments to sweep clipping norm $C$ (with all other hyperparameters fixed).
We recommend to choose a logarithmic scale for the sweep of $C$ such as $\{\ldots, 0.01, 0.1, 1, 10, 100, \ldots\}$ or more fine-grained if resources allow.
\item Identify the \textbf{smallest value} of clipping norm $\tilde C$ such that model utility is (adversely) \textbf{affected only slightly} (as compared to the utility of a non-private model).

If not sure which $C$ to pick, err on the side of smaller $C$.
\end{compactenum}
\end{specialistbox}

The idea behind this strategy is to find a value of clipping norm which causes actual clipping of gradients, which could be observed empirically by a slight drop of utility.
This loss of utility can typically be almost regained by further tuning the learning rate (see Figure~\ref{fig:c_lr_sweep}).
However, to save computational resources it's recommended to tune the learning rate after setting non-zero noise multiplier $\sigma$.

If running a clipping norm sweep is too costly, a practitioner may consider using adaptive clipping norm algorithm~\cite{andrew2021adaptiveclip} which should save the compute at a cost of more complicated implementation.

\paragraph{Tuning given privacy and computation constraints.}

This strategy assumes that a practitioner has a specific $\epsilon$ target in mind. 
As discussed in Section~\ref{sec:hyperparams_eps}, it is important to decide whether this privacy budget applies only to the training of the final model, or needs to cover potential privacy losses during hyperparameter tuning on private data as well. In any case, the hyperparameter tuning process should be reported (please see Section \ref{sec:hyperparams_eps} for an additional discussion).

Additionally, we assume that the network architecture and data preprocessing pipeline are fixed a priori.
We also expect that the practitioner has a way to choose good hyperparameters for a non-private version of the model.
This is typically the case when practitioner starts with some non-private model with the goal of making it differentially private.
With that in mind we condense all above-mentioned considerations into the following strategy:

\begin{specialistbox}{Hyperparameter tuning strategy under computation and privacy constraints}
\begin{compactenum}
\item \textbf{Identify the maximum number of training epochs $N$ and the largest batch size $B$ that is computationally feasible.}
Typically a practitioner should start with batch size and number of training epochs which is used for non-private training and then simultaneously scale both of them until computational limit is reached.

\item \textbf{Tune the model in a non-private setting with chosen $N$ and $B$.} Identify the optimal learning rate $\alpha_{nodp}$ and possibly other hyperparameters,
like weight decay, regularization and so on.
To simplify the tuning, all non-private hyperparameters other than the learning rate are considered frozen after this step. 

\item \textbf{Choose the clipping norm $C$ using the subroutine ClipSearch}

\item \textbf{Compute noise multiplier $\sigma$} based on desired privacy budget $\epsilon$, batch size $B$ and number of training epochs $N$.
\item \textbf{Perform the learning rate sweep.} Set the noise multiplier to $\sigma$, clipping norm to $C$ and run a full search of a learning rate (for example, using grid or random search). Additionally, if it is computationally feasible, add a concurrent sweep over the vicinity of the clipping norm $C$ chosen previously.
\end{compactenum}
\end{specialistbox}

Due to resource constraints, it might be hard to perform the multiple hyperparameter sweeps suggested above.
The following heuristics can reduce the computational burden:
\begin{compactitem}
\item Tune the hyperparameters on a smaller model with similar architecture and then re-use most of these hyperparameter value for the final large model training.
\item Tune the hyperparameters on a smaller batch size; then linearly increase noise multiplier and the batch size. This approach can be potentially useful when attempting to meet a constraint on both privacy and utility, as discussed next. 
\end{compactitem}

\paragraph{Tuning given privacy and utility constraints.}
As discussed previously, good privacy and utility can sometimes be achieved by choosing a sufficiently large batch size (at the cost of increased computation); this is likely to be possible in a setting where non-private models with reasonably high accuracy were trained using only a fraction of the dataset, which is relatively common in production settings. However, the required batch size might make training even a single model quite computationally expensive, and hence alternative approaches to hyperparameter tuning may be required.

Building on the suggestions above, the following assumption (introduced by \citet{McMahan2018dplm}) can be quite useful: \emph{for a sufficiently large batch size $B$, the accuracy of the model is essentially determined by the standard deviation $\bar{\Sigma} := \frac{\sigma C}{B}$ of the noise in the average model gradient $\bar{g_t}$} (using the notation of Algorithm~\ref{algo:abadi}). Importantly, we assume the number of training iterations and other hyperparameters including $C$ remain fixed.  That is, even though the privacy cost would be quite different, we assume we get the same accuracy whether we choose $(\sigma, B)$ or $(K\sigma, K B)$ for any multiplier $K$. Thus, we may estimate test-set accuracy as a function of $\bar{\Sigma}$ using a small batch size (and other hyperparameters) that provide good accuracy for non-private training; \citet[Fig.3]{McMahan2018dplm} provides an example of such a relationship; there is a clear ``knee'' in these curves, indicating that as long as noise is below a certain threshold, model accuracy is essentially unaffected. Of course, other hyperparameters like the learning rate and number of iterations may influence these curves. 

With this relationship established, one can choose a $\bar{\Sigma}^*$ that achieves the desired accuracy, and then use an accounting line search\footnote{This is equivalent to plugging $\bar{\Sigma}^*$ into the $y$-axis of Figure~\ref{fig:batchsize}, reading off the corresponding $B$ for the desired privacy level, and then computing $\sigma = B \bar{\Sigma}^*/C$.} (like \texttt{dp\_accounting.calibrate\_dp\_mechanism} from the Google DP libraries, see Section~\ref{sec:calculating-and-reporting-eps}) to find a $B$ and $\sigma$ such that a desired privacy target is achieved while satisfying $\frac{\sigma C}{B} = \bar{\Sigma}^*$. This approached can be summarized as:

\begin{specialistbox}{Hyperparameter tuning strategy under privacy and accuracy constraints}
\begin{compactenum}
\item Identify a (small) batch size $B_{small}$ and learning rate that gives reasonable model utility in a non-private setting.
\item Identify an appropriate clipping norm $C$ using the subroutine ClipSearch.
\item Varying the noise level $\sigma$, plot model utility vs noise in the average gradient $\bar{\Sigma} := \frac{\sigma C}{B_{small}}$.
\item Identify the smallest noise level $\bar{\Sigma}^*$ that achieves the desired utility (if possible).
\item Holding $\epsilon$ fixed and varying batch size $B$ and $\sigma$ subject to the constraint $\frac{\sigma C}{B} = \bar{\Sigma}^*$, find the appropriate level of noise $\sigma$ and (hopefully computationally feasible) batch size $B$ that are estimated (via $\bar{\Sigma}^*$) to achieve the desired utility.
\item Train the final model using batch size $B$. If computation allows, consider trying several slightly smaller learning rates as well.
\end{compactenum}
\end{specialistbox}

The success of this approach depends on whether the above assumption holds. The primary concern is the now-well-known ``generalization gap'' phenomenon in non-private training, where using too large of a batch size may harm the model's generalization (test set accuracy), even though train-set accuracy may be unaffected or improved \cite{keskar17largebatch,hao18visualizing}. However, both papers point out that the stochastic noise in SGD gradients (due to sampling a batch of examples) may be important to this phenomenon, and hence one might conjecture that adding additional noise as in DP-SGD should offset the over-fitting tendencies of large-batch training. Further, \citet{hoffer17trainlonger} observes that the generalization gap may be due to keeping the number of training epochs constant, and that it is the decrease in the number of iterations due to larger batches (when epochs are fixed) that is problematic, rather than the batch size itself. The fact that learning rates may need to be adjusted when changing batch sizes (particularly if the number of iterations is also changed) further complicates the situation.
More work in this area is certainly needed, particularly with respect to determining the significance of the ``generalization gap'' phenomenon to DP training. In any case, the tuning assumption above has proved useful, with both \citet{McMahan2018dplm} and \citet[Fig. 12]{kairouz21practical} verifying that this assumption holds for next-word-prediction language models in the regimes considered. We encourage practitioners to try this approach, and report if large batches produce worse generalization than predicted by small-batch experiments; however, some caution is advised if it is known that lager batches (with fixed iterations) can reduce generalization for non-private training.

\paragraph{Large Language Models Peculiarities.\Difficult}

For giant models like Large Language Models (LLMs), it might be impossible to implement the tuning procedure outlined above. In particular, such models may take long time to converge (e.g., days), and running a sweep over hyperparameters is prohibitively expensive. A common strategy for tuning such models is to reuse the optimal hyperparameter values found on small models for training larger models \cite{yang2021tuning}. Although some papers, such as \cite{yang2021tuning},  state that regularization and optimizer parameter values might not be transferable, \citet{li2022llmdp} reports success finding appropriate hyperparameters on smaller GPT models and reusing them for larger GPT models. Another peculiarity of LLMs is that some implementations like \citet{roberts2022t5x} by default do not normalize the loss to account for the length of the sequence. This results in large gradient norms (that depend on sequence length). This in turn makes it hard to find an appropriate clipping norm and can give preference to longer sentences. \citet{ponomareva-etal-2022-training} suggests for DP-training to use a loss normalization strategy that averages the loss incurred over all target tokens in the target sequence.

Finally, giant models require a lot of training data. While some authors \cite{ponomareva-etal-2022-training,anil2021large} were able to fully pre-train LLMs with DP, by far the most common strategy is to take some pre-trained LLM checkpoint and do only private fine-tuning on the private data. For example, \citet{li2022llmdp} showed that private fine-tuning can maintain accuracy given a good pretrained model. \citet{hoory-etal-2021-learning} similarly explored DP fine-tuning on a medical domain for a pre-trained BERT checkpoint. However, \citet{li2022llmdp} highlight that for such a strategy to work well, the pretraining and fine-tuning tasks should have similar objectives.

\paragraph{Peculiarities of Models with Very Sparse Gradient Updates.\Difficult}

Certain layer types in a deep learning model can generate very sparse gradients. As an example, if a lookup table is used to encode categorical variables into an embedding space, then per-example gradient updates for all but one row of the embedding table will be exactly zero. 
When using a model with very sparse gradient updates, it is important to use an optimizer that keeps gradient updates in sparse form: if the optimizer stores and passes only non-zero parameter updates, the computation and memory cost of many operations becomes smaller. 
Since these operations are cheaper with sparse updates, one can use much larger batches than an optimizer which materializes the entire gradient for each sample. This can significantly improve the computational speed of the model and reduce the memory footprint. 
In turn this results in an improved utility of the model both due to the increase of the possible batch size and due to the ability of training for more epochs (using the same amount of computational resources). Finally, it is important to make sure that the sparsity-aware optimizer is implemented in such a way that the noise is nevertheless applied to \textit{all} the weights, even the ones with zero gradient. If this is not done, the differential privacy guarantees will not hold.

\subsubsection{How Hyperpameter Tuning Can Increase $\epsilon$}\label{sec:hyperparams_eps}
\begin{specialistbox}{Privacy cost of hyperpameter tuning}
The simplest approach, recommended when possible, is to do all model architecture search and hyperparameter tuning on a proxy public dataset (with a distribution similar to the private data), and only use the private training dataset to train the final DP model. When hyperpameter tuning must be performed on private data, any privacy guarantees reported should clearly state what ``data touches'' are accounted for (Section \ref{sec:dp-gur-practice-reporting}). This should include at least a guarantee that applies only to the use of private data in training the final model, but ideally also a (weaker) guarantee that accounts for the use of private data in hyperparameter tuning.
\end{specialistbox}

In the remainder of this section, we expand on the context for this recommendation, and touch on tools that allow us to formally account for hyperparameter tuning costs.

As discussed in the previous sections, it is generally recommended to tune multiple hyperparameters of the model. Ideally, DP guarantees should account for all uses of the private data that influenced anything being released. In practice, this can be difficult to formalize for the long-lived, evolving datasets that are common in industry (with users signing up or leaving the system, and/or updating their data). For example, it is likely impossible to precisely account for the hypothetical privacy cost of using a set of initial hyperparameters selected as ``probably pretty good'' by an ML engineer with years of experience working with the dataset in question.

Nevertheless, when hyperparameter tuning using private data is undertaken more methodically, it can be possible to account for the privacy cost of specific hyperparameter tuning runs. At least in theory, this may be important, as doing hyperparameter tuning on private data without accounting for the privacy cost can inadvertently leak private information. For example, recent work \cite{DBLP:conf/iclr/Papernot022} shows that the choice of hyperparameters can reveal sensitive data of outliers in small SVM models, unless each hyperparameter trial was trained with DP-Training. Although such attacks exist in theory, the most important thing is to train the final released model with DP. 

In this section we aim to show that more rigorous accounting of the hyperparameter tuning process is also possible. There are two general classes of strategy for tuning hyperparameters while preserving DP. 

\paragraph{Using public data.} The first class is to tune hyperparameters by training models on publicly available data which is drawn from a similar distribution as the private data of interest. This does not carry an additional privacy cost and is a reasonable first choice when such data is available. In the absence of public data, an alternative is to fix the values of hyperparameters to some reasonable defaults and forgo hyperparameter tuning altogether. This approach is often used with extremely large models like language models, where the (compute) cost of tuning is prohibitively expensive \cite{ponomareva-etal-2022-training}. 

\paragraph{Tuning on private data.}
The second class is to train each model (with DP-SGD or other DP-Training method) during the hyperparameter sweep and account for these runs in the final privacy guarantees. There are several accounting methods in the literature. The simplest way to account for the privacy cost of multiple hyperparameter tuning runs is using sequential composition, i.e.,  simply adding the individual $\epsilon$ and $\delta$ costs from each of the runs. However, these bounds can be significantly improved due to the fact that we train many models during hyperparameter tuning but only the best model is released and used for inference.
Below we list several ways to account for the privacy cost of a set of hyperparameter tuning runs. The relative utility of these approaches is dependent on the specifics of the task and the number of hyperparameter runs desired or needed. For a specific problem, one should compute the privacy costs with each accounting method and use the one that provides the best privacy-utility trade off.

\begin{figure}[h]
\centering
\begin{subfigure}[t]{0.03\textwidth}
    \textbf{A}
\end{subfigure}
\begin{subfigure}[t]{0.45\textwidth}
    \includegraphics[width=\textwidth]{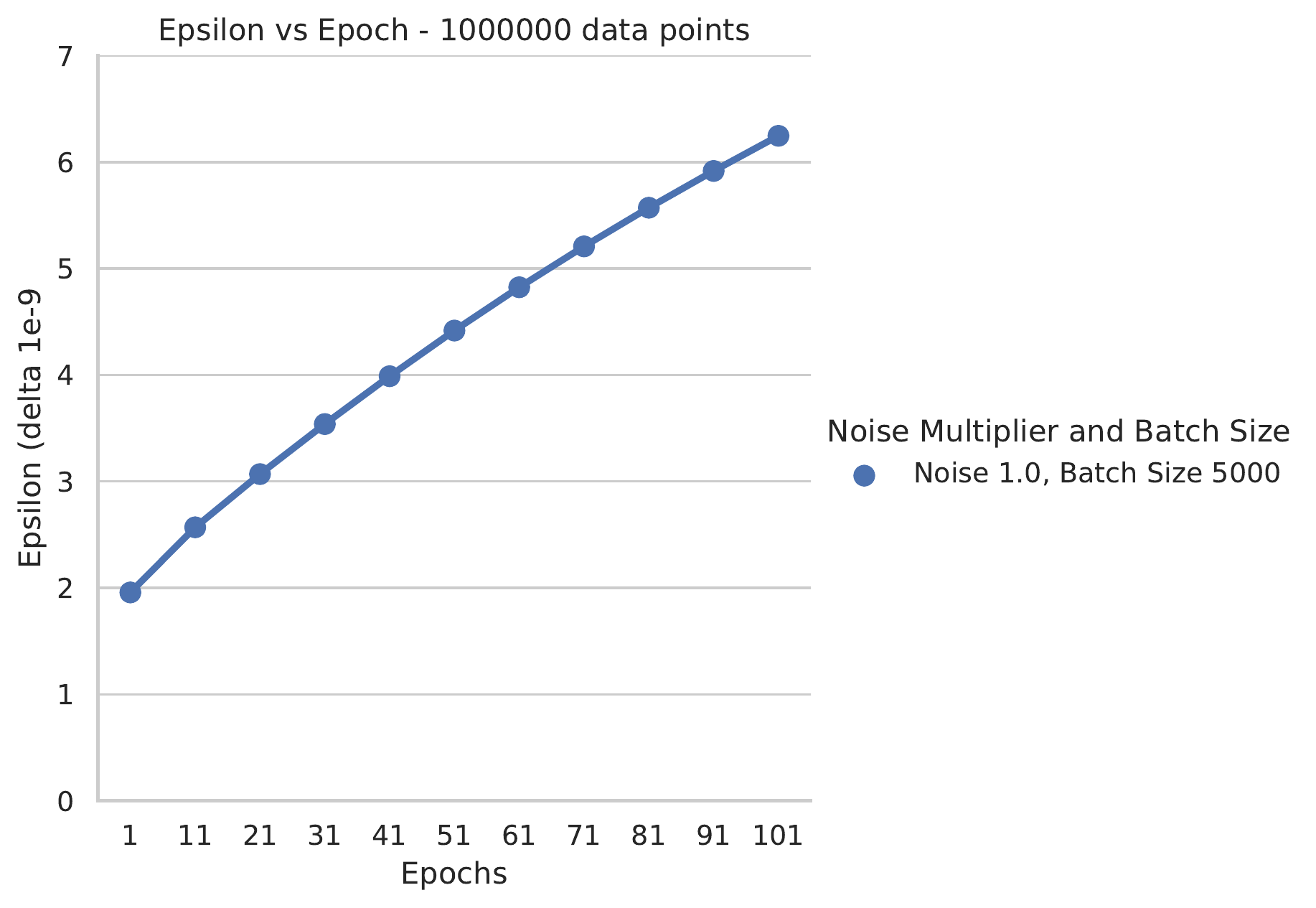}
\end{subfigure}
\begin{subfigure}[t]{0.03\textwidth}
    \textbf{B}
\end{subfigure}
\begin{subfigure}[t]{0.45\textwidth}
 \includegraphics[width=\textwidth]{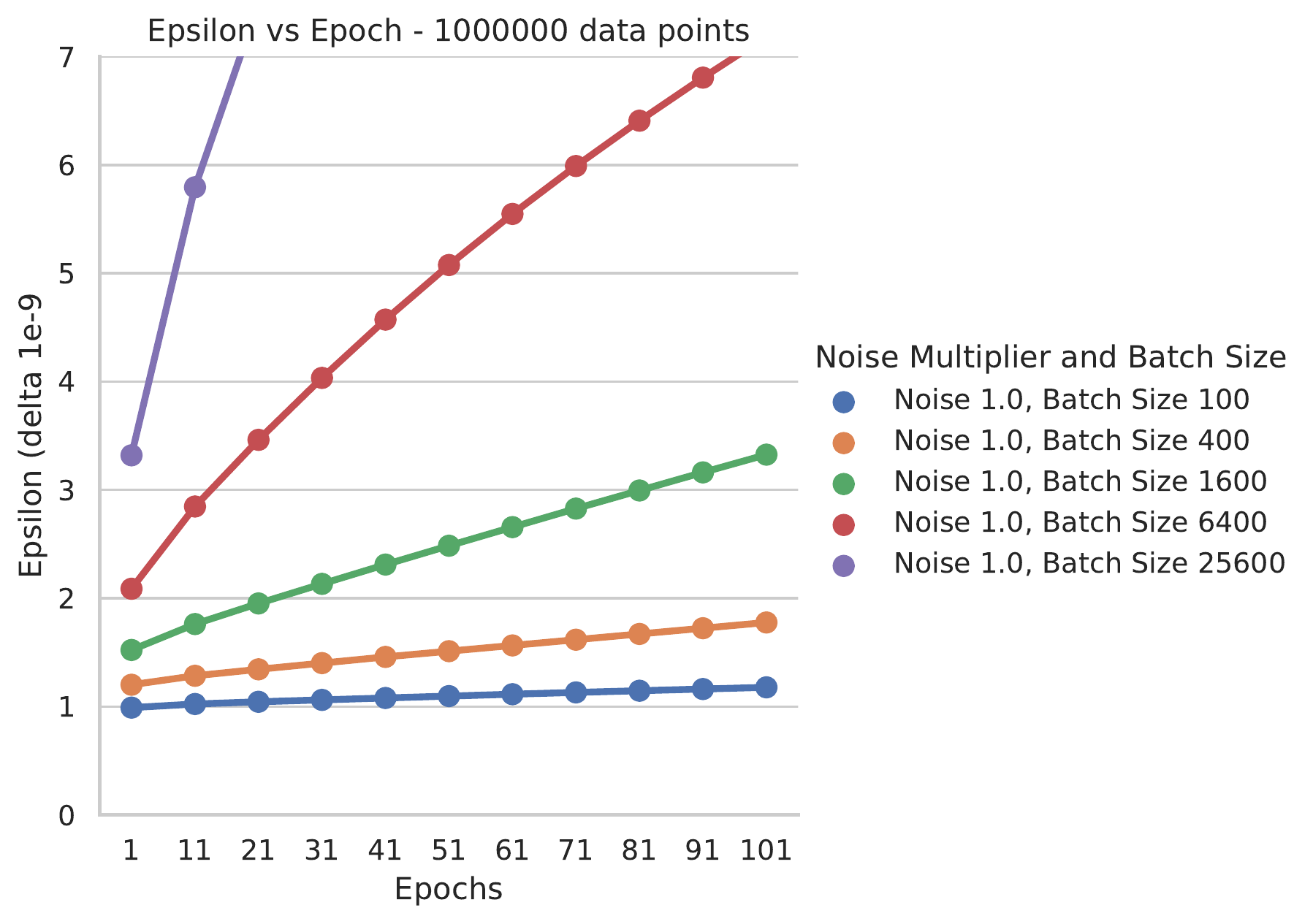} 
\end{subfigure}

\caption{\textbf{A} Privacy costs when treating each hyperparameter  tuning run like an extra epoch. The first epoch has a high privacy budget cost, and each subsequent epoch is "cheaper". 
\textbf{B} Training with smaller batches has lower privacy cost for the same total number of epochs.}
\label{figure:epsilon-vs-epoch}
\end{figure}

First, improvement over sequential composition can be achieved by treating each hyperparameter tuning run as an additional epoch and computing with RDP the cost of training for the total number of epochs used across all hyperparameter tuning runs. This approach makes the privacy cost for the first epoch significantly higher than the cost of an additional epoch. This is due to the sub-linear scaling introduced by either advanced composition or R\'enyi Differential Privacy \cite{dwork}. For the Figures below, we use R\'enyi differential privacy because it provides a tighter bound. In Figure \ref{figure:epsilon-vs-epoch}, we show the privacy costs for various numbers of epochs and various hyperparameters\footnote{We open source the code used to generate these plots at the following URL: \url{https://gist.github.com/carsondenison/6ca3890e3231de9be461cc04510e962e}}. As discussed in Section~\ref{sec:dp-gur-practice}, the functional form of the privacy cost $\epsilon$ is 
$$\epsilon \approx A \dfrac{q \sqrt{k}}{\sigma} + B \dfrac{k q^2}{\sigma^2}$$
where $k$ is the number of steps in DP-Training, $q$ is the sampling rate and $A$ and $B$ are some ``constants'' that hide a (small) dependence on $q$, $\delta$, and clipping norm $C$ (refer to Appendix \ref{app:epsilon_epoch_derivation} for derivations). Notice in Figure \ref{figure:epsilon-vs-epoch} (B) that using smaller batch sizes has significantly lower privacy cost for the same number of epochs. Several papers, such as \citet{anil2021large} and \citet{TAN_without_a_burn}, recommend using small batches for a fixed number of steps for hyperparameter tuning, then training the final model with the largest possible batch size, given the computational resources, and adjusting the learning rate, given the computational budget, to maximize the privacy-utility trade-off.

A second approach is to use the notion of Privacy Loss Distributions (PLD) \cite{koskela2020tight}. Privacy loss distributions offer tighter composition of DP events than R\'enyi differential privacy. The approach is similar to that of RDP accounting described above, except it exactly tracks the privacy loss distribution.

A third approach to account for hyperparameter tuning cost was proposed by \citet{Abadi_2016}. On a high-level, this approach randomly samples a set of $M$  configurations from a search space consisting of $K$ hyperparameter configurations. Then a model is DP-Trained on each sampled configuration. The best model among these trials is consequently released in a differentially private manner using the exponential mechanism \cite{dwork}.
This approach presents a triple trade-off between privacy, number of hyperparameter configurations explored, and the expected utility of the released model. 
Specifically, the larger the search space size $K$, the lower the privacy cost and the more hyperparameter configurations $M$ may be tried. However, the accuracy of the exponential mechanism degrades with increasing $K$, meaning that potentially a model other than the best is selected as the winner. 
Refer to appendix D of \citet{Abadi_2016} for the algorithmic bounds.

Recently \citet{DBLP:conf/iclr/Papernot022} provided tight bounds for their hyperparameter tuning algorithms using RDP accounting. In this family of algorithms, the number of hyperparameter tuning runs $M$ is treated as a random variable drawn from a chosen probability distribution. 
One then draws $M$ sets of hyperparameters to test, selected randomly with replacement, from a set of $K$ potential hyperparameters. 
Finally, the true best set of hyperparameters is returned.
If $M$ is drawn from a Poisson distribution, the authors show that the cost is only logarithmic in the number of tuning runs. Alternatively, if $M$ is drawn from a truncated negative binomial distribution, then the privacy cost increases very slowly with the number of tuning runs. 

Table \ref{table:tuning-comparison} summarizes aforementioned methods and presents their asymptotic bounds. In Appendix \ref{app:hparam-tuning-comparison} we compare the privacy costs of the above methods for 100 hyperparameter tuning runs of a dataset with $1,000,000$ data points, a batch size of $5,000$, and a noise multiplier of $1.0$. We find that using the PLD accountant to compute the single-epoch cost, and using the Poisson-distribution based method from \citet{DBLP:conf/iclr/Papernot022} gives the best trade off between privacy and reliability in this particular setup.

Finally, no matter which hyperparameter tuning strategy is used for privacy accounting, it is important to remember that differentially private hyperparameter tuning is in its infancy. 
Additionally, it is common to use some form of iterative Bayesian hyperparameter tuning which creates interdependent hyperparameter trials \cite{https://doi.org/10.48550/arxiv.1206.2944}. When each run is chosen adaptively based on the results of previous hyperparameter trials, advanced composition, RDP composition, and PLD composition are still valid. However, the tighter bounds from \citet{DBLP:conf/iclr/Papernot022} and \citet{Abadi_2016} no longer hold. Tighter accounting of adaptively chosen hyperparameters is an open area of research.

\begin{table}[ht]
\small
\begin{tabular}{p{1.8cm} p{4cm} p{1.8cm} p{2.6cm} p{3.2cm}}
 \toprule
\textbf{Tuning Strategy} &\textbf{Method} &\textbf{Privacy Cost} &\textbf{Accuracy (w.r.t. best trial)} &\textbf{Additional Considerations} \\
 \midrule
 \multirow{2}{2cm}{Not using Private Data} & Use default hparams &0  &Best &Requires good defaults. Can be useful when computational cost of hparam tuning is too high\\
 &Tune on public data &0  &Best   &Works only if public data of similar distribution is available\\
 \hline
 \multirow{4}{2cm}{Using private data} &Simple composition &$\mathcal{O}$(trials) &Best &Strictly dominated by advanced composition\\
 &Advanced composition or Renyi DP composition or PLD composition & $\mathcal{O}(\sqrt{\text{trials}})$ &Best & $\mathcal{O}(\sqrt\text{trials})$ when steps * sampling ratio $<<$ noise multiplier, $\mathcal{O}(\text{trials})$ otherwise\\
 &Using Exponential mechanism \cite{Abadi_2016} &$\mathcal{O}(\sqrt{\text{trials}})$  &$\text{Best} - \mathcal{O}(\sqrt{\text{trials}})$ &Known number of trials, but best run is not always returned.\\
 &Randomized Number of Trials \cite{DBLP:conf/iclr/Papernot022} &$\mathcal{O}(\log{\text{trials}})$ &Best & Number of trials to run is randomized, but best run is always returned.\\
 \hline
\end{tabular}
\caption{Comparison of Hyperparameter Tuning Strategies}
\label{table:tuning-comparison}
\end{table}

\subsection{Model Architecture Considerations}\label{sec:architecture}
In general, a number of architectural choices should be made in order to successfully apply DP-Training. In Section \ref{sec-privacy-choices}, we go over model components that are potentially not private and need to be modified for use with DP-Training. Then, in Section  \ref{sec:utility-considerations} we describe components that affect model utility when DP-Training is used. 
\subsubsection{Model Components Which Affect Privacy}\label{sec-privacy-choices}
It is worth remembering that DP-Training (e.g., DP-SGD) and its variants provide guarantees by limiting the impact that each individual instance has on the overall model. 
Each instance is considered to be sampled independently with some fixed probability (defined indirectly by the ratio of the batch size to the overall dataset size). 
The contribution of each sample to the overall model is limited (by applying per-example gradient clipping) and the final aggregated gradient is further distorted by adding random noise.
These steps result in bounded per-sample contribution and allow one to reason about privacy guarantees of the overall model. 
However, some components and architectural decisions that are commonly used in neural networks may break this reasoning of limited per-example contribution. 
For example, layers that calculate and/or store some batch statistics like batch normalization layers, or losses that can't be decomposed into per-example losses, such as pairwise losses. 
Some libraries like Opacus in PyTorch \cite{opacus} chose to disallow usage of such components and require users to replace them with components that don't calculate batch statistics. 

Below, we briefly show how to reason about several popular layers that are commonly used. Then, we proceed with a table listing additional commonly used neural networks components (layers, optimizers, tokenizers, etc.) and state whether they are inherently private when used with DP-Training or need to be modified. This table is not meant to be exhaustive, but rather to highlight that all parts of a complex ML model should be examined.

\begin{compactenum}
\item \textbf{Batch Normalization} layer \cite{ioffe2015batch}
normalizes each layer's input to zero mean and unit variance by rescaling the batch using its calculated on the fly batch mean and variance. During inference, the batch mean and variance  are fixed to that of the exponential running average of training mean and variance. 

A BatchNorm layer has trainable parameters that are updated during backpropagation. These parameters don't represent a problem from a DP standpoint, as long as DP-Training method like DP-SGD was used during backpropagation. However, BatchNorm uses current batch' mean and standard deviation information to rescale each instance in the batch during the forward pass. This creates dependency between instances from the batch and makes it hard to reason about per-example sensitivity for DP-SGD.

In order to make BatchNorm private, one has several options:
\begin{compactenum}
\item In settings when public data is available, one can instead calculate the mean and variance BatchNorm statistics on this public data that is injected into the batch during the training, as in \cite{davody2020effect}. To obtain such public data, authors use data close in semantics: for example, KMnist data as public data for MNIST, CIFAR100 for Cifar10 data etc.. During inference, statistics over the same public data are used.
\item It is possible to privatize BatchNorm per-batch mean and standard deviation calculation. For example, to privatize the mean, per example clipping (with a norm different from DP-SGD clipping norm) and Gaussian noise addition to the sum can be used. Such privatized batch mean then would be employed during the forward pass, as well as for updating running mean statistics that BatchNorm layer maintains and subsequently uses for inference. Privacy accounting for a sequential combination of Gaussian Mechanism for a BatchNorm and Gaussian Mechanism for DP-SGD can be handled for example via accounting for adaptive streams \cite{denisov22matfact}.
\end{compactenum}

\item \textbf{Layer Normalization} is another popular normalization layer that improves training time. It removes the dependency on the batch size that BatchNormalization exhibits, and it can be used, unlike BatchNorm, for recurrent nets \cite{layernorm}. This layer computes means and variances for all the neurons in the layer using only one instance (as opposed to the whole batch in BatchNorm). Because this normalization works on a per-instance basis, and because means and variances of all neurons will be public if neurons were updated via DP-Training process, this layer poses no problems from DP-standpoint.

\item \textbf{Group Normalization} is a normalization layer introduced by \citet{groupnorm} and specifically designed for vision tasks. This layer is essentially a LayerNormalization applied to groups of channels from the image input and is equivalent to LayerNormalization when the number of groups is 1. Just as LayerNormalization, this layer does not pose privacy concerns.

\item \textbf{WeightNormalization} \cite{DBLP:journals/corr/SalimansK16} is another normalization layer that does not introduce dependencies between examples from the same batch. It works by reparameterizing the weight vector by decoupling its direction from its length, with these two new parameters (a scalar and a vector) learnt via gradient descent. As long as the gradients w.r.t. to all parameters are appropriately privatized as per DP-SGD, this layer does not pose problems.
\end{compactenum}
We refer readers to Table \ref{layers-privacy-table}, which  highlights other commonly used components or processes that apply to NNs. 

\renewcommand{\arraystretch}{1}
\setlength{\tabcolsep}{5pt}
\footnotesize
\begin{longtable}[ht!]{p{0.01\linewidth} p{0.3\linewidth} p{0.05\linewidth} p{0.55\linewidth}}
 \toprule
\textbf{}  & \textbf{Name} & \textbf{RM\footnote{Requires \textbf{additional} modification (beoynd clipping and noise as per DP-SGD Algorithm \ref{algo:abadi}) to be DP-Compatible}} &\textbf{~Comments}                                         \\
\midrule

\multirow{5}{*}{\rotatebox[origin=c]{90}{Normalizing}} 

&Batch Normalization \cite{ioffe2015batch}  & Yes & Either add noise to Batch norm mean and mean of squares calculation; or use public data to calculate these statistics. \\

&Layer Normalization \cite{layernorm}  & No &  \\
&GroupNormalziation \cite{groupnorm}  & No &  \\
& Weight Normalization \cite{DBLP:journals/corr/SalimansK16} & No \\
\hline

\multirow{13}{*}{\rotatebox[origin=c]{90}{Specialized Layers}} 

&GNN \cite{4700287} & Yes & Node or edge level GNN layers augment the features of the instance (node or edge) with features and labels of their direct neighbours, making this process not private. Further, graph structure is also leaked through such aggregation \cite{DBLP:journals/corr/abs-2010-00906}. An additional complication of GNN networks is that during inference time, the same (training) graph structure is reused for predictions, and it needs to be DP-protected as well  (on top of DP-Training of GNN models) \cite{DBLP:journals/corr/abs-2111-15521}. There are attempts at adding noise to the aggregation function, but the authors also had to change the structure of the network \cite{DBLP:journals/corr/abs-2111-15521}, while \citet{10.1145/3460120.3484565} considered nodes and labels private but treated edges as public data.The area of DP with GNN models is very much nascent. \\
\hline

\multirow{20}{*}{\rotatebox[origin=c]{90}{Optimizers}} 
& SGD \cite{Robbins2007ASA}  & No & As long as DP-SGD version is used \cite{Abadi_2016}, which dictates per example clipping and adding noise to the aggregated batch gradients  \\
&Adaptive and Accelerated First-order Optimizers (Adam \cite{adam}, Adagrad \cite{JMLR:v12:duchi11a} etc)  & No &
These optimizers maintain additional statistics (e.g., momentum), which are only functions of the gradients obtained by the optimizer at current and previous steps. Thus, as long as the gradients are accessed using a DP mechanism (e.g., per \cite{Abadi_2016}), adaptive and accelerated first-order methods are DP. However, recent research suggests that DP versions of adaptive optimizers can accumulate extra noise, which may negatively affect utility \cite{li2022private}. Designing adaptive optimizers for DP-Training is an active area of research
\cite{kuru2022differentially,asi2021private,li2022private,kairouz2021nearly}. Refer to Section \ref{sec:utility-considerations} for additional discussion.
\\

&Second-order Optimizers (e.g., Newton) & Yes & These optimizers compute the Hessian  of the loss (or an approximation of it) and use it to rescale the gradients. To ensure differential privacy, both the gradients and the Hessian should be accessed using a DP mechanism. The gradients can be privatized using the same techniques used in first-order methods. However, privatizing the Hessian is more challenging. For empirical risk minimization with convex objectives, one possibility is to add noise to the per-example Hessians (technically, this approach requires the Hessian to have a known, bounded norm) \cite{avella2021differentially}. However, the latter approach can be computationally prohibitive for large problems (since each per-example Hessian scales quadratically in the number of parameters) and it can also lead to potential convergence (e.g., loss stabilization) issues\footnote{After adding noise, the Hessian matrix may no longer be positive definite, which may negatively affect convergence.}. As a more efficient alternative, \citet{mehta2022differentially} clips the input features (instead of the per-example Hessians) and then adds calibrated noise to the (full) Hessian. When sufficient public data is available, another possibility is to use the public data to estimate the Hessian (while using private data to compute the noised gradients) \cite{ji2014differentially}.
\\

& Sparsity-aware Optimizers & Maybe & Some (implementations of) optimizers are aware of the fact that only a small proportion of the weights will have nonzero gradients. This is often the case for models with large dimensional embedding/lookup tables. While clipping is not affected by the sparsity (e.g., we need to clip only non zero gradients), it is important to make sure that the noise is added for all the weights, even to those that originally have a zero gradient update. Failure to do so voids the privacy guarantees of DP-Training. \\
\hline

\multirow{10}{*}{\rotatebox[origin=c]{90}{Losses}}
& Cross-Entropy & No & Standard loss that is easy to reason about for DP-Training analysis \\
&
Pair-wise, triplet etc. loss & Yes & Losses that operate on a number of instances at the same time are commonly used for contrastive learning, metric learning, pairwise ranking etc. For instance-level privacy, gradients for each example will depend on another example in the pair. The easiest way to do DP-Training is to add the noise to the already trained model (output/weights perturbation), but this will add more noise (compared to gradient noise injection algorithms like DP-SDG) (refer back to Section \ref{sec:noise-injection}) and thus will affect the utility. There is some work that tackles this setting for convex loss functions via loss perturbation \cite{Huai_Wang_Miao_Xu_Zhang_2020}. \citet{ijcai2021p73} instead look into modifying Projected Gradient Descent (PGD) to achieve DP-Training, under Lipschitz continuity and convexity assumptions. The authors bound the sensitivity of PGD and add noise based on the Gaussian mechanism. 
Alternatively, one can provide (group-level) guarantees per pair of example (if fixed pair assignment is available), which requires running DP-Training and clipping the gradient at a pair level. \\

& Energy-based & Yes & While such losses are common for convolutional deep-belief networks, this loss makes it hard to reason about global sensitivity \cite{DBLP:journals/corr/PhanWD17}. Custom approaches for privacy accounting or approximations of the loss will be required for DP-Training. For example, \citet{DBLP:journals/corr/PhanWD17} derived new polynomial approximation of energy-based loss using Chebyshev's Expansion and injected noise into these polynomials. \\

\hline

\multirow{10}{*}{\rotatebox[origin=c]{90}{Compression}}
& Pruning & No & Usually pruning removes weights or neurons based on their magnitude, and these values are already considered public if the model is trained with DP-Training \\
& Weight quantization &No & The weights of a DP-Trained model are already ``public'' and can be quantized. \\
& Distillation  & Maybe & If one trained a DP teacher model and attempts to distill it into a smaller model, whether this final student model will be private (DP) will depend on the distillation data. If distillation is done on public (student) data, then the student model is still DP, protecting the original teacher's private training data. If one however distills using private (student) data, the student also needs to be trained with DP-Training. Alternatively, DP can be added only during the student training and the resulting model will be considered private, protecting only student training data (and non DP w.r.t. to the teacher training data).
 \\
\hline
\multirow{2}{*}{\rotatebox[origin=c]{90}{Tokenizers (Language)}}
& WordPiece \cite{wordpiece} & Yes & The tokenizer is trained based on the training data prior to the model training. As such, the tokenizer needs to be privatized;  for example as in \citet{hoory-etal-2021-learning}, and the privacy budget consumed by the tokenizer should be accounted for. An alternative is to use a tokenizer that was pretrained on a different public dataset \cite{ponomareva-etal-2022-training}\\
& SentencePiece \cite{DBLP:journals/corr/abs-1808-06226} & Yes & Just as WordPiece, it needs to be privatized; for example, as in \cite{ponomareva-etal-2022-training}. An alternative is to use a tokenizer that was pretrained on a different public dataset \cite{ponomareva-etal-2022-training}\\

\caption{This table lists components and processes that are commonly used in deep learning and describes whether special modifications are required for these modules to be compatible with DP.}
\label{layers-privacy-table}
\end{longtable}

\normalsize
\subsubsection{Design Choices Affecting Model Quality}\label{sec:utility-considerations}
It is fair to say that there is no consensus on how and if the architecture and components choices of an ML model should change when going from the model without DP to its DP version. By far, the most common approach is not to change anything and simply retune the hyperparameters, as discussed in Section \ref{sec:tuning_hyperparams}. Nevertheless, below we attempt to summarize the current state of the research on how model design choices affect model utility.

\paragraph{Activation functions.}
\citet{papernot2020tempered} argued that the choice of activation functions (e.g., RELU, sigmoid, etc.) has an important effect on DP model utility. The authors stated that bounded activation functions like tempered sigmoids outperform unbounded functions like RELU in DP-Training settings. This recommendation stems from the authors' observation that during DP-SGD training, the  activations explode as training  progresses. This in turn results in a  more drastic clipping of the gradients, and therefore, may lead to a worse utility due to the information loss. Tempered sigmoid functions, however, control the gradient norm and reduce the amount of actual gradient clipping that takes place during DP-Training. The authors report improved privacy-utility tradeoffs on three popular datasets. However, tempered sigmoids introduce another hyperparameter, the temperature, which needs to be tuned as well. As in many other papers, the cost of such tuning is not taken into account for the final $\epsilon$ guarantees. However, one limitation of this study is that it does not take into account possible connections between activation functions and other architectural choices \cite{DBLP:journals/corr/abs-2110-08557}. Additionally, unbounded activation functions like RELU have been shown previously to drastically improve the performance of the neural networks  and significantly improve  convergence (e.g., loss stabilization) \cite{NIPS2012_c399862d}.

One the other hand, one of the insights from neural architecture search for DP models performed by \citet{DBLP:journals/corr/abs-2110-08557} is that SELU activation \cite{DBLP:journals/corr/KlambauerUMH17} is more suitable than tempered sigmoids for DP Training. SELU functions, however, result in internal normalization and may require reconfiguration of the architecture and removal of regularization layers like Batch and Layer norm. Another empirical observation presented by \citet{DBLP:journals/corr/abs-2110-08557} is that activation functions that keep negative values (unlike RELU for example) are more effective for DP-Training.

\paragraph{Regularization.}
There are two conflicting groups of work with respect to using additional regularizers when training with DP. Firstly, several works argue that regularization is important for obtaining better utility-privacy tradeoff for DP-Training methods. For example \citet{davody2020effect} argue that normalization layers like BatchNorm are extremely beneficial due to the fact that they make the networks robust to the additional noise (in the weights) during training, and therefore should improve the performance of DP-Training methods like DP-SGD. Authors report substantial improvements (7 to 10\% with a privacy budget of $\epsilon = 0.1$ and 0.05) in performance when BatchNorm layers are introduced and accounted for in privacy calculations. The experiments demonstrate this on both image and natural language models. \citet{anil2021large} report an opposite effect that scale invariant layers like BatchNorm have on the utility: the Gaussian noise injected into the gradients increases the Frobenious norm of the weights during training, which in turn reduces the magnitude of the gradients and slows down the training process. They argue that for the models that use such layers, a large weight decay parameter is needed for Adam optimizer.
\citet{deepmind2022dpimagenet} however observe that for DP trained models, improvements in performance on training data are directly correlated to improvements on test data, hinting at reduced overfiting of DP-Trained models. Therefore, \citet{deepmind2022dpimagenet} argue that explicit regularization like dropout, label smoothing, weight decay, stochastic depth etc can and should be removed.

\paragraph{Optimizers.} 
Choosing an appropriate optimizer and its hyperparameters, e.g., learning rate, are among the most important choices for training machine learning models. SGD and its variants are the most common optimizers for training deep neural networks \cite{Goodfellow-et-al-2016}. Adaptive optimizers, such as Adam \cite{adam} and AdaGrad \cite{JMLR:v12:duchi11a,mcmahan2010adaptive}, have been widely used due to their stability, less need for tuning and fast convergence (e.g., loss stabilization), especially for language models and generative models \cite{DBLP:journals/corr/abs-1810-04805,gpt,brock2018large,ho2022cascaded}. In general, clipping and Gaussian noise can be also applied to the variants of SGD. A simple strategy is to use privatized gradients (e.g., clipped and aggregated with the noise) to compute the moment statistics (first and second moments). This method allows one to reuse the same privacy calculations as for DP-SGD, since the privatized gradients are considered to be public and can be used freely, due to the post-processing property. This simple strategy, e.g., DP-Adam, DP-AdaGrad etc., has been used in many previous works \cite{zhou2020private,anil2021large,li2022llmdp,yu2021differentially}~and is probably most commonly used by practitioners, who ``privatize'' for DP training their best performing optimizer from a non-private setting. However, there are also concerns that this strategy is suboptimal as the noise added to privatize the gradient will reduce the effectiveness of the preconditioner: because prior gradients are noised, statistics that include non-linear transformations (like scaling in Adagrad and root mean square propagation in RMSProp) may accumulate extra noise.
Empirical results in \cite{li2022private} seem to confirm this concern: the authors show that DP-Adam can perform worse than DP-SGD when Adam performs better than SGD in a non-private setting on the same tasks, especially in the high-noise-strong-privacy regime. Some heuristic (empirical)  suggestions to counter this noise were explored recently. For example,  \citet{anil2021large} advocated for a high weight decay value when training with Adam optimizers. Additionally, there is a line of work that explores more sophisticated adaptive optimizers for differentially private training, which often needs additional public data to estimate a more accurate preconditioner \cite{asi2021private,li2022private,kairouz2021nearly}. Designing theoretically grounded differentially private \textit{adaptive} optimizers is an open question and active research topic. 
Finally, a recent study suggested that gradient clipping can improve the performance of standard SGD for non-private training and thus SGD can potentially be used as a replacement of adaptive optimizers \cite{zhang2020adaptive}.
Based on all of the above, we advocate for the following (heuristic) strategy for choosing the optimizer. 
\begin{specialistbox}{Choosing the optimizer}
\begin{compactitem}
    \item If SGD (possibly with momentum and gradient clipping) and adaptive optimizers (e.g., AdaGrad or Adam) perform similarly for non-private training, consider using (momentum) DP-SGD.
    \item  If adaptive optimizers are performing significantly better than non-adaptive methods (SGD) in non-private training, chose the private version of the preferred optimizer and tune its hyperparameters. 
    \item Alternatively, the choice of the optimizer can be viewed as another hyperparameter to tune.
\end{compactitem}
\end{specialistbox}
\paragraph{Model size: small vs large models.}
It has been widely believed that using smaller models with DP-SGD results in better privacy-utility tradeoffs. For example \citet{DBLP:journals/corr/BassilyST14} show that training larger models results in worse generalization when using DP training. This is due to the fact that the norm of the noise needed for DP training is proportional to the (square root of) the number of model parameters \cite{DBLP:journals/corr/BassilyST14,DBLP:journals/corr/abs-2011-11660,kurakin2022dpimagenet,https://doi.org/10.48550/arxiv.2203.00324}. Recently,  however, \citet{deepmind2022dpimagenet} demonstrated that large overparameterized models can perform well with proper hyperparamter tuning and some architectural modifications (e.g., less regularization), achieving new SOTA on CIFAR-10 (by approx. 10\%).

\paragraph{Automatic architecture selection.}
\citet{DBLP:journals/corr/abs-2110-08557} investigated the effect of different model architectures on the utility of the resulting DP-trained models. The authors argue that instead of reusing an architecture that works well for data in a non-DP setting, it is necessary to redesign the model for DP-training. They then propose a neural architecture search based on reinforcement learning. Their approach takes into the account the interplay between various architectural choices. The authors also provide a number of empirical observations based on the best models obtained using neural architecture search; for example, MaxPool layers tend to perform better than Average Pooling layers for DP-Trained models.

To summarize the aforementioned research, 
\begin{specialistbox}{Architectural adjustments for DP ML models}
There is no clear theoretical or empirical consensus on how to adjust the architecture and model components for DP Training in order to maximize the utility. It seems likely that, just as in conventional empirical risk minimization, the choice of architecture is a hyperparameter and the most utility is achieved with proper hyperparameter tuning. \end{specialistbox}

\subsection{Microbatches\Difficult}\label{sec:microbatches}
In general, DP-SGD's requirement of per-example gradient clipping is computationally and memory expensive. Some implementations, such as  Tensorflow Privacy \cite{tf-privacy}, allow one to split the batch into a number of smaller microbatches, process microbatches separately, and aggregate the result. This makes it easier to deal with larger batches, since empirically DP-SGD's utility seems to improve with the batch size.
To our knowledge, there is no clear information in the literature as to whether microbatching changes the privacy guarantees, and below we attempt to remedy this. In the literature and practical implementations there are so far two ways of implementing microbatching.  
\begin{compactenum}
\item \textbf{Option 1}: Split the batch into microbatches (alternatively, draw a number of microbatches and distribute them onto devices). For each microbatch, clip each \textbf{per-example gradient} to have a maximum clipping norm $C$. Aggregate the \textbf{sum} of the gradients from all the microbatches, add noise proportional to the clipping norm and  $\sigma$, as per Algorithm \ref{algo:abadi}.
This option is essentially a classical DP-SGD for multi-device training. This is also equivalent to so-called virtual batch or gradient accumulation (see Section \ref{sec:problems-dp-sgd}). \textbf{The privacy guarantees remain the same} as in no-microbatching setting, and \textbf{the amount of noise added is also the same}.
\item \textbf{Option 2}, implemented in some libraries, for example in  Tensorflow Privacy \cite{tf-privacy}, clips \textbf{the average per microbatch gradient}. The correct way to implement this is presented in Algorithm \ref{algo:microbatch}: for each microbatch, calculate the average gradient. Clip the \textbf{average per microbatch} gradient to the maximum norm $C$. \textbf{Sum the averaged (per-microbatch) gradients}, add  noise proportional to the clipping norm and $\sigma$, divide by the number of microbatches. 
\textbf{The privacy guarantee $\epsilon$ remains the same} as in the no-microbatching setting, but this approach adds more noise (the standard deviation used is \textbf{2k} times larger, where $k$ is the size of each microbatch). Thus, the utility of such algorithm is expected to be worse than the  no-microbatching setting. Smaller microbatches (alternatively, more microbatches) are expected to do better due to less additional noise. Specifically, if each microbatch consists of just one  example, the Algorithm \ref{algo:microbatch} reduces to a standard DP-SGD.
This additional noise in microbatching setting is due to two factors:
\begin{compactenum}
\item The per-example sensitivity changes to \textbf{$2C$}\footnote{Assuming commonly used add-or-remove notion of ``neighbouring'' datasets that was used so far in this paper. For ``replace-one-record'' adjacency (see Section \ref{sec:recordadjacency}), the sensitivity will be equal to 2C in both the microbatch and non-microbatch settings.} in microbatching setting. To see this, recall Definition \ref{def:sensitivity}. A particular aberrant example from a microbatch can change the average per-microbatch gradient from $g$ to $-g$, where $g\ge C$, making the sensitivity $2C$.
\item $k$ times more noise (per example) is added due to the microbatching.
\end{compactenum}

\begin{algorithm}
\caption{DP-SGD with microbatching \cite{DBLP:journals/corr/abs-1812-06210} with a minor correction to account for \textbf{2C} sensitivity}
\label{algo:microbatch}
\begin{algorithmic}
\Require Training data, consisting of features  $X := \{x_1, x_2, ..., x_N\}$ and labels $Y := \{y_1, y_2, ..., y_N\}$. \\
$f(x; \theta)$ is the output of a model parameterized by $\theta$ and applied to an input $x$. \\
$L(Y,f(X; \theta))=\frac{1}{N}\sum_{i} L(y_i, f(x_i; \theta))$ is the empirical risk. \\
SGD hyperparameters: $\eta$ learning rate, $T$ number of iterations, $B$ batch size. \\
DP hyperparameters:  clipping norm $C$,  noise level $\sigma$, number of microbatches per batch $M$.
\Ensure $\theta_T$ final model parameters
\State $\theta_0 \gets $ randomly initialized values
\For{$t \gets 1$ to $T$}                     
     \State Randomly sample a batch $B_t$ with sampling probability $B/N$     
    
    \State $k \gets $ B/M \Comment Number of examples per microbatch.
     
    \For{$m \gets 1$ to $M$} \Comment Process each microbatch. 
     \State $b_m$ $\gets$ indices of $k$ examples from $B_t$ 
    
       \State \HiLi $g_{t}^{(m)} \gets \frac{1}{k} \sum_{i \in b_m} \nabla_{\theta_t}{L(y_i, f(x_i; \theta_t))}$
        \Comment{ Compute average microbatch gradient}
        
         \State  \HiLi $g_{t}^{(m)} \gets g_{t}^{(m)}/max(1, \frac{||g_{t}^{(m)}||_2}{C})$
        \Comment Clip average microbatch gradient
        
     \EndFor
     
     \State  \HiLiLong $\bar g_t  \gets \frac{1}{M}(\sum_m{g_{t}^{(m)}}+\mathcal{N}(0,\,4 \sigma^{2} C^2 \identity ) )$
     \Comment Add noise

    \State $\theta_{t+1} \gets \theta_t - \eta \bar g_t $
    \Comment Gradient descent step
    
    \EndFor     

\end{algorithmic}
\end{algorithm}

Some additional caveats of this microbatching approach are
\begin{compactenum}
\item It does not provide group $k$ level privacy (Section \ref{sec:unitofprivacy}). For $k$ group privacy, the definition of neighbouring datasets change to datasets that differ in \textbf{any} $k$ instances. If fixed partitioning into batching and micrbatches was used for all epochs, then this microbatching algorithm would provide $k$ group privacy; however, such fixed partitioning is not the case for standard training of ML models.
\item When switching from standard DP-SGD to microbatch DP-SGD, the clipping norm threshold $C$ should be retuned. This is due to the fact that in microbatch DP-SGD clipping is applied to the average of the gradients in microbatches, so an appropriate clipping norm can be smaller than in standard DP-SGD.
\end{compactenum}
\end{compactenum}

\subsection{Frameworks and Libraries for DP}
There are many frameworks and libraries for differentially-private training.
They typically differ in their capabilities, and some of them can be faster than the others.
Moreover, DP-frameworks are typically tightly coupled with corresponding machine learning frameworks (e.g.,  Tensorflow, PyTorch, and JAX).
Thus, when dealing with existing code or pre-determined ML framework, practitioners usually have very limited options on what DP framework to choose.
In Table~\ref{table:dp_frameworks}, we provide a list of most popular DP frameworks for various machine learning frameworks.

\begin{table}[htb]
\small
\begin{tabular} {p{0.1\linewidth} p{0.3\linewidth}  p{0.53\linewidth}}
 \toprule
 & \textbf{DP Framework} & \textbf{Description}  \\
\midrule
\multirow{3}{*}{Tensorflow} 
 & Tensorflow Privacy (TFP)\footnote{\url{https://github.com/tensorflow/privacy}} & This is the default framework to perform DP-training in Tensorflow.
 It is mature, feature-rich and well maintained.
 The main drawback is that it is  relatively slow.\\\cmidrule{2-3}
 & Tensoflow Federated (TFF)\footnote{\url{https://www.tensorflow.org/federated}} & This framework works together with Tensorflow Privacy to facilitate federated learning and user-level differentially private training. \\ 
 \midrule
{Google DP \footnote{\url{https://github.com/google/differential-privacy/tree/main/python}}} & General purpose/any &
This library provides general-purpose DP accounting functionality for many common mechanisms including DP-SGD and DP-FTRL. It works well with TFP, but can easily be used to compute privacy costs for other implementations. \\
\midrule
Pytorch & Opacus~\cite{opacus} & This is the main DP-framework for Pytorch.
It provides efficient implementations of per-example gradients for various common neural network layers. \\
\midrule
JAX & Various Libraries & At the time of publishing of this survey, JAX did not have a single universal framework for differential privacy. Nevertheless, various authors released differentially private implementations for specific tasks~\cite{deepmind2022jax-privacy,kurakin2022dpimagenet,ponomareva-etal-2022-training}. \\
\bottomrule
\end{tabular}
\caption{This table lists various frameworks for machine learning with differential privacy. DP frameworks are grouped by machine learning frameworks they are compatible with (left column).}\label{table:dp_frameworks}
\end{table}

\section{Conclusion}
While Differential Privacy is gaining popularity in academic and industrial settings, training a complex ML model like a deep neural net with DP remains a non-trivial task, both due to utility drop, computational cost and a number of model components that should be made DP (like tokenizers, various layers, different losses etc.). 

In this survey paper we compiled a summary of the current research body related to making ML models DP, and provided practical tips on how to achieve the best privacy-utility tradeoffs and what $\epsilon$ guarantees to target. We argued for careful consideration and explicit reporting of commonly glanced over areas such as whether amplification assumptions hold, the unit of privacy that was used, the definition of ``neighbouring'' datasets and how hyperparameter tuning was performed. We drew attention of practitioners to the fact that for complex models careful examination and possible adjustment of the model components is often required in order to both preserve privacy and to improve model performance. 

Our hope is that this self-contained guide will make applications of DP to ML models easier and faster to adopt and will serve as a reference point for the practioners who want to know \textit{``just enough''} in order to correctly apply DP to complex ML models.

\section*{Acknowledgements}
The authors wish to thank Peter Kairouz, Ryan McKenna, Daniel Ramage and Thomas Steinke
 for useful comments and discussions on the manuscript. Ryan McKenna suggested visualizing the importance of batch size via Figure~\ref{fig:batchsize}. Alina Oprea contributed helpful discussion and writing for the coverage of empirical privacy attacks and their implications for the proposed privacy tiers.

\bibliography{biblio}
\bibliographystyle{apa-good}

\appendix

\section{DP-Training for non-differentiable models.}\label{app:nondif}
We have shown in the main paper that applying differential privacy to differentiable models for the large part requires mostly straightforward changes to the optimization algorithm (e.g. going from SGD to DP-SGD) and a careful choice of hyperparameter values. This recipe is universal for all differentiable models. However non-differentiable models require custom adaptations to their algorithm in order to induce privacy. In general, one can either
\begin{compactitem}
\item Replace a non differentiable algorithm with a differential approximation and apply existing DP-Training methods like DP-SGD. Privacy accounting is well established and easy. An example of this approach would be replacing a non differentiable tree-based model like CART \cite{BreiFrieStonOlsh84}  with a soft tree model like \cite{DBLP:journals/corr/abs-2002-07772}, which is differentiable and can be trained with DP-SGD.
\item Alternatively, one can modify the original algorithm to make all the statistics calculated over the data private, using well established mechanisms like Laplace, exponential, Gaussian etc. \cite{dwork}. Careful custom privacy accounting is required.
\end{compactitem}

For the latter approach of making all the statistics of the model private, there are a number of generic rules that hold for all the models. They are as follows:

\begin{compactitem}
\item Any statistics calculated using the original data need to be modified by applying an appropriate noise mechanism. For example, if a model calculates and subsequently uses quantiles of the feature values, or if a final prediction is calculated over all or some of the data like taking a mean etc -- these statistics will meed tp be privatized.
\item When introducing noise to various  parts of the algorithm, careful consideration as to whether to introduce noise uniformly or split the privacy budget and use less or more noise in some parts of the algorithms is important.
\item Custom privacy accounting to calculate overall $\epsilon$ of the algorithm is required. R\'enyi DP provides easy composition and tight bounds for conversion back to $(\epsilon, \delta)$. Refer back to Section \ref{sec:dpaccounting}.
\item Any distributed calculation of statistics should be implemented carefully. For example, when an algorithm calls for $\sigma^2$ noise added to an average of the data, when it is distributed accross a number of workers so each worker calculates an average of the data it processes and adds the \textit{independent} (e.g. not the same) noise, the noise per worker should be scaled by the number of workers (e.g. each worker adds more noise).  
\end{compactitem}

Below we go through a number of popular non-differentiable models and briefly outline a) what data-dependant statistics these models calculate and b) point the reader to the papers that implement the changes required for making an algorithm differentially private,  c)survey papers that investigate how to split the privacy budget between various steps of the algorithm  d) outline some alternatives.

\subsection{Tree-based algorithms}
\textbf{Statistics to be privatized}: Splitting rules (histogram of values, feature value for the split), leaves prediction values.

Many of the most popular non-differentiable models are based on decision trees. These include models like CART \cite{BreiFrieStonOlsh84} and more modern variants like Gradient Boosted Decision Trees \cite{gbm}. Such trees are usually learnt in a greedy top-down layer-by-layer fashion. At each iteration, an optimal split (a feature and its value) is chosen and the data is routed to the children according to the split decision. At the bottom level, some statistics like average value of the instances in a leaf or average of gradients, are calculated. These values determine leaves values and are used for subsequent prediction \cite{BreiFrieStonOlsh84}. 

A good comprehensive survey on applying DP to tree based models can be found in \cite{DBLP:journals/corr/FletcherI16a}. Authors investigate in detail various trade offs when applying DP (what noise mechanism to chose, whether to use a random splits vs best split etc). They highlight that when choosing the best splits the statistics used for choice need to be noised, but also the actual continuous best value of the feature should be privatized - for example, by being drawn randomly from some range. Further, \cite{DBLP:journals/corr/abs-2012-10602} introduce a new generic top down learning algorithm DP-TopDown, that is applicable to a distributed setting. A private split subroutine that selects the best split is provided. This paper advocates for a decay budgeting strategy, where less noise is added for the earlier splits which are more important. Surprisingly, they report that the private algorithm sometimes results in a better utility than its non-private counterparts. They hypothesize that it is due to greedy algorithms choosing "unlucky" splits that overfit, whereas injecting noise into the split decision allows the model to avoid such overfitting. Similarly \cite{Li_Wu_Wen_He_2020} adapts gradient boosted decisions trees to be differentially private. 

A common alternative to choosing the best possible split is to choose a split randomly, like in \cite{Geurts03extremelyrandomized}. For such an algorithm, only the final leaf value is based on the data and thus needs to be privatized by adding appropriate level of noise.
Finally, as mentioned previously, tree based models can be replaced with their differentiable counterparts - Soft trees (e.g. \cite{DBLP:journals/corr/abs-2002-07772}) and subsequently trained conventionally using DP training procedures. 

\subsection{Clustering algorithms}

In this section, we discuss practical algorithms for differentially private k-means. Given a set of data points, k-means aims to partition the points into k disjoint sets (clusters), with the objective of minimizing the within-cluster sum of squares (WCSS). Solving k-means to optimality is generally NP-Hard \cite{aloise2009np}, so many efficient heuristics and approximation algorithms have been developed. A standard heuristic is  Lloyd's algorithm, which iteratively improves WCSS by repeating the following two steps until meeting some convergence criterion: (i) assign each point to its closest cluster center, and (ii) update each cluster center to the arithmetic mean of the points in it. In what follows, we focus on DP-Lloyd--a popular differentially private variant of Llyod's algorithm. After discussing DP-Lloyd, we give pointers to alternative popular algorithms in the literature.

\textbf{Statistics to be privatized}: centroid values, possibly the process of choosing the closest centroid.

\textbf{DP-Lloyd}: This was first proposed by \cite{blum2005practical} and later implemented in the PINQ platform \cite{mcsherry2009privacy}. The idea is to use the standard Lloyd's algorithm and simply add noise during each cluster center update, where the noise is chosen according to a standard DP mechanism (e.g., Laplace mechanism). Specifically, when computing the arithmetic mean of the points in the cluster, noise is added to the sum (numerator) and the counts (denominator). Next, we discuss practical tips on how to make two key choices: the number of iterations and the initial cluster centers factors. 

\textsl{Number of iterations and Privacy Budget}: Since noise is added during each iteration, the privacy budget will be divided across iterations. For a fixed budget, running the algorithm with more iterations will require adding noise with larger magnitudes, which might have an adverse effect on the clustering quality. On the other hand, too few iterations may not be sufficient for the algorithm to converge. Thus, it is important  to choose a suitable number of iterations. For example, in \cite{mcsherry2009privacy}'s implementation, the default number of iterations for DP-Lloyd is 5, and \cite{su2016differentially}'s experiments indicate that DP-Lloyd converges (gets close to a local minimizer) in 5 iterations on a variety of real datasets with $2$ to $10$ dimensions. However, generally, the number of iterations needed  depends on the dataset, $k$ number of clusters and the quality of the initial cluster centers.

If the number of iterations is set a-priori, a common strategy is to distribute the privacy budget equally across iterations. If the number of iterations is unknown, one strategy is to decrease the privacy budget allocated to each subsequent iteration, in a way that respects the overall privacy budget. More formally, let $\epsilon$ be the overall privacy budget and $\epsilon_t$ be the budget allocated to the $t$-th iteration of the algorithm. Any sequence $\{ \epsilon_t \}$ satisfying  $\sum_{k=1}^{\infty} \epsilon_t =\epsilon$ will respect the overall privacy budget. For example, one possible choice is the geometric sequence $\epsilon_t = 2^{-t} \epsilon$ \cite{dwork2011firm}. However, we note that the performance is expected to deteriorate after a certain number of iterations (as the noise level increases indefinitely with the number of iterations).

\textsl{Initial Cluster Centers}: The quality of the initial cluster centers determines the number of iterations needed by DP-Lloyd to converge and consequently the level of noise needed. For example, if the initial cluster centers are very close to the optimal centers, the algorithm can converge in a small number of iterations and thus requires minimal noise. In the standard (non-private) Lloyd's algorithm, it is common to run the algorithm for many randomly sampled initial centers and pick the center with the lowest WCSS, or use careful seeding strategies such as k-means++ \cite{arthur2006k}. For DP-Lloyd, if there is a sufficient privacy budget, one strategy is to run the algorithm for multiple, randomly sampled initial clusters--with the privacy budget distributed across runs. \cite{su2016differentially} suggests an alternative seeding method for DP-Lloyd, which samples cluster centers randomly but under that constraint that each sampled cluster center is sufficiently far from all existing cluster centers (specifically, the distance between each initial cluster center is above a user-specified threshold). \cite{su2016differentially} reports that the latter method typically works better than using a single random initialization. 

\textbf{Popular Alternative Algorithms} 

\textsl{Sample and Aggregate:} Another popular DP k-means algorithm is based on the Sample and Aggregate (SaF) framework  \cite{nissim2007smooth}. The high-level idea is to first partition the dataset into multiple subsets, and then run a non-private k-means algorithm (e.g., Lloyd's algorithm) on each subset. Finally, the cluster centers obtained from each subset are aggregated under a standard DP mechanism. A main advantage from SaF is that the partitioning leads to a relatively low sensitivity--removing one example only affects a single partition, which makes a small contribution on the final (aggregated) cluster centers (assuming a large number of partitions). However, theoretically, for SaF to work well, the data should be well-separated so that the cluster centers can be well-estimated from small samples. Thus, if the data is  well-separated, SaF may outperform DP-Lloyd because SaF is expected to require less noise (due to its low sensitivity). However, \cite{su2016differentially}'s experiments indicate that DP-Lloyd outperforms SaF on a collection of synthetic and real datasets.

\textsl{Synopsis and Hybrid Algorithms:} \cite{su2016differentially} proposes an alternative synopsis-based algorithm. The algorithm divides the input space into $M$ equi-sized cells (boxes), and then outputs a DP synopsis consisting of (i) the center of each box, and (ii) the count of points in each box with noise added according to a standard DP mechanism. Since the synopsis is private, any non-private k-means algorithm could be applied to it. \cite{su2016differentially}'s experiments show that the synopsis approach outperforms DP-Lloyd on datasets with 2 or 3 dimensions, but performs worse on datasets with larger dimensions. Thus, the synopsis method seems suitable only for very low-dimensional problems. \cite{su2016differentially} also reported  success with a hybrid approach, which uses the output of the synopsis method as a initial cluster centers for DP-Lloyd (with the privacy budget split in half between the two methods).

\section{Derivation of DP-SGD cost per epoch}
\label{app:epsilon_epoch_derivation}
We make the assumption that our training is done with uniform sampling \textit{with replacement}. This is almost always violated in practice, but is necessary for the analysis.
For each batch, we sample $L$ of the $N$ total data points. This gives us a sampling ratio of $q = \frac{L}{N}$. We have a noise multiplier $\sigma$ and a clipping norm $C$. For DP-SGD, our standard deviation of noise is $\sigma C$. Using the gaussian mechanism, if our noise standard deviation is 
$$ \sigma C = \frac{\sqrt{2 \ln{\frac{1.25}{\delta}}}}{\epsilon}$$
Then the mechanism is $(\epsilon, \delta)$ differentially private. Since each sample has probability $q$ of being in the batch, this mechanism is actually $(q\epsilon, q\delta)$ (assuming $\epsilon \le 1$, Section \ref{sec:amplification}) differentially private with respect to the whole batch. This means that:
$$ \sigma C = q \frac{\sqrt{2 \ln \frac{9 q}{8 \delta}}}{\epsilon}$$
Rearranging this formula, we see that for a single batch:
$$  \epsilon = q \frac{\sqrt{2 \ln \frac{9 q}{8 \delta}}}{\sigma C}$$
Using the advanced composition formula \cite{dwork}, we can compose $k$ steps at a privacy cost of:
$$ \Tilde{\epsilon} = \epsilon \sqrt{2 k \ln \frac{1}{\delta'}} + k \epsilon \dfrac{e^{\epsilon} - 1}{e^{\epsilon} + 1}$$
Since the $\epsilon$ for a single batch satisfies $\epsilon << 1$ we can approximate $$\dfrac{e^{\epsilon} - 1}{e^{\epsilon} + 1} \approx \frac{\epsilon}{2}$$
Making this approximation:
$$\Tilde{\epsilon} = \epsilon \sqrt{2 k \ln \frac{1}{\delta'}} + \dfrac{k \epsilon^2}{2}$$
Then substituting in for $\epsilon$
$$\Tilde{\epsilon} = q \frac{\sqrt{2 \ln \frac{9 q}{8 \delta}}}{\sigma C} \sqrt{2 k \ln \frac{1}{\delta'}} + \dfrac{k (q \frac{\sqrt{2 \ln \frac{9 q}{8 \delta}}}{\sigma C})^2}{2}$$
We care about the relationship between the sampling ratio $q$, the noise multiplier $\sigma$, and the number of batches $k$ on $\epsilon$. To get a more manageable expression we first drop logarithmic factors and then combine factors which do not depend on $\sigma$, $k$, or $\epsilon$.
$$\Tilde{\epsilon} = A \dfrac{q \sqrt{k}}{\sigma} + B \dfrac{k q^2}{\sigma^2}$$
This shows that the privacy cost of the first few batches is high, and then the change in the $\sqrt{~}$ term quickly becomes zero, causing the overall cost to be linear in the number of batches. 

Using smaller batches but keeping the total number of epochs fixed changes $q \rightarrow q / x$ and $k \rightarrow k x$ where $x$ is the ratio of the original to new batch size. This means 
$$\epsilon_{new} = A \dfrac{q \sqrt{k}}{\sigma \sqrt{x}} + B \dfrac{k q^2}{x \sigma^2} < \epsilon_{old}$$

Thus using smaller batches for hyperparameter tuning leads to a lower privacy cost.

\section{Example comparison of hyperparameter tuning accounting methods}
In this Section we provide an example that demonstrates how to reason about various hyperparameter tuning algorithms. In particular, we compare RDP accounting, PLD accounting, the Exponential Mechanism \cite{Abadi_2016}, and Randomized Number of Trials Algorithm \cite{DBLP:conf/iclr/Papernot022} with truncated negative binomial distribution and Poisson distribution.
\label{app:hparam-tuning-comparison}

We open source the code used for this section \footnote{Code used for this appendix is here: \url{https://gist.github.com/carsondenison/d69e0b86f98af6d4f2d086d26859f6ec}}.

Since it is nontrivial to compare the three algorithms for a general case, we instead will work through one particular example here.

Assume we have 1,000,000 training data points, we do DP-Training (DP-SGD) with a batch size of 5,000 using noise multiplier $\sigma=1.0$, we want to test 100 distinct sets of hyperparameters, and we have a validation set of 10,000 data points.

The $(\epsilon, \delta)$ DP cost of a single epoch is $(\epsilon_{single~run} = 1.2, \delta = 1e-6)$, which occurs at an RDP of $(\lambda = 10.29, \epsilon = 0.0839)$.

\subsection{RDP composition}
Using RDP composition, we find that the total cost of $100$ epochs of training is $(\epsilon, \delta) = (4.95, 1e-6)$.\\

\subsection{PLD composition}
Using the Privacy loss distribution composition instead of RDP, we find that the total cost of one epoch is only $(\epsilon, \delta) = (0.59, 1e-6)$ and the cost for $100$ epochs of training is $(\epsilon, \delta) = (4.62, 1e-6)$.\\

\subsection{Exponential mechanism from \citet{Abadi_2016}}
Using the scheme from \citet{Abadi_2016}, we must select a target accuracy. We want our best chosen trial to be within 1\% accuracy of the actual best model, with probability $0.99$. This means that our answer must be within $100$ samples of the best, so:
$$100 = \frac{4}{\epsilon'} \ln\frac{100 * 100}{\epsilon'}$$
Solving this equation, and substituting in $\epsilon_{tuning} = 8 \epsilon'$, we find that the total epsilon cost of hyperparameter tuning is $3.24$. As a remark in appendix D of \citet{Abadi2016}, the authors note that the total privacy cost is $\max(\epsilon_{single\_run}, \epsilon_{tuning})$, for a total $(\epsilon, \delta)$ cost of $(3.24, 1e-6)$. 

This has the additional complications that the accuracy of the returned model is up to 1\% worse than the best set of hyperparameter values (because of the exponential mechanism) with probability $0.99$, and with probability $0.01$ the accuracy is even worse. Additionally, because we randomly choose hyperparameters with replacement, there is a chance of 
$$\big(\frac{99}{100}\big)^{100} \approx \frac{1}{e}$$
that each set of hyperparameter values never gets run.\\

\subsection{Randomized number of trials from \citet{DBLP:conf/iclr/Papernot022}}
Finally, we use the approaches from \citet{DBLP:conf/iclr/Papernot022}. There are three approaches we will cover. Two use the Truncated-Negative-Binomial-distribution to draw the number of hparam runs, and one will use the Poisson distribution.

The Truncated-Negative-Binomial-distribution scheme from \citet{DBLP:conf/iclr/Papernot022} has two parameters: The first, $\eta$ controls the shape of the distribution. The second, $\gamma$ controls the mean of the distribution, given a fixed $\eta$. Larger $\eta$ leads to better concentration around the mean, but worse privacy.

\subsubsection{Truncated negative binomial distribution with $\eta = 0$}

We first use the Truncated-Negative-Binomial-distribution with $\eta = 0$. This is also known as the logarithmic distribution. This has probability density function:

$$ \mathds{P}[K=k] = \dfrac{(1 - \gamma)^k}{k \cdot \ln(1/\gamma)}$$
Where $\gamma$ is a chosen parameter. The mean is:
$$ \mathds{E}[K] = \dfrac{\frac{1}{\gamma} - 1}{\ln(\frac{1}{\gamma})}$$
We want our mean to be $100$, so:
$$K = 100 = \dfrac{\frac{1}{\gamma} - 1}{\ln \frac{1}{\gamma}} \implies \gamma = 0.00154212$$

We can get the privacy cost of tuning a number of runs drawn from the above distribution using Theorem 2 of \citet{DBLP:conf/iclr/Papernot022}. This states that given two RDP pairs $(\lambda, \epsilon)$ and $(\hat \lambda, \hat \epsilon)$ values for a single run, the total Renyi DP cost of drawing a number $K$ from the above distribution and training $K$ models in the hyperparameter sweep and releasing the best set of hyperparameters is $(\lambda, \epsilon')$ where:
$$\epsilon' = \epsilon + (1 + \eta) \cdot \big( 1 - \frac{1}{\lambda}\big) \hat \epsilon + \frac{(1 + \eta) \cdot \ln(1/\gamma)}{\hat \lambda} + \frac{\ln(\mathds{E}[K])}{\lambda - 1}$$
In our case, we will take the optimal values from above: $(\lambda = 10.29, \epsilon = 0.0839)$ as our $\hat \lambda$ and $\hat \epsilon$ and then compute the increased cost at each $\lambda$ and re-convert to $(\epsilon, \delta)$ privacy.

Plugging in our values to Theorem 2 of \citet{DBLP:conf/iclr/Papernot022} gives us new RDP values for each order, which we can convert to a new $(\epsilon, \delta)$ of $(2.42, 1e-6)$.

This is the tightest privacy bound of the three approaches, but it comes with two constraints: First, like in the previous approach, there is a chance that each set of hyperparameters is not used. Second, although the mean number of runs is 100, the distribution is very un-concentrated. In particular, the mode of the distribution is the value 1 - which occurs with probability $0.154$, and there is a 60\% chance that the number of trials will be less than 50. This means there is a decent chance that hyperparameter tuning with this method will lead to poor results.

\subsubsection{Truncated negative binomial distribution with $\eta = 1$}

We can do this analysis again with the Truncated-Negative-Binomial-distribution, using $\eta = 1$, which is just the Geometric distribution. This makes the probability mass around very small numbers lower, which is good for the reliability of the tuning algorithm, but will have a higher privacy cost. Now the probability distribution is:

$$ \mathds{P}[K=k] = \dfrac{(1 - \gamma)^k}{\frac{1}{\gamma} - 1}$$
Where $\gamma$ is a chosen parameter. The mean is:
$$ \mathds{E}[K] = \frac{1}{\gamma}$$
We want our mean to be $100$, so:
$$K = 100 = \frac{1}{\gamma} \implies \gamma = 0.01$$

Doing the same steps as above, this gives us an epsilon delta privacy cost of $(\epsilon = 2.76, \delta=1e-6)$ to do $K$ runs and return the best set of hyperparameters. This is much better than above, because now although the mode is still $1$, there is only a probability of $0.01$ of getting $1$ hyperparameter tuning run, and there is only a probability of $0.39$ of getting fewer than $50$ runs. This is still not great, but it is an improvement.

If we increase our desired mean to $1000$, to be sure to get more runs, we will get a higher privacy cost as well. In this case we set $\gamma = 0.001$. This increases the total privacy cost to $(\epsilon = 3.45, \delta=1e-6)$, which is on par with the method from \citet{Abadi_2016}. Now there is only a probability of $0.094$ of being able to run fewer than $100$ runs, and less than a $1\%$ chance of being able to run fewer than $10$. However, a $~10\%$ chance of being able to run fewer than 10 runs is still a daunting prospect for a machine learning practitioner.

\subsubsection{Poisson distribution}
The Poisson-sampled method from the same paper is much more centered, with a negligible chance of getting less than 50 trials. However, it has a much larger privacy cost than the previous two methods. 

With this method, we draw from the distribution:
$$\mathds{P}[K=k] = e^{-\mu} \frac{\mu^k}{k!},$$
where $\mu$ is the mean of the distribution.

For this privacy accounting scheme, we need an $(\hat \epsilon, \hat \delta)$ guarantee. We denote this as $\hat \epsilon$ and $\hat \delta$ to distinguish from our $(\lambda, \epsilon)$ guarantee. For each RDP order $\lambda$, we must compute the optimal $\hat \delta$ such that with our $(\epsilon, \delta)$-DP guarantee we have that 
$$\epsilon = 1 + \frac{1}{\lambda - 1}$$
Then we compute the new RDP as:
$$\epsilon' = \epsilon + \mu \cdot \hat \delta + \frac{\ln \mu}{\lambda - 1}$$
As before, we then convert back to $\epsilon,\delta$ DP, and get the final privacy cost. 

There are two ways to do this. \\

We need an $(\epsilon, \delta)$ guarantee for a single epoch in order to use this method, and we can get that either through RDP or PLD accounting. If we use RDP, our single epoch cost is $(\epsilon=1.2, \delta=1e-6)$, which leads to a final epsilon of: $(\epsilon = 4.18, \delta=1e-6)$. This is worse than privacy with the Truncated-Negative-Binomial distribution, and worse than the method from \citet{Abadi_2016}, but it is extremely well concentrated around the mean unlike the Negative-Binomial approach, \textit{and} we return the true best set of hyperparameters instead of having to use the exponential mechanism as in \citet{Abadi_2016}. 

However, if we use the PLD accountant instead, our $\hat \delta$ for a single epoch become \textit{much} smaller: $(\epsilon=0.59, \delta=1e-6)$. This means our final epsilon cost for 100 epochs with the poisson distribution is actually only $(\epsilon=2.63, \delta=1e-6)$. This is much better than RDP, PLD, and exponential mechanism accounting. Additionally, it is nearly as good as the truncated negative binomial approach with $\eta = 0$. However, it also has \textit{much better} concentration around the mean. In this case this is by far the best approach, but only when using the PLD accountant to compute the single-epoch epsilon and delta. 

\section{Additional notes on terms used}
We want to highlight that in ML community a number of terms is used loosely with different meaning. While we attempted to clarify such terms in the paper, below we list some that have several widely accepted meanings. 
\begin{compactitem}
\item \textbf{Privacy guarantees} - in this scope of this work, we use this term to describe \textit{data anonymization guarantees} \cite{bonawitz22cacm}.
\item \textbf{Convergence}: The term ``convergence'' is often used to refer to different notions, including (i)  ``convergence to a stationary solution (e.g., zero gradient)'', (ii) convergence to a global optimum, and (iii)  ``convergence in the loss'' (i.e., the loss stabilizes).
\item \textbf{Batch and microbatch}: we use "batch" to refer to a portion of a training data used for SGD (Stochastic Gradient Descent) update. This is in contrast to a ``full batch'' that implies full training data (and is used for Gradient Descent Algorithm). Batch can be split into microbatches, for example for distributing on different cores.

\end{compactitem}

\end{document}